\documentclass[10pt,journal,compsoc]{IEEEtran}

\ifCLASSINFOpdf
\else
\usepackage[dvips]{graphicx}
\fi
\usepackage{url}

\usepackage[colorlinks,linkcolor=blue,anchorcolor=blue,
citecolor=blue]{hyperref}
\usepackage{graphicx}
\usepackage{amsmath,amssymb}
\usepackage{booktabs,tabularx,multirow,threeparttable}
\usepackage{float}
\usepackage{subfigure}
\usepackage{amssymb}
\usepackage{subfigure}
\usepackage{framed}

\usepackage{comment}
\usepackage{amsmath,algorithm,algpseudocode,amsfonts}

\usepackage{amsthm,amsmath,amsfonts,amssymb,bm}
\ifCLASSOPTIONcompsoc
  \usepackage[nocompress]{cite}
\else
  \usepackage{cite}
\fi

\newcommand{\zhang}[1]{{\color{black} #1}}

\newcommand{\rev}[1]{{\color{black} #1}}

\begin{document}
	\title{Detecting Rotated Objects as Gaussian Distributions and Its 3-D Generalization}

	\author{Xue~Yang, Gefan~Zhang, Xiaojiang~Yang, Yue~Zhou, Wentao~Wang, \\Jin~Tang, Tao~He, Junchi~Yan~\IEEEmembership{Senior Member,~IEEE}
		
	
	\thanks{X.~Yang, G.~Zhang, X.~Yang, Y.~Zhou, W.~Wang, J.~Yan are with School of Electronic Information and Electrical Engineering, and MoE Key Lab of Artificial Intelligence, Shanghai Jiao Tong University, Shanghai, China. J. Yan is also with Shanghai AI Laboratory, Shanghai, China. J. Tang is with Anhui Province Key Laboratory of Multimodal Cognitive Computation, Hefei, China, and Anhui University, Hefei, China. T. He is with Cowa Robot, Co Ltd, Wuhu, China, and Anhui Province Key Laboratory of Multimodal Cognitive Computation, Hefei, China. G. Zhang is also with Cowa Robot, Co Ltd.}
	\thanks{E-mail: \{yangxue-2019-sjtu, lizaozhouke, yangxiaojiang, sjtu\_zy, wwt117, yanjunchi\}@sjtu.edu.cn, tj@ahu.edu.cn, tommie.he@cowarobot.com}
	\thanks{Correspondence author: Junchi Yan.}
	}

	\markboth{}%
	{Shell \MakeLowercase{\textit{et al.}}: Bare Demo of IEEEtran.cls for IEEE Journals}

\IEEEtitleabstractindextext{
	\begin{abstract}
		\rev{Existing detection methods commonly use a parameterized bounding box (BBox) to model and detect (horizontal) objects and an additional rotation angle parameter is used for rotated objects. We argue that such a mechanism has fundamental limitations in building an effective regression loss for rotation detection, especially for high-precision detection with high IoU (e.g. 0.75). Instead, we propose to model the rotated objects as Gaussian distributions. A direct advantage is that our new regression loss regarding the distance between two Gaussians e.g. Kullback-Leibler Divergence (KLD), can well align the actual detection performance metric, which is not well addressed in existing methods. Moreover, the two bottlenecks i.e. boundary discontinuity and square-like problem also disappear. We also propose an efficient Gaussian metric-based label assignment strategy to further boost the performance. Interestingly, by analyzing the BBox parameters' gradients under our Gaussian-based KLD loss, we show that these parameters are dynamically updated with interpretable physical meaning, which help explain the effectiveness of our approach, especially for high-precision detection. We extend our approach from 2-D to 3-D with a tailored algorithm design to handle the heading estimation, and experimental results on twelve public datasets (2-D/3-D, aerial/text/face images) with various base detectors show its superiority.}

	\end{abstract}
	
	}
	\maketitle
	
	\section{Introduction}\label{sec:introduction}
	\IEEEPARstart{R}{otated} objects are ubiquitous for visual detection scenarios, such as aerial images~\cite{yang2018automatic, yang2019scrdet,ming2021sparse}, scene text~\cite{zhou2017east, liu2018fots, ma2018arbitrary, liao2018rotation}, face~\cite{shi2018real} and 3-D objects~\cite{zheng2020rotation, yin2021center}, retail scenes~\cite{chen2020piou, pan2020dynamic}, etc. Compared with the abundant literature on horizontal object detection~\cite{ren2015faster,lin2017feature,lin2017focal}, many oriented detectors build themselves upon the well established horizontal detection pipelines. However, these detectors still encounter challenges in the emerging rotation detection cases, e.g. large aspect ratio objects, dense scenes with rotated objects, especially for high-precision purpose detection, namely the detector is required to achieve high Intersection over Union (IoU) e.g. 0.75 or even higher.
	
	\begin{figure}[!tb]
    	\centering
    	\subfigure{
    		\begin{minipage}[t]{0.47\linewidth}
    			\centering
    			\includegraphics[width=0.98\linewidth]{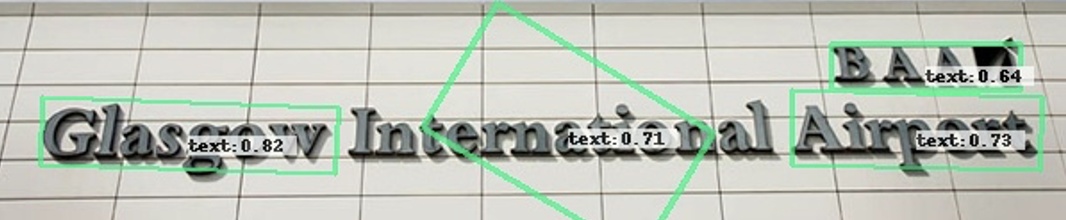}
    		\end{minipage}%
    		\label{fig:compare_vis_1}
    	}
    	\subfigure{
    		\begin{minipage}[t]{0.47\linewidth}
    			\centering
    			\includegraphics[width=0.98\linewidth]{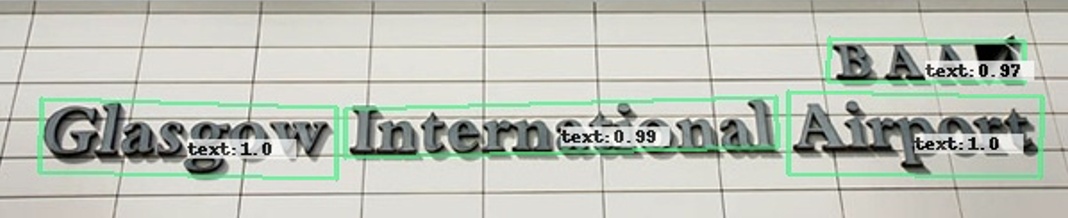}
    		\end{minipage}
    		\label{fig:compare_vis_2}
    	}\\
    	\vspace{-6pt}
    	\subfigure{
    		\begin{minipage}[t]{0.47\linewidth}
    			\centering
    			\includegraphics[width=0.98\linewidth]{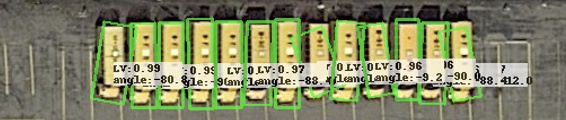}
    		\end{minipage}%
    		\label{fig:compare_vis_3}
    	}
    	\subfigure{
    		\begin{minipage}[t]{0.47\linewidth}
    			\centering
    			\includegraphics[width=0.98\linewidth]{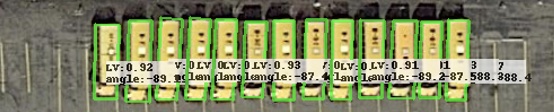}
    		\end{minipage}
    		\label{fig:compare_vis_4}
    	}\\
    	\vspace{-6pt}
    	\subfigure{
    		\begin{minipage}[t]{0.47\linewidth}
    			\centering
    			\includegraphics[width=0.98\linewidth]{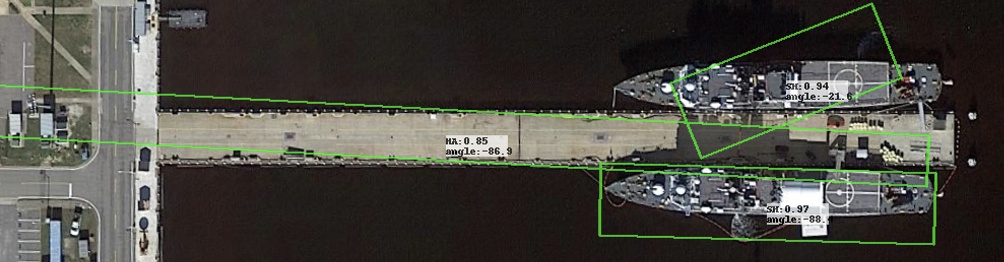}
    		\end{minipage}%
    		\label{fig:compare_vis_5}
    	}
    	\subfigure{
    		\begin{minipage}[t]{0.47\linewidth}
    			\centering
    			\includegraphics[width=0.98\linewidth]{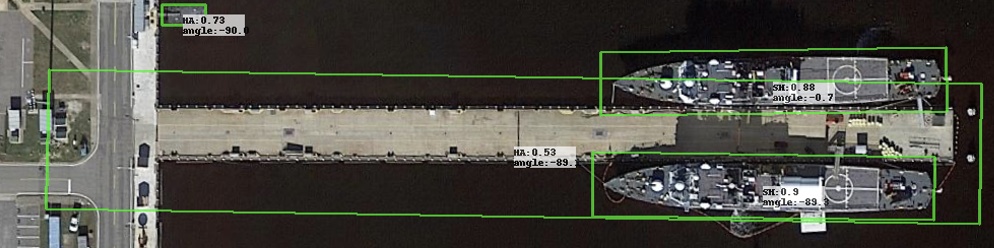}
    		\end{minipage}
    		\label{fig:compare_vis_6}
    	}\\
    	\vspace{-6pt}
    	\subfigure{
    		\begin{minipage}[t]{0.47\linewidth}
    			\centering
    			\includegraphics[width=0.98\linewidth]{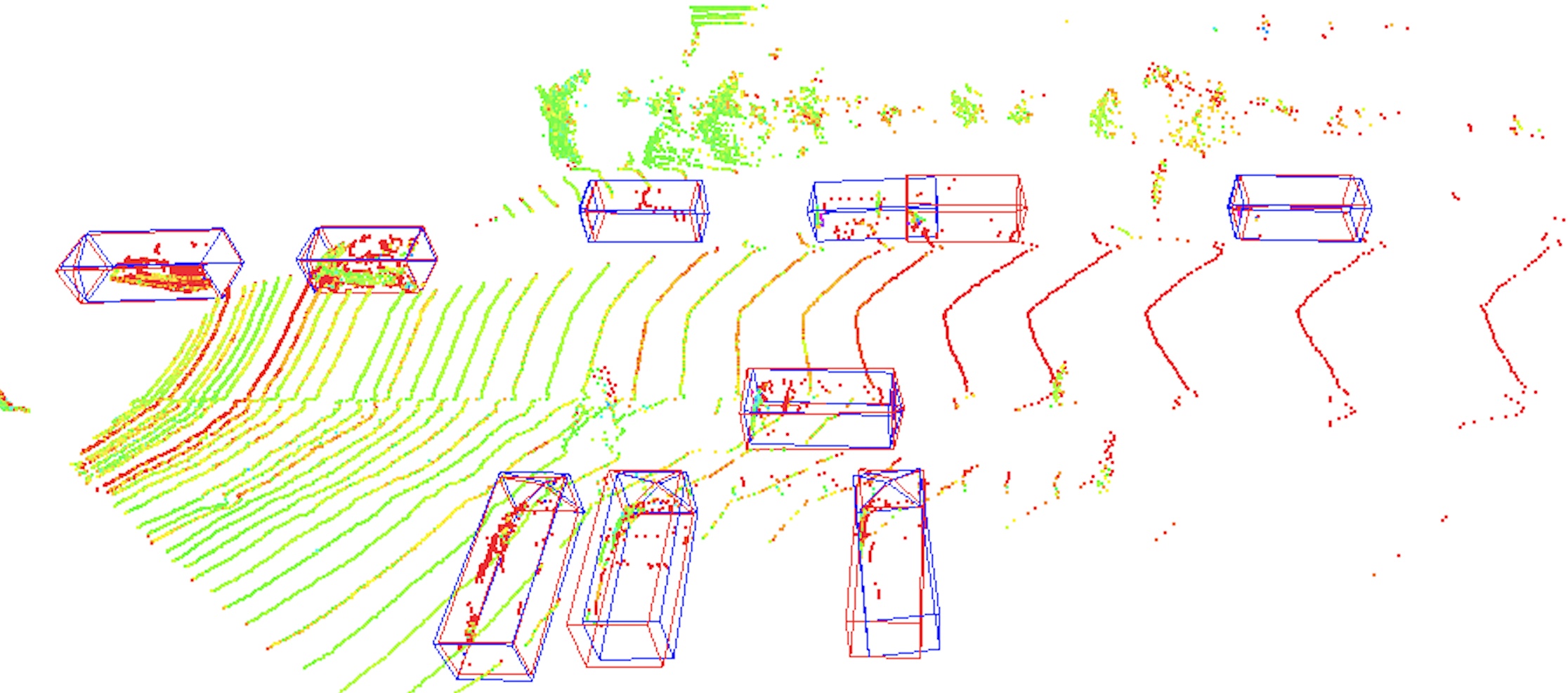}
    		\end{minipage}%
    		\label{fig:compare_vis_7}
    	}
    	\subfigure{
    		\begin{minipage}[t]{0.47\linewidth}
    			\centering
    			\includegraphics[width=0.98\linewidth]{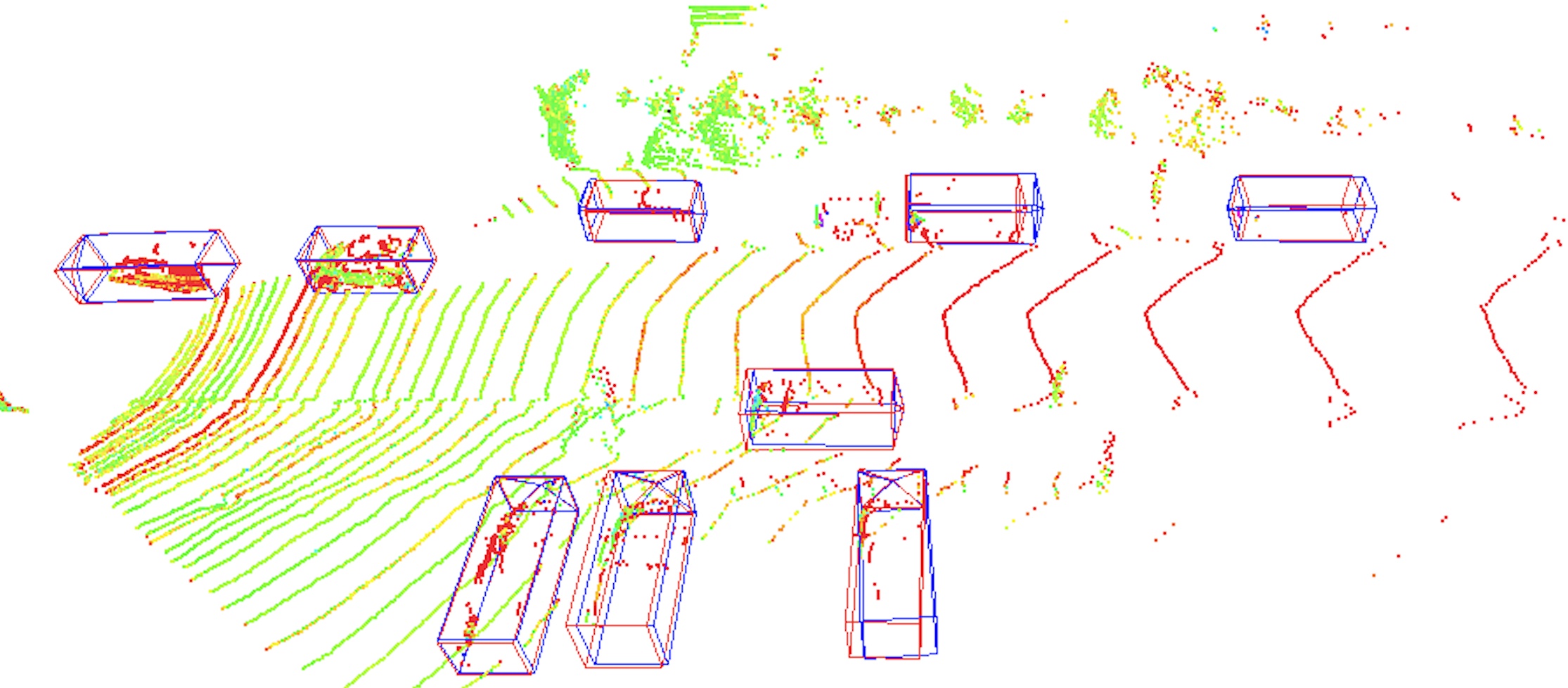}
    		\end{minipage}
    		\label{fig:compare_vis_8}
    	}\\
    	\centering
    	\vspace{-7pt}
    	\caption{\rev{Detection results comparison (top: 2-D, bottom: 3-D) at the boundary condition (i.e. horizontal or vertical rotation) between Smooth L1 loss-based (\text{left}) and the Gaussian-based (\text{right}) detectors. See illustration in Fig.~\ref{fig:gpd} for the Gaussian-based bounding box detection.}}
    	\label{fig:compare_vis}
    	\vspace{-12pt}
    \end{figure}
	
	\rev{\textit{Challenges in rotation regression loss design.} Although the dominant line of works~\cite{azimi2018towards, ding2018learning, yang2019scrdet, yang2021r3det} take a regression methodology to predict the rotation angle and have achieved state-of-the-art performance, the angle regression model suffer a few issues: i) the inconsistency between final detection metric e.g. mAP and loss function as not well addressed~\cite{rezatofighi2019generalized,zheng2020distance}, ii) boundary discontinuity~\cite{yang2019scrdet,yang2020arbitrary} when the loss jumps at boundary condition due to periodicity of angle and exchangeability of edges, and iii) square-like problem~\cite{yang2021dense} which refers to the case that the loss still sensitive to the rotation angle, when the objects are approximately in square shape under the long edge definition. See more details in Sec.~\ref{sec:revisit}. These issues remain open and there lacks a unified approach. In fact, they can largely hurt the final performance, especially at the boundary condition in the sense of horizontal or vertical rotation due to the periodicity of angles, as shown in Fig.~\ref{fig:compare_vis}. }
	
	\rev{\textit{Challenges in rotation regression loss implementation.}  To resolve the inconsistency between final detection metric which has been a pronounced challenge in literature, Skew Intersection over Union (SkewIoU) induced loss have been devised \cite{zheng2020rotation,zhou2019iou} which is unfortunately very hard-to-implement due to the need of handling the complicated corner cases of geometry overlapping\footnote{\rev{See an open-source version with thousands of lines of code for implementing the loss in \cite{zhou2019iou}: \url{https://github.com/open-mmlab/mmcv/pull/1854}. While our new loss only costs tens of lines of code.}}.}
    
     \rev{\textit{Challenges in rotation regression loss optimization.} Meanwhile, we argue that (and will be verified in our later technical analysis) for devising an effective rotation regression loss for high-precision rotation detection, the importance of different parameters of the bounding box (BBox) to different types of objects can vary. For example, the angle parameter ($\theta$) and the center point parameters ($x,y$) are important for high aspect ratio objects and small objects, respectively. In other words, the regression loss should be self-modulated during the learning process and calls for a more dynamic optimization strategy.}

\begin{figure}[!tb]
    	\centering
    	\subfigure{
    		\begin{minipage}[t]{0.46\linewidth}
    			\centering
    			\includegraphics[width=0.88\linewidth]{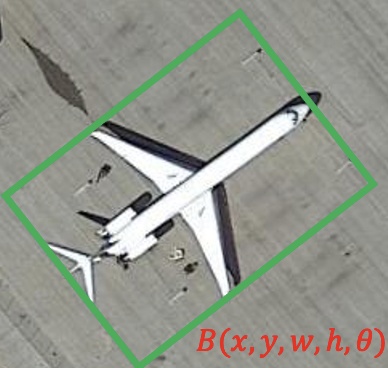}
    		\end{minipage}%
    		\label{fig:2-D_object}
    	}
    	\subfigure{
    		\begin{minipage}[t]{0.46\linewidth}
    			\centering
    			\includegraphics[width=0.88\linewidth]{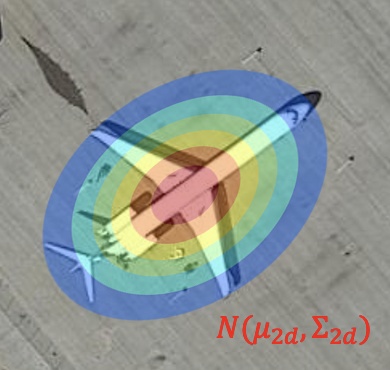}
    		\end{minipage}
    		\label{fig:2-D_gaussian}
    	}\\
    	\vspace{-4pt}
    	\subfigure{
    		\begin{minipage}[t]{0.47\linewidth}
    			\centering
    			\includegraphics[width=0.88\linewidth]{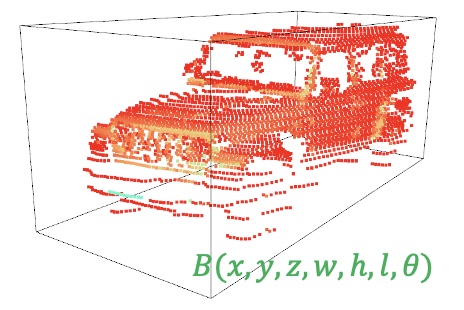}
    		\end{minipage}%
    		\label{fig:3-D_object}
    	}
    	\subfigure{
    		\begin{minipage}[t]{0.47\linewidth}
    			\centering
    			\includegraphics[width=0.88\linewidth]{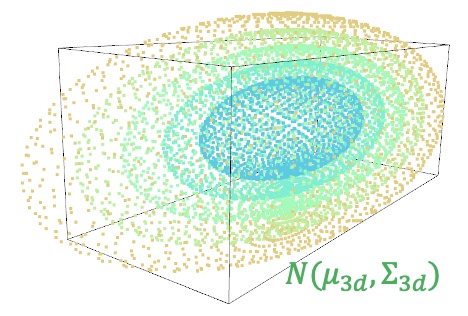}
    		\end{minipage}
    		\label{fig:3-D_gaussian}
    	}
    	\centering
    	\vspace{-10pt}
    	\caption{\rev{A schematic diagram of modeling a rotating bounding 2-D (top) and 3-D (bottom) box by a Gaussian distribution instead of the BBox.}}
    	\label{fig:gpd}
    	\vspace{-10pt}
    \end{figure}

	Seeing the above challenges, we propose to use a Gaussian distribution to model a rotated BBox for 2-D/3-D rotated object detection, as shown in Fig. \ref{fig:gpd}. \rev{Specifically, our method approximates the aforementioned metric/loss-consistent yet hard-to-implement SkewIoU loss \cite{zheng2020rotation,zhou2019iou} between two boxes by directly calculating their distance, which can be readily fulfilled (thanks to the Gaussian parameterization)} by popular metrics e.g. Gaussian Wasserstein Distance (GWD)~\cite{villani2008optimal}, Bhattacharyya Distance (BCD)~\cite{bhattacharyya1943measure} and Kullback-Leibler Divergence (KLD) \cite{kullback1951information}. \rev{Our Gaussian parameterization of BBox also makes our loss immune from both boundary discontinuity \cite{yang2019scrdet,yang2020arbitrary} and square-like problem \cite{yang2021dense} as shown on the right of Fig.~\ref{fig:compare_vis}.} 

	
	In particular, by analyzing the gradient of the parameters during learning, we show that the optimization of one parameter will be affected by the morphological parameters of the object (as the gradient weight). It means that the model will adaptively adjust the optimization strategy given a specific configuration of an object for detection, which can lead to excellent performance in high-precision detection. In addition, KLD and BCD are proven to be scale invariant, which is an important property that Smooth L1 loss and GWD do not possess. As the horizontal BBox is a special case of the rotated BBox, we show that KLD can also be degenerated into the $l_{n}$-norm loss as commonly used in horizontal detection pipeline.

	The preliminary content has partly appeared in the conference papers: \cite{yang2021rethinking} (GWD-based) and \cite{yang2021learning} (KLD-based)\footnote{\rev{This journal version extends the previous two conference versions in the following aspects:
	i) The Bhattacharyya Distance is employed and analyzed to further show the advantages of Gaussian modeling and the importance of scale invariance, see Sec. \ref{sec:bcd} and Sec. \ref{sec:ablation}; 
	ii) We extend the framework based on Gaussian distribution modeling from 2-D to 3-D object detection as specified in Sec. \ref{sec:3-D_loss}, Fig. \ref{fig:degeneration}-\ref{fig:3-Ddetect}, and Tab. \ref{tab:waymo_val}-\ref{tab:post_processing}; 
	iii) We propose a novel label assignment strategy based on Gaussian metric, combined with the dynamic threshold division by ATSS to further improve the performance, see Sec. \ref{sec:la} and Tab. \ref{tab:label_assign}.
	iv) We have added a more robust baseline, which eliminates the boundary discontinuity problem by predicting the two $\cos$ and $\sin$ components of the angle, see Sec. \ref{sec:loss} and Tab. \ref{tab:ablation_study}; 
	v) We verify our approach on additional more challenging datasets, including FDDB, DIOR-R, and DOTA-v1.5/v2.0, see Tab. \ref{tab:ablation_high_precision}, Tab. \ref{tab:ablation_study} and Tab. \ref{table:dior_r}. Among them DOTA-v1.5/v2.0 contain more samples and tiny objects (less than 10 pixels) than DOTA-v1.0;
	vi) We add comparative experiments between different approximate SkewIoU losses to demonstrate that the proposed technique is an easy-to-implement and better-performing alternative, as shown in Tab. \ref{tab:iou_related}.}}. 
	The contributions of this extended journal version are:
	
	\rev{1) We propose to model and detect general 2-D/3-D objects using a Gaussian distribution, which is in contrast to the commonly used BBox parameterization protocol regarding with the shape (and rotation) in existing object detection works. Our work is also beyond the a few works using Gaussian to model for the specific ellipse detection tasks e.g. lesion~\cite{LesionGaussian19} and knot~\cite{KnotsWACV21}. In 3-D case, the object heading can be estimated by our devised post-processing algorithm.}

	\rev{2) Our approach naturally address the bottlenecks in existing rotation detection methods: First, we can easily derive a new and easy-to-implement SkewIoU induced regression loss (compared to the plain SkewIoU loss~\cite{zhou2019iou}) regarding the distance between two Gaussians e.g. KLD, can well align the actual detection accuracy metric. Second, the boundary discontinuity and square-like problem, naturally disappear regardless how the rotated BBox is defined.}

	\rev{3) We apply three metrics between two Gaussian distributions to establish the regression loss: Gaussian Wasserstein Distance (GWD), Bhattacharyya Distance (BCD) and Kullback-Leibler Divergence (KLD), and perform experiments on twelve public datasets (2-D/3-D, aerial/text/face images) using three popular detectors (RetinaNet \cite{lin2017focal}, R$^3$Det \cite{yang2021r3det}, FPN \cite{lin2017feature}). The results show the effectiveness of our approach, especially for high-precision detection with high IoU (e.g. IoU = 0.75). Source code is made publicly available for both 2-D and 3-D cases (see experiment part).}

	\rev{4) By studying the BBox parameters' gradients with the  KLD loss, we show that the  regression  loss is self-modulated  during  the  learning  process and parameters are dynamically optimized. It further explains the effectiveness of our loss. }
	
	\rev{5) We propose to replace the IoU with Gaussian metric for label assignment to efficiently divide positive and negative samples for better training, so that the label assignment is consistent with the regression loss. We dynamically calculate the threshold via Adaptive Training Sample Selection (ATSS) \cite{zhang2020bridging}, and achieve further improvements.}

	\section{Related Work}\label{sec:related}
    We first review the works on rotated 2-D/3-D object detection, followed by the key challenges analysis in existing rotation detection methods. Readers are referred to \cite{LiuDetSurveyIJCV20} for comprehensive literature review on horizontal detection.
    
	As we will show later in the paper, \rev{our proposed regression loss is coherent to existing horizontal detection loss, in the sense that it degenerates to the popular $l_{n}$-norm loss when the rotation is horizontal.}
    

    \subsection{Rotated 2-D/3-D object detection}
    As an emerging direction, advance in this area try to extend classical horizontal detectors to the rotation case by adopting the rotated BBoxes. Compared with~\cite{yang2020arbitrary,yang2021dense} that treat the rotation detection tasks as an angle classification problem, regression based detectors are more dominant. For aerial images, ICN \cite{azimi2018towards}, ROI-Transformer \cite{ding2018learning}, SCRDet \cite{yang2019scrdet}, Gliding Vertex \cite{xu2020gliding} and ReDet \cite{han2021redet} are representative two-stage methods whose pipeline comprises of object localization and classification, while DRN \cite{pan2020dynamic} and RSDet \cite{qian2021learning} are single-stage methods. To pursue the trade-off of accuracy and speed, single-stage based refined detectors, such as R$^3$Det \cite{yang2021r3det} and S$^2$A-Net \cite{han2021align}, have been proposed.
    
    For scene text detection, RRPN \cite{ma2018arbitrary} employs rotated RPN to generate rotated proposals and further perform rotated BBox regression. TextBoxes++ \cite{liao2018textboxes++} adopts vertex regression on SSD \cite{liu2016ssd}. RRD \cite{liao2018rotation} improves TextBoxes++ by decoupling classification and BBox regression on rotation-invariant and rotation sensitive features, respectively. Most of the above detectors extend the $l_{n}$-norm loss from horizontal detector by adding extra parameters.
    
    3-D object detection tasks requires predicting rotated BBoxes in three dimensional space. 3-D object detectors can be classified as camera-based or LiDAR-based according to the sensing modality, and our work mainly focuses on the LiDAR-based methods. Many prior works of LiDAR-based 3-D detectors focus on the design of the feature encoding paradigm from raw points. For instance, PointRCNN~\cite{shi2019pointrcnn} uses PointNet++~\cite{qi2017pointnet++} to encode the per-point features and aggregate points' features by multi-scale set abstraction. VoxelNet~\cite{zhou2018voxelnet} partitions the 3-D spaces into rasterized voxels and uses PointNet~\cite{qi2017pointnet} to encode the voxel features from raw points inside each voxel and thus generates a unified feature representation of the 3-D space. SECOND~\cite{yan2018second} simplifies VoxelNet and implements computationally efficient sparse convolution operators for its feature encoder. PointPillars~\cite{lang2019pointpillars} partitions the space into a grid of pillars, resulting in single voxel per location in the bird-eye-view feature map, which improves backbone efficiency. Similar to 2-D detection, all those methods adopt the $l_{n}$-norms regression loss. 


    \rev{\subsection{Inconsistency between Metric and Rotation Loss} 
    It has been shown in classic horizontal detectors that the use of IoU induced loss e.g. GIoU \cite{rezatofighi2019generalized}, DIoU \cite{zheng2020distance} can ensure the consistency of the final detection metric and loss. However, the efforts~\cite{zheng2020rotation,zhou2019iou} of adapting such losses to rotation detection for differentiable learning is nontrivial because the calculation of SkewIoU needs to determine whether the BBox intersect, and how many intersection points, etc. incurring significant engineering as mentioned in Sec.~\ref{sec:introduction}.}



    
    \rev{Efforts have been made to gradient-friendly approximate SkewIoU loss. One representative work is PolarMask~\cite{xie2020polarmask}, whose calculation yet is discrete, which incurs numerical calculation error and the granularity of discretization can greatly affect the final calculation accuracy. PIoU~\cite{chen2020piou} is devised by simply counting the number of pixels. To tackle the uncertainty of convex caused by rotation, \cite{zheng2020rotation} proposes a projection operation to estimate the intersection area. SCRDet~\cite{yang2019scrdet} combines SkewIoU and Smooth L1 loss to develop an IoU-Smooth L1 loss, which partly circumvents the need for differentiable SkewIoU loss.  Polygon-to-Polygon distance loss \cite{yang2022polygon} is derived from the area sum of triangles specified by the vertexes of one polygon and the edges of the other. KFIoU \cite{yang2022kfiou} achieves a trend-level alignment with SkewIoU by Gaussian modeling and Kalman filtering.}
    
    \subsection{Boundary Discontinuity and Square-like Problems} 
    
\begin{figure}[!tb]
        \begin{center}
        \subfigure{
    		\begin{minipage}[t]{0.95\linewidth}
    			\centering
    			\includegraphics[width=0.98\linewidth]{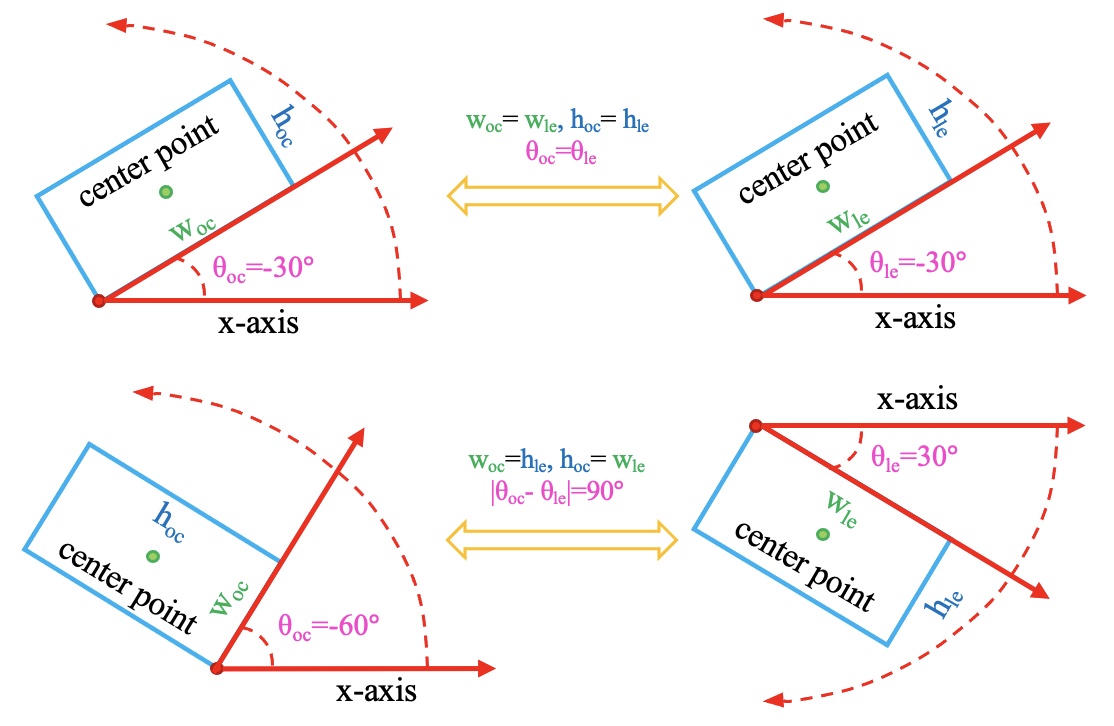}
    		\end{minipage}%
    	}\centering\vspace{-7pt}
        \caption{Two classic definitions of rotated BBoxes. \text{Left:} OpenCV Definition $D_{oc}$ \cite{yang2019scrdet,yang2021r3det}, \text{Right:} Long Edge Definition $D_{le}$ \cite{ding2018learning,han2021redet}.}
        \label{fig:definition}
        \end{center}
    \end{figure}

    \rev{In general, due to the periodicity of angle parameters and the fundamental limitation of the classic BBox definitions (see Fig.~\ref{fig:definition}), regression-based rotation detectors often suffer from the so-called boundary discontinuity and square-like problem. The first problem refers to the loss jump at the boundary condition, while the latter leads to the sensitivity of loss to the rotation change when the object is approximately square. As will be detailed in Sec.~\ref{sec:revisit}, these two issues in depend on the choice of BBox definition.}

    Existing methods try to solve part of these problems by different means. For instance, SCRDet~\cite{yang2019scrdet} and RSDet~\cite{qian2021learning} propose IoU-Smooth L1 loss and the so-called modulated loss to smooth the boundary loss jump. CSL~\cite{yang2020arbitrary,yang2022on} transforms angular prediction from a regression task to a classification one. DCL~\cite{yang2021dense} further solves square-like object detection problem under the long edge definition. Instance segmentation-based methods are practical, and relevant methods (e.g. Mask OBB~\cite{wang2019mask}) have been proposed. However, there still exist limitations. First, using rotated boxes as binary masks will introduce background area, which will reduce the classification accuracy of pixels and affect the accuracy of the final prediction box. Secondly, for the top-down methods (e.g. Mask RCNN~\cite{he2017mask}), dense scenes will limit the detection of horizontal boxes because of the excessive suppression of dense horizontal overlapping BBoxes due to non-maximum suppression (NMS), thereby affecting subsequent segmentation. Aerial images often show large scenes with a large number of dense and small objects, which is not suitable for the bottom-up methods, such as SOLO \cite{wang2020solo} and CondInst \cite{tian2020conditional}, which assign different instances to different channels. This is the main reason why regression-based rotation detection algorithms still dominate in the field of aerial imagery.


    
    \begin{figure*}[!tb]
    	\centering
    	\subfigure{
    		\begin{minipage}[t]{0.23\linewidth}
    			\centering
    			\includegraphics[width=1.\linewidth]{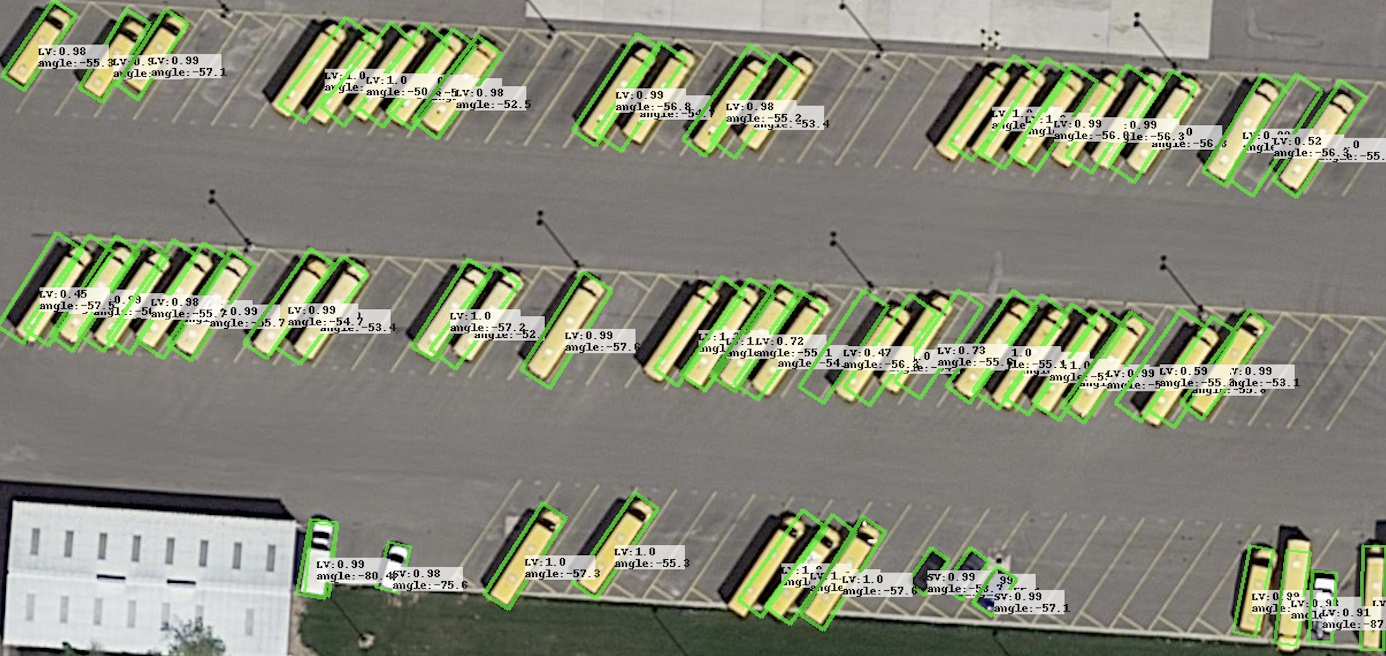}
    		\end{minipage}
    		\label{fig:retinanet_dota_vis}
    	}
    	\subfigure{
    		\begin{minipage}[t]{0.23\linewidth}
    			\centering
    			\includegraphics[width=1.\linewidth]{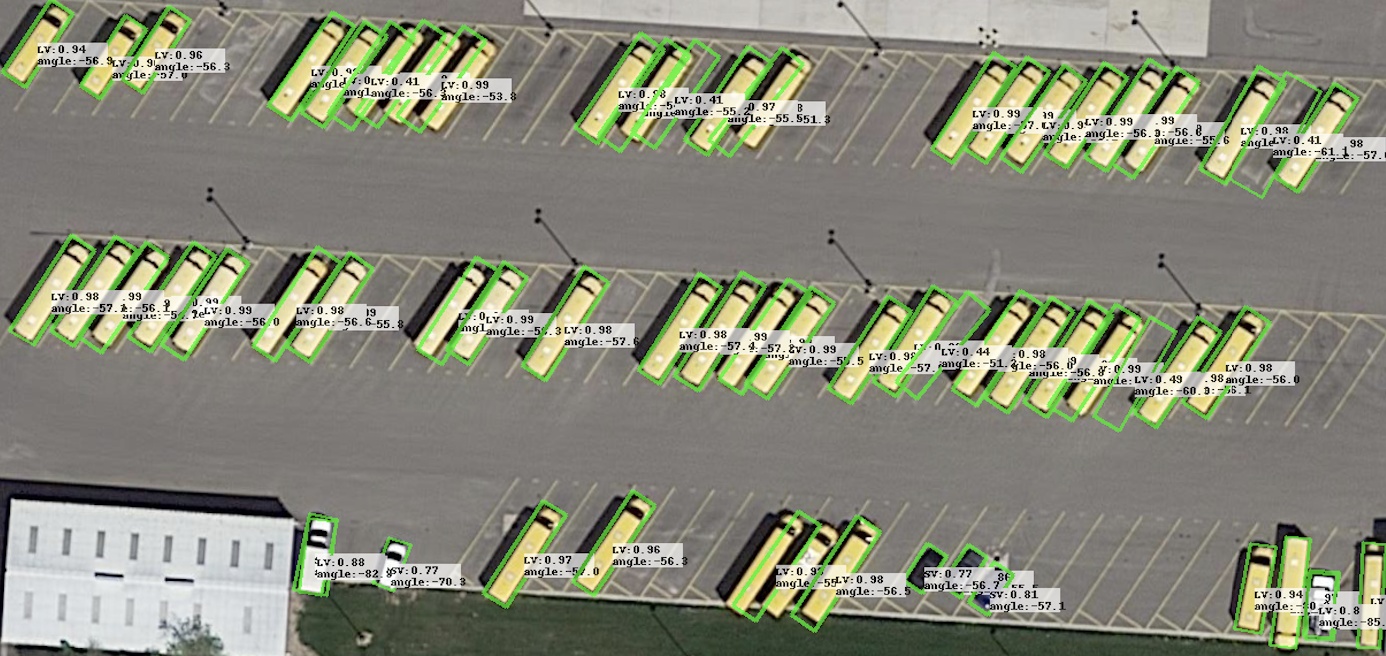}
    		\end{minipage}
    		\label{fig:gwd_dota_vis}
    	}
    	\subfigure{
    		\begin{minipage}[t]{0.23\linewidth}
    			\centering
    			\includegraphics[width=1.\linewidth]{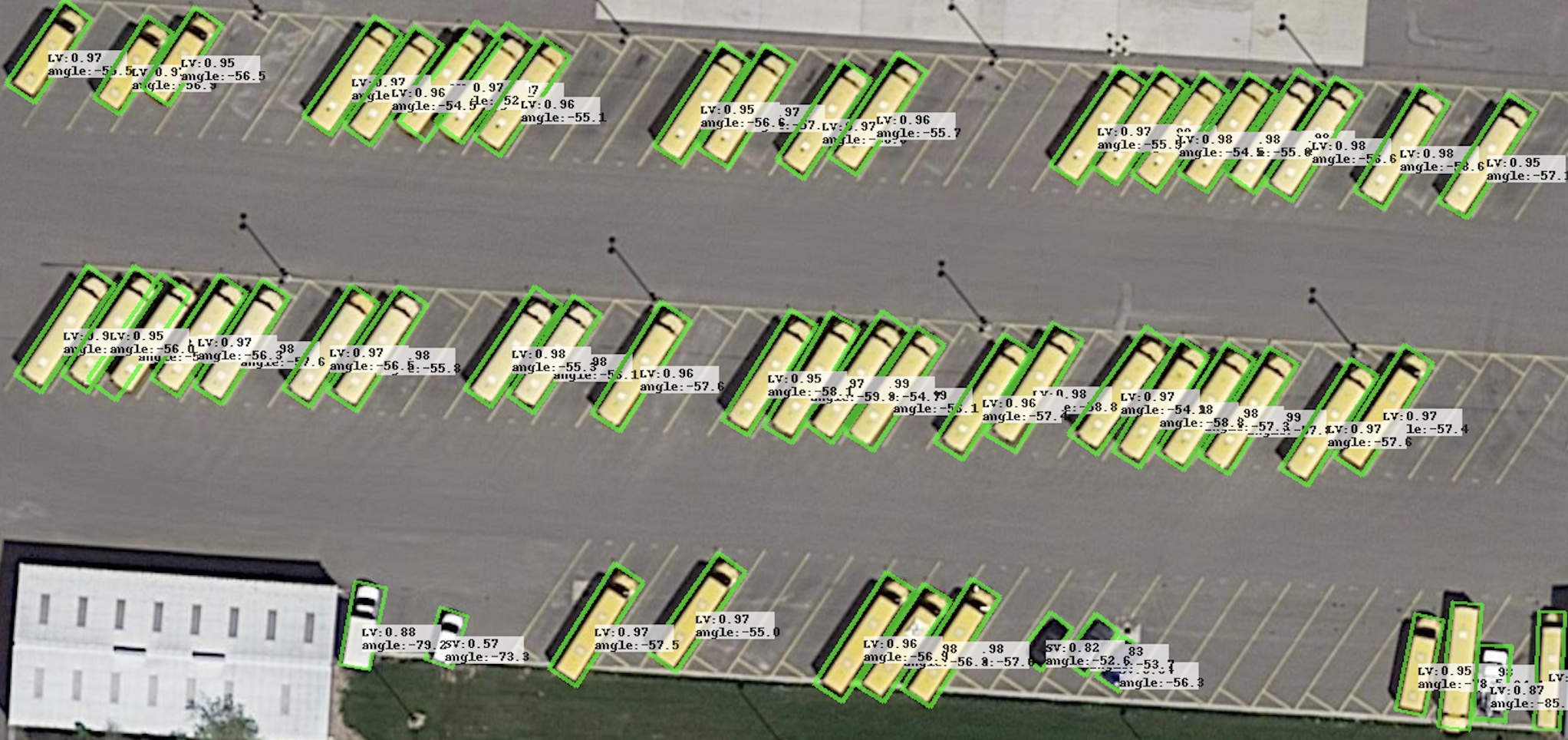}
    		\end{minipage}
    		\label{fig:bcd_dota_vis}
    	}
    	\subfigure{
    		\begin{minipage}[t]{0.23\linewidth}
    			\centering
    			\includegraphics[width=1.\linewidth]{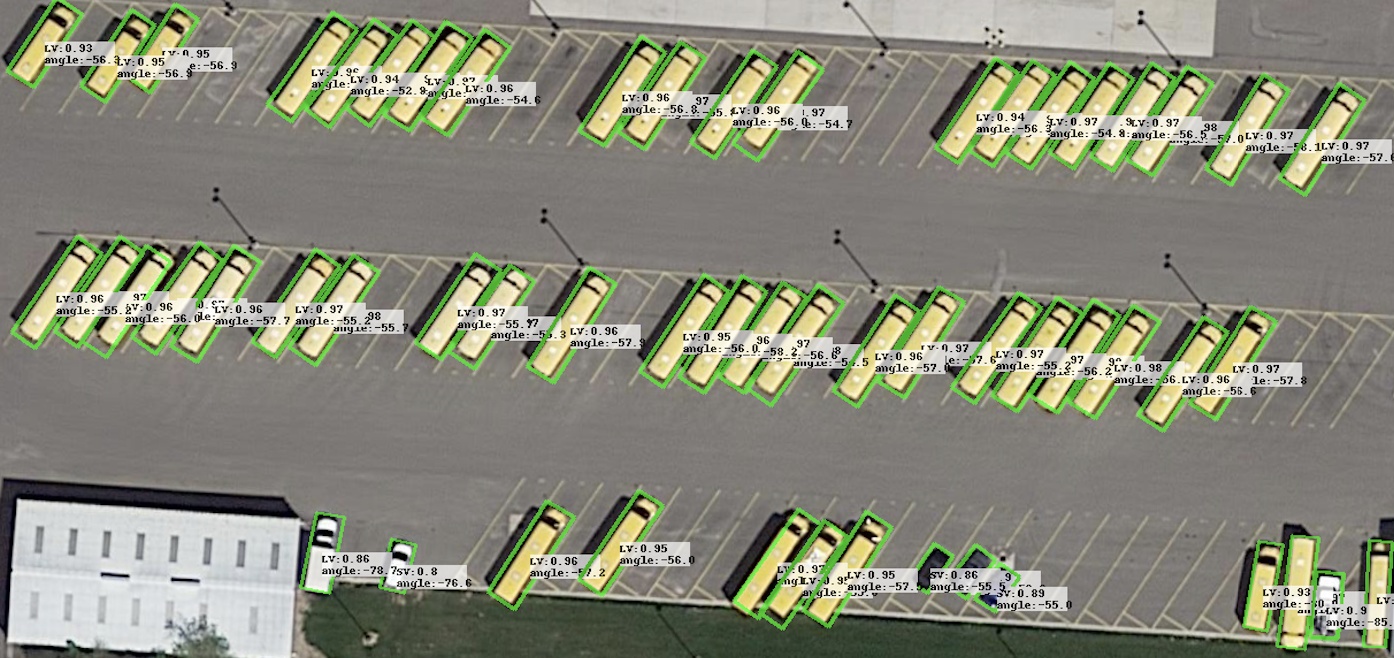}
    		\end{minipage}
    		\label{fig:kld_dota_vis}
    	}\\
    	\vspace{-5pt}
    	\subfigure{
    		\begin{minipage}[t]{0.23\linewidth}
    			\centering
    			\includegraphics[width=1.\linewidth]{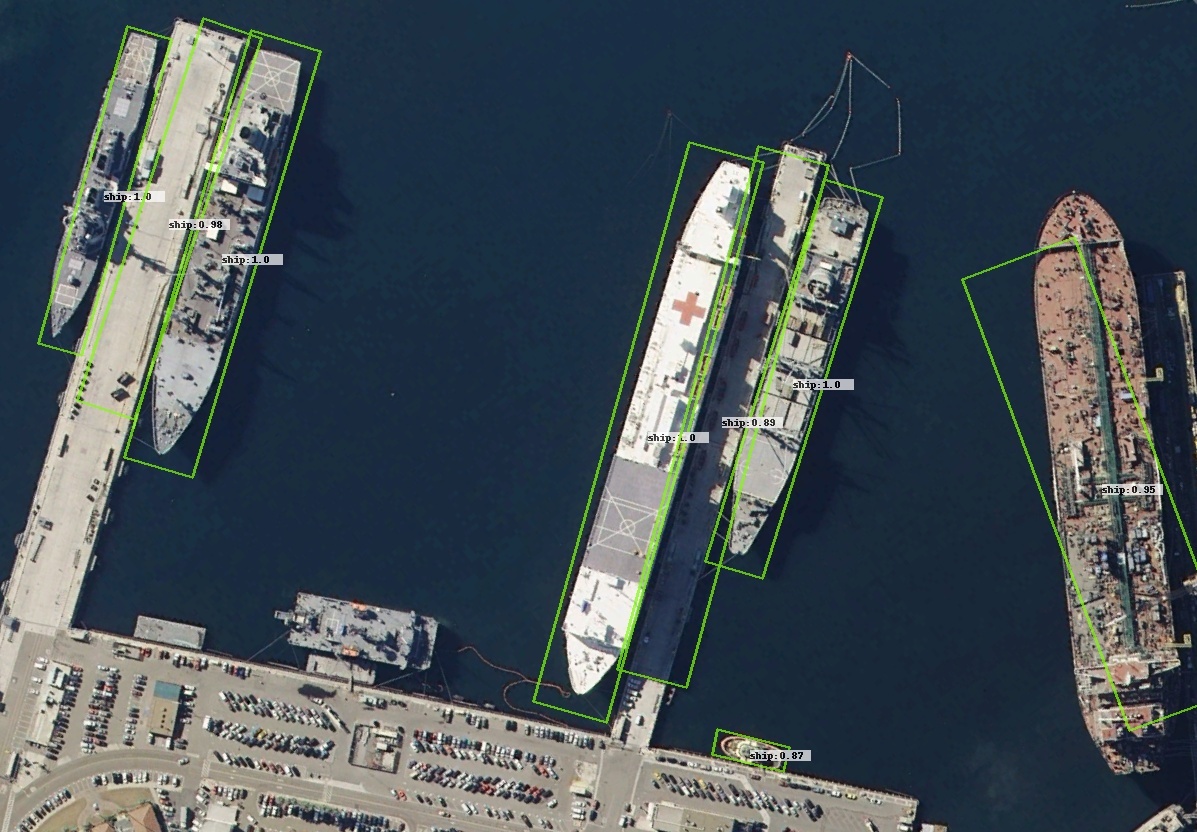}
    		\end{minipage}
    		\label{fig:retinanet_hrsc2016_vis}
    	}
    	\subfigure{
    		\begin{minipage}[t]{0.23\linewidth}
    			\centering
    			\includegraphics[width=1.\linewidth]{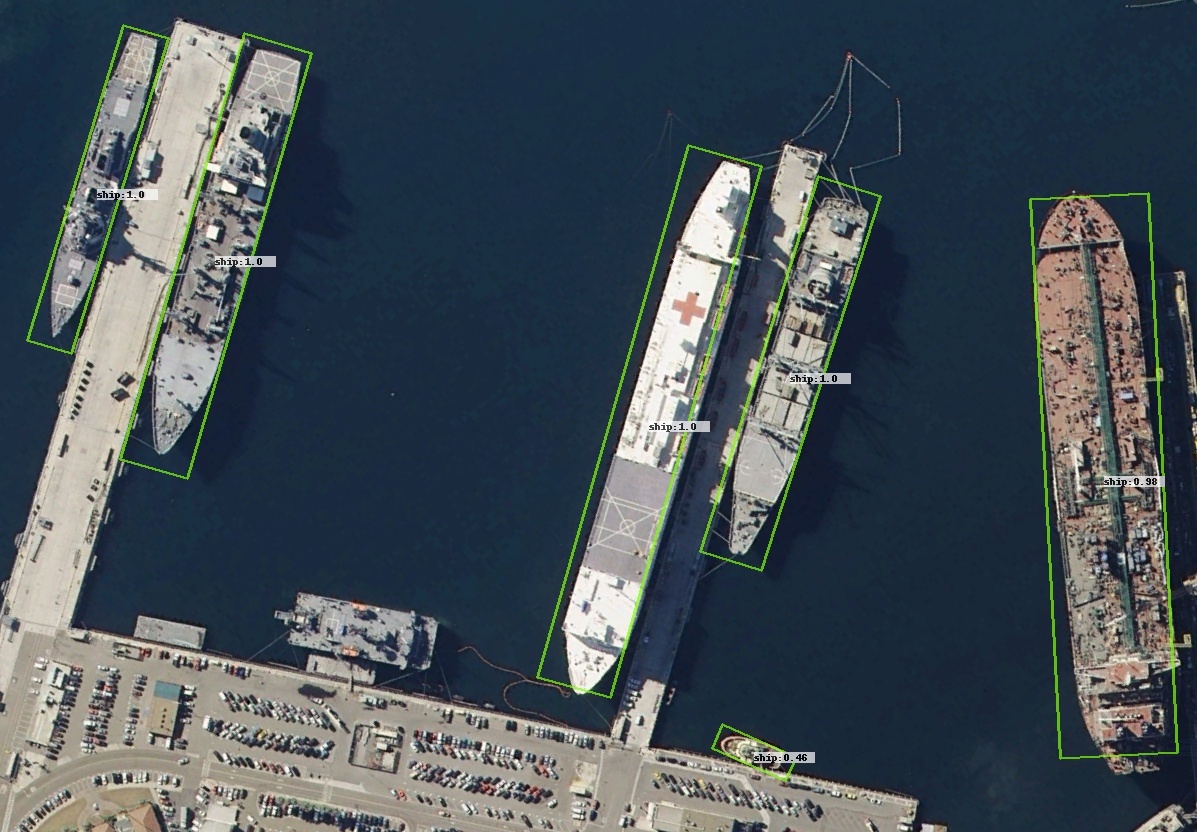}
    		\end{minipage}
    		\label{fig:gwd_hrsc2016_vis}
    	}
    	\subfigure{
    		\begin{minipage}[t]{0.23\linewidth}
    			\centering
    			\includegraphics[width=1.\linewidth]{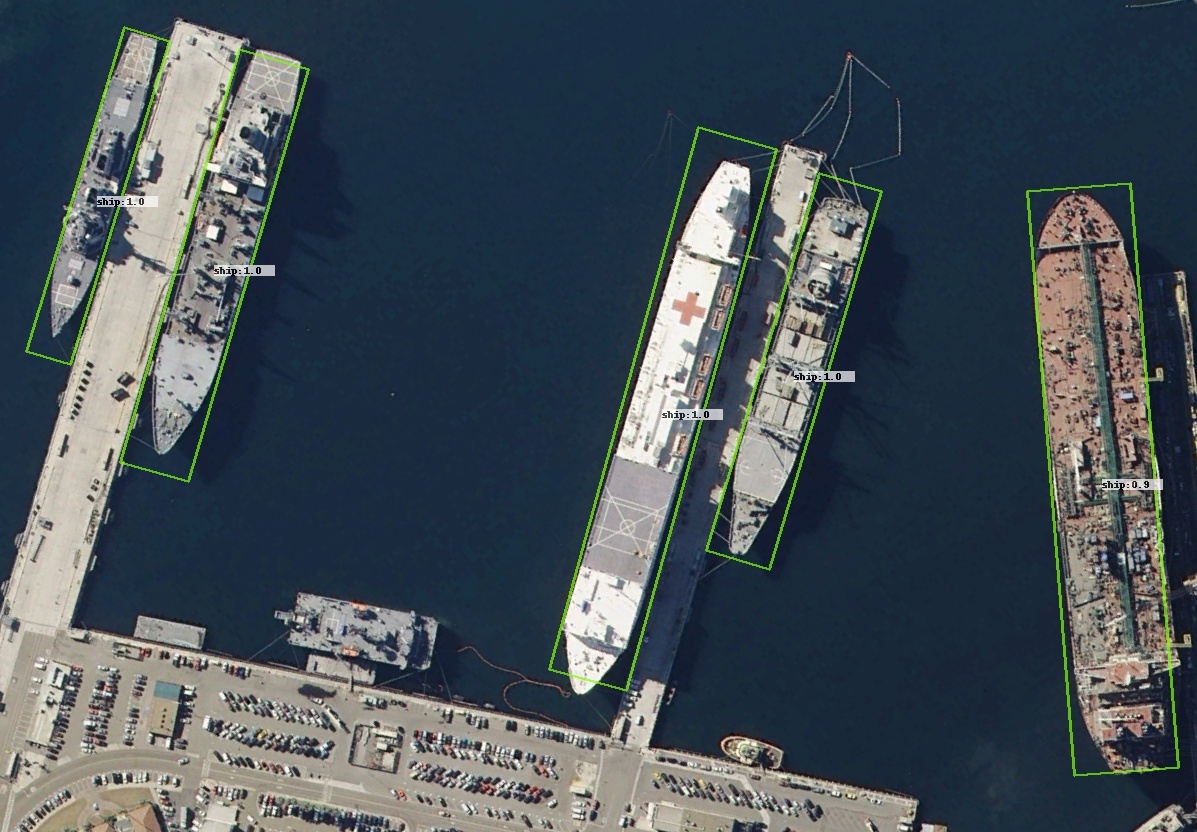}
    		\end{minipage}
    		\label{fig:bcd_hrsc2016_vis}
    	}
    	\subfigure{
    		\begin{minipage}[t]{0.23\linewidth}
    			\centering
    			\includegraphics[width=1.\linewidth]{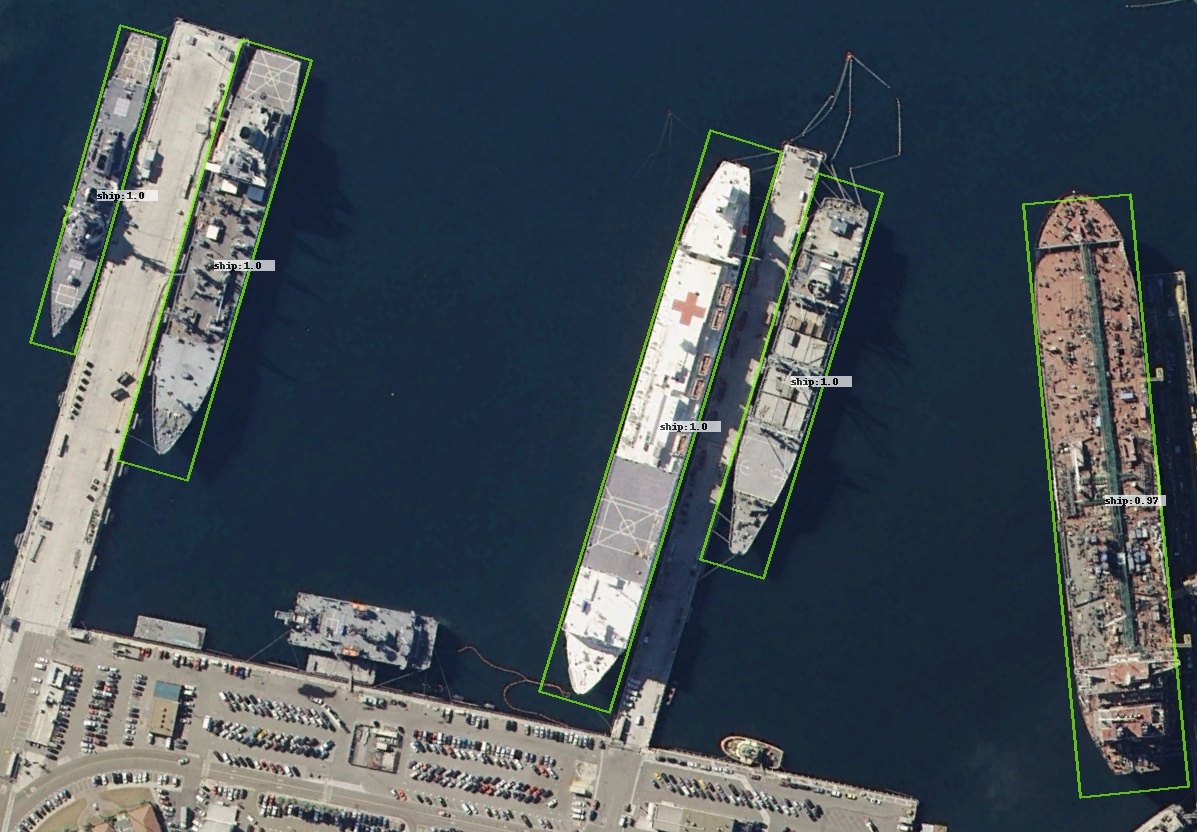}
    		\end{minipage}
    		\label{fig:kld_hrsc2016_vis}
    	}\\
    	\vspace{-5pt}
    	\subfigure{
    		\begin{minipage}[t]{0.23\linewidth}
    			\centering
    			\includegraphics[width=1.\linewidth]{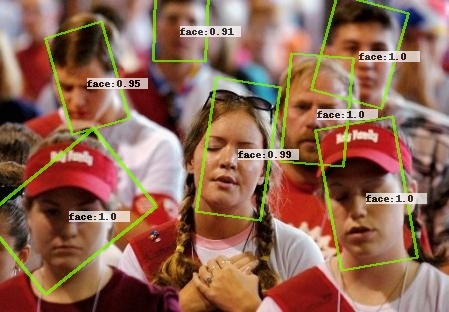}
    		\end{minipage}
    		\label{fig:retinanet_fddb_P2170}
    	}
    	\subfigure{
    		\begin{minipage}[t]{0.23\linewidth}
    			\centering
    			\includegraphics[width=1.\linewidth]{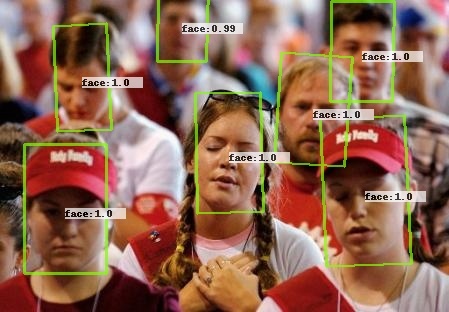}
    		\end{minipage}
    		\label{fig:gwd_fddb_P2170}
    	}
    	\subfigure{
    		\begin{minipage}[t]{0.23\linewidth}
    			\centering
    			\includegraphics[width=1.\linewidth]{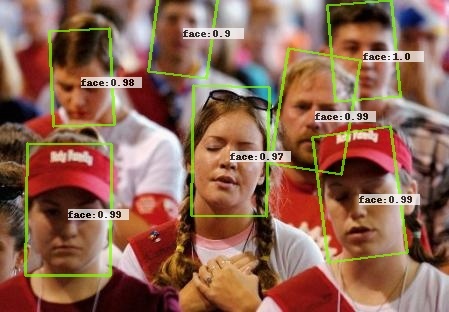}
    		\end{minipage}
    		\label{fig:bcd_fddb_P2170}
    	}
    	\subfigure{
    		\begin{minipage}[t]{0.23\linewidth}
    			\centering
    			\includegraphics[width=1.\linewidth]{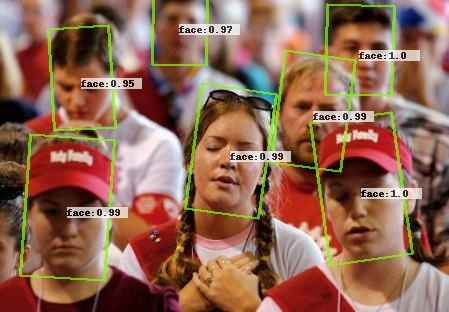}
    		\end{minipage}
    		\label{fig:kld_fddb_P2170}
    	}
    	\centering\vspace{-7pt}
    	\caption{High-precision detection by Smooth L1 loss, GWD, BCD and KLD (left to right). Datasets: DOTA (top) \cite{xia2018dota}, HRSC2016 (bottom) \cite{liu2017high} and FDDB \cite{jain2010fddb}. Since the center point parameters in Smooth L1 Loss and GWD are independently optimized, their prediction results are slightly shifted. In contrast, the KLD-based prediction results are closer to the object boundary and show strong robustness in dense scenes. Similarly, the prediction angle of Smooth L1 Loss is not as accurate as KLD. 
    	}
    	\label{fig:high_precision_compare_vis}
    \end{figure*}
    
    \section{Rotated Regression Detector Revisit}\label{sec:revisit}
    To motivate this work, in this section, we introduce and analyze some deficiencies in state-of-the-art rotating detectors, which are mostly based on angle regression.
    
    \subsection{Bounding Box Definition Specific Detector Design}
    Existing rotated object detection are mainly developed by the adaption of horizontal detectors, with classic regression loss~\cite{ding2018learning, yang2019scrdet, yang2022scrdet++, yang2021r3det}, e.g. $l_{n}$-norms, defined on the parameterization of two BBoxes in 2-D/3-D to maximize the IoU. 
    
    Fig.~\ref{fig:definition} shows two popular definitions for parameterizing rotated BBox: OpenCV protocol denoted by $D_{oc}$ \cite{yang2019scrdet,yang2021r3det}, and long edge definition denoted by $D_{le}$ \cite{ding2018learning,han2021redet}. Note $\theta \in [-90^\circ, 0^\circ)$ of the former denotes the acute or right angle between $w_{oc}$ of BBox and $x$-axis. While $\theta \in [-90^\circ, 90^\circ)$ of the latter definition is the angle of BBox's long edge $w_{le}$ and $x$-axis. The two definitions are convertible to each other:
    \begin{equation*}
    \footnotesize
        D_{le}(w_{le},h_{le},\theta_{le}) = \\
    	\left\{ \begin{array}{rcl}
    	D_{oc}(w_{oc},h_{oc},\theta_{oc}), & w_{oc} \geq h_{oc} \\ D_{oc}(h_{oc},w_{oc},\theta_{oc} + 90^\circ), & otherwise
    	\end{array}\right.
    	\label{eq:oc2le}
    \end{equation*}
    \begin{equation*}
    \footnotesize
        D_{oc}(w_{oc},h_{oc},\theta_{oc}) = \\
    	\left\{ \begin{array}{rcl}
    	D_{le}(w_{le},h_{le},\theta_{le}), & \theta_{le} \in [-90^\circ,0^\circ) \\ D_{le}(h_{le},w_{le},\theta_{le} - 90^\circ), & otherwise
    	\end{array}\right.
    	\label{eq:le2oc}
    \end{equation*}
    
    \rev{Unfortunately, regardless the BBox definition, there always exist specific issues to solve, which is coupled with the definition choice (e.g. SCRDet~\cite{yang2019scrdet}, R$^3$Det~\cite{yang2021r3det} using OpenCV protocol $D_{oc}$ while CSL~\cite{yang2020arbitrary}, DCL~\cite{yang2021dense} adopting $D_{le}$). Specifically, both definitions would raise the boundary discontinuity issue with a common reason of periodicity of angle (PoA)~\cite{yang2020arbitrary}, while the OpenCV protocol suffers an additional cause of the exchangeability of edges (EoE)~\cite{yang2020arbitrary}. While the long edge definition meanwhile incurs the aforementioned square-like problem but not for OpenCV protocol.}
    
    
    Moreover, the effects of different definitions can be entangled with other factors including the network, datasets, and hyperparameters. There lacks an elegant solution to decouple the detector design from the definition choice.
    
    \begin{figure*}[!tb]
    	\centering
    	\subfigure[Angle difference case]{
    		\begin{minipage}[t]{0.35\linewidth}
    			\centering
    			\includegraphics[width=0.98\linewidth]{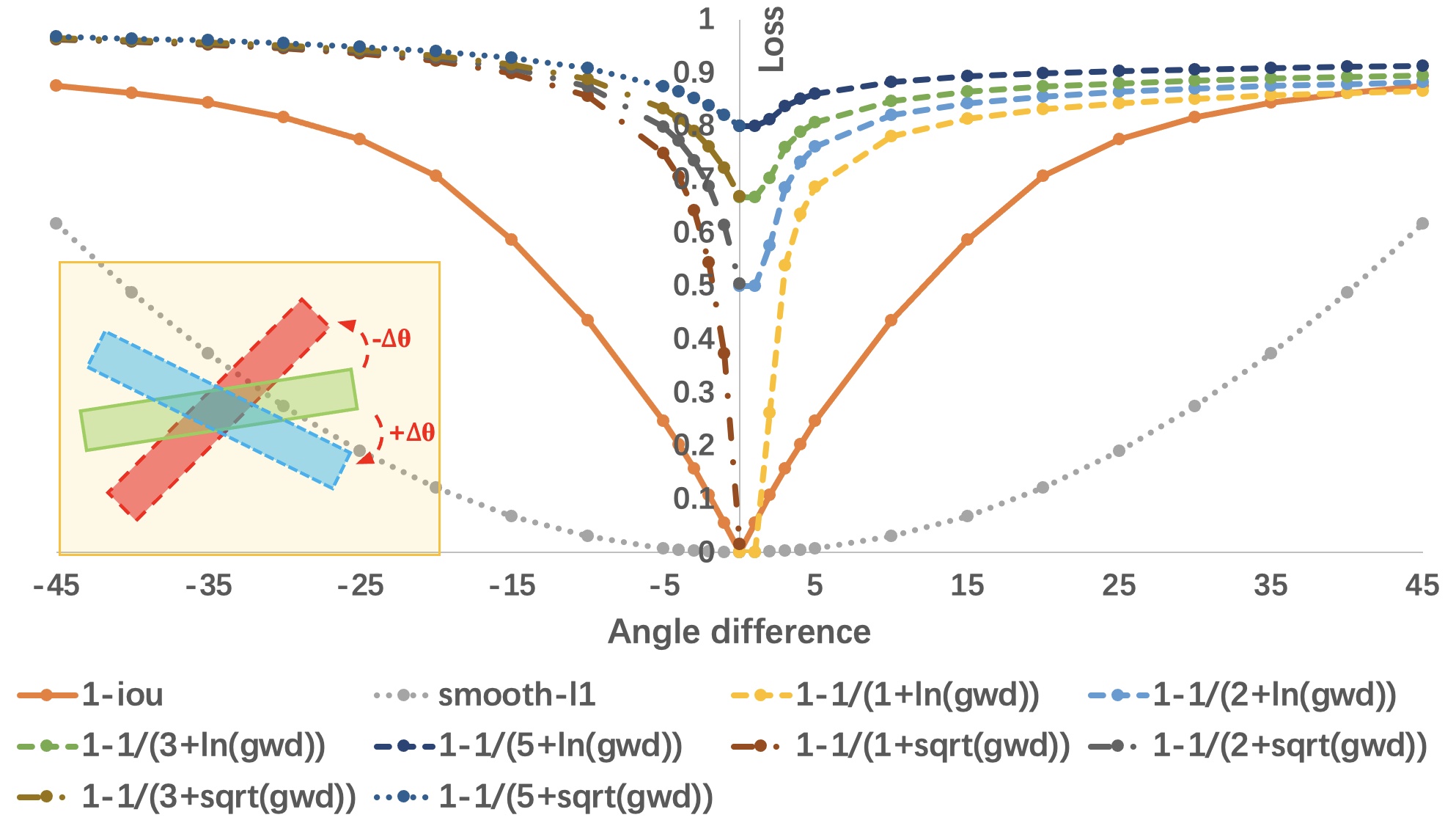}
    		\end{minipage}%
    		\label{fig:iou_smooth-l1-1}
    	}
    	\subfigure[Aspect ratio case]{
    		\begin{minipage}[t]{0.29\linewidth}
    			\centering
    			\includegraphics[width=0.98\linewidth]{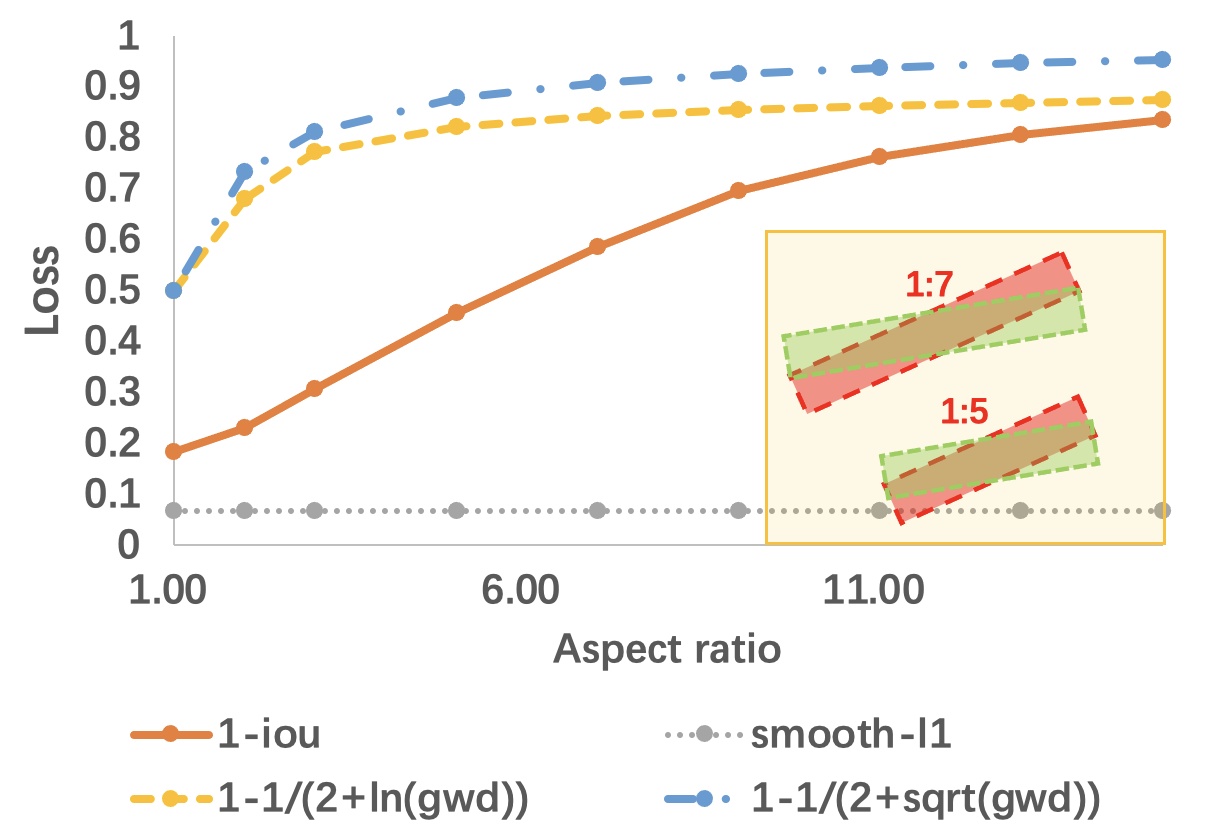}
    		\end{minipage}%
    		\label{fig:iou_smooth-l1-2}
    	}
    	\subfigure[Center shifting case]{
    		\begin{minipage}[t]{0.29\linewidth}
    			\centering
    			\includegraphics[width=0.98\linewidth]{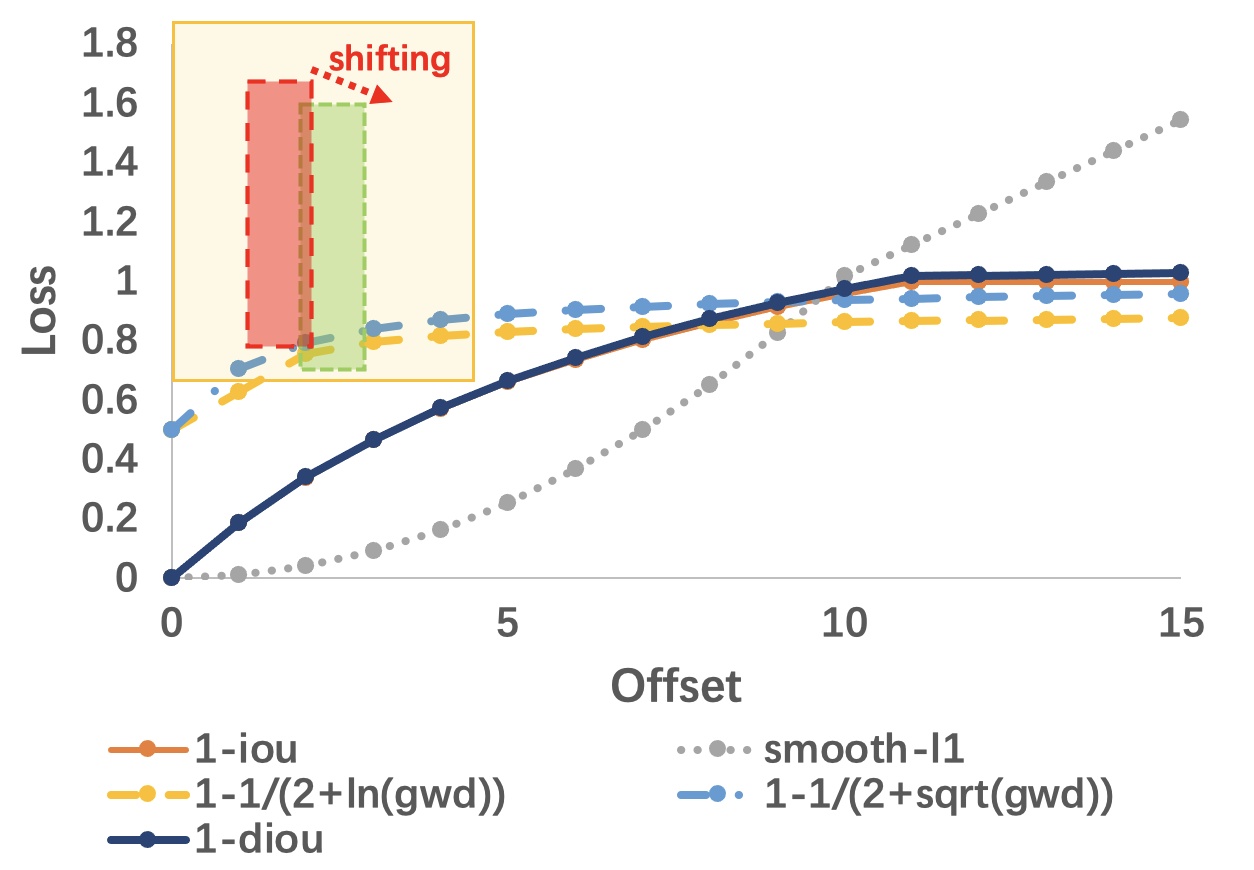}
    		\end{minipage}
    		\label{fig:iou_smooth-l1-3}
    	}
    \centering\vspace{-7pt}
    	\caption{Behavior comparison of different losses: Smooth L1 Loss, IoU Loss and Gaussian-based loss (e.g. GWD) in different detection cases.}
    	\label{fig:iou_smooth-l1}
    \end{figure*}
    
    \subsection{Regression Loss Design Revisit: From Horizon to Rotation Detection}\label{sec:simple_reg_loss}
    Regression loss is widely used in visual object detectors. For horizontal BBox regression, the model \cite{ren2015faster,lin2017feature,lin2017focal} mainly outputs four items for location and size:
    \begin{equation}
    \small
    	\begin{aligned}
    	t_{x}^{p}=\frac{x_{p}-x_{a}}{w_{a}}, t_{y}^{p}=\frac{y_{p}-y_{a}}{h_{a}}, t_{w}^{p}=\ln\left(\frac{w_{p}}{w_{a}}\right), t_{h}^{p}=\ln\left(\frac{h_{p}}{h_{a}}\right)
    	\label{eq:reg_pred}
    	\end{aligned}
    \end{equation}
    to match the four targets from the ground truth:
    \begin{equation}
    \small
    	\begin{aligned}
    	t_{x}^{t}=\frac{x_{t}-x_{a}}{w_{a}}, t_{y}^{t}=\frac{y_{t}-y_{a}}{h_{a}}, t_{w}^{t}=\ln\left(\frac{w_{t}}{w_{a}}\right), t_{h}^{t}=\ln\left(\frac{h_{t}}{h_{a}}\right)
    	\label{eq:reg_target}
    	\end{aligned}
    \end{equation}
    where $x,y,w,h$ denote the center coordinates, width and height, respectively. $x_{t}, x_{a}, x_{p}$ are for the ground-truth box, anchor box, and predicted box, respectively (so for $y,w,h$).
    
    Extending the above horizontal case, existing rotation detection models~\cite{yang2018automatic,ding2018learning,han2021align,pan2020dynamic,ming2021dynamic} also use regression loss which simply involves an extra angle parameter $\theta$:
    \begin{equation}
    \small
    	\begin{aligned}
    	t_{\theta}^{p}=f(\theta_{p}-\theta_{a}), t_{\theta}^{t}=f(\theta_{t}-\theta_{a})
    	\label{eq:reg_theta}
    	\end{aligned}
    \end{equation}
    where $f(\cdot)$ is used to deal with angular periodicity, such as trigonometric functions, modulo, etc.
    
    The overall regression loss for rotation detection is:
    \begin{equation}
    \small
    	\begin{aligned}
    	L_{reg} = l_{n}\text{-norm}\left(\Delta t_{x}, \Delta t_{y}, \Delta t_{w}, \Delta t_{h}, \Delta t_{\theta}\right)
    	\label{eq:reg_loss}
    	\end{aligned}
    \end{equation}
    where $\Delta t_{x} = t_{x}^{p} - t_{x}^{t} = \frac{\Delta x}{w_{a}}$, $\Delta t_{y} = t_{y}^{p} - t_{y}^{t} = \frac{\Delta y}{h_{a}}$, $\Delta t_{w} = t_{w}^{p} - t_{w}^{t} = \ln(w_{p}/w_{t})$, $\Delta t_{h} = t_{h}^{p} - t_{h}^{t} = \ln(h_{p}/h_{t})$, and $\Delta t_{\theta} = t_{\theta}^{p} - t_{\theta}^{t}$.
    
    It can be seen that $l_n$-norm focuses on the difference of individual BBox parameters and the parameters will not be dynamically optimized according to the shape of object, making the loss (or detection accuracy) sensitive to the under-fitting of any of the parameters. 
    This mechanism is fatal to high-precision detection. Taking the left side of Fig. \ref{fig:high_precision_compare_vis} as an example, the detection result based on the Smooth L1 loss often shows the deviation of the center point or angle. Moreover, different types of objects have different sensitivity to these five parameters. For example, the angle parameter is very important for detecting objects with large aspect ratios. This requires to select an appropriate set of weights given a specific single object sample during the training, which is nontrivial or even unrealistic. Thus, regression loss should be self-modulated during the learning process and calls for a more dynamic optimization strategy. This provides inspiration for the design of regression loss in this paper.

    \subsection{Major Challenges in Rotation Detection}
    \subsubsection{Inconsistency between Metric and Loss}
    Intersection over Union (IoU) has been the standard metric for both horizontal detection and rotation detection. However, there is an inconsistency between the metric and regression loss (e.g. $l_{n}$-norms), that is, a smaller training loss cannot guarantee a higher performance, which has been extensively discussed in horizontal detection \cite{rezatofighi2019generalized, zheng2020distance}. This misalignment becomes more prominent in rotating object detection due to the introduction of angle parameter in regression based models. To illustrate this, we use Fig.~\ref{fig:iou_smooth-l1} to compare IoU induced loss and Smooth L1 loss \cite{ren2015faster}:
    
    \textit{Case 1:} Fig.~\ref{fig:iou_smooth-l1-1} depicts the relation between angle difference and loss functions. Though they all bear monotonicity, only Smooth L1 curve is convex while the others are not.
    	
    \textit{Case 2:} Fig.~\ref{fig:iou_smooth-l1-2} shows the changes of the two loss functions under different aspect ratio conditions. It can be seen that the Smooth L1 loss of the two BBoxe are constant (mainly from the angle difference), but the IoU loss will change drastically as the aspect ratio varies.
    
    \textit{Case 3:} Fig.~\ref{fig:iou_smooth-l1-3} explores the impact of center point shifting on different loss functions. Similarly, despite the same monotonicity, there is no high degree of consistency.
    
    Seeing the above flaws of classic Smooth L1 loss, IoU-induced losses emerge for horizontal detection e.g. GIoU \cite{rezatofighi2019generalized}, DIoU \cite{zheng2020distance}. It could help to fill the gap between metric and regression loss for rotation detection. Thanks to the introduction of Gaussian distribution for object modeling, the between-distribution metric becomes a simple yet effective approximate SkewIoU loss. Moreover, we will show later that the Gaussian framework has unique properties to solve boundary discontinuity and square-like problem. 
    

    \begin{figure*}[!tb]
    	\centering
    	\includegraphics[width=0.95\linewidth]{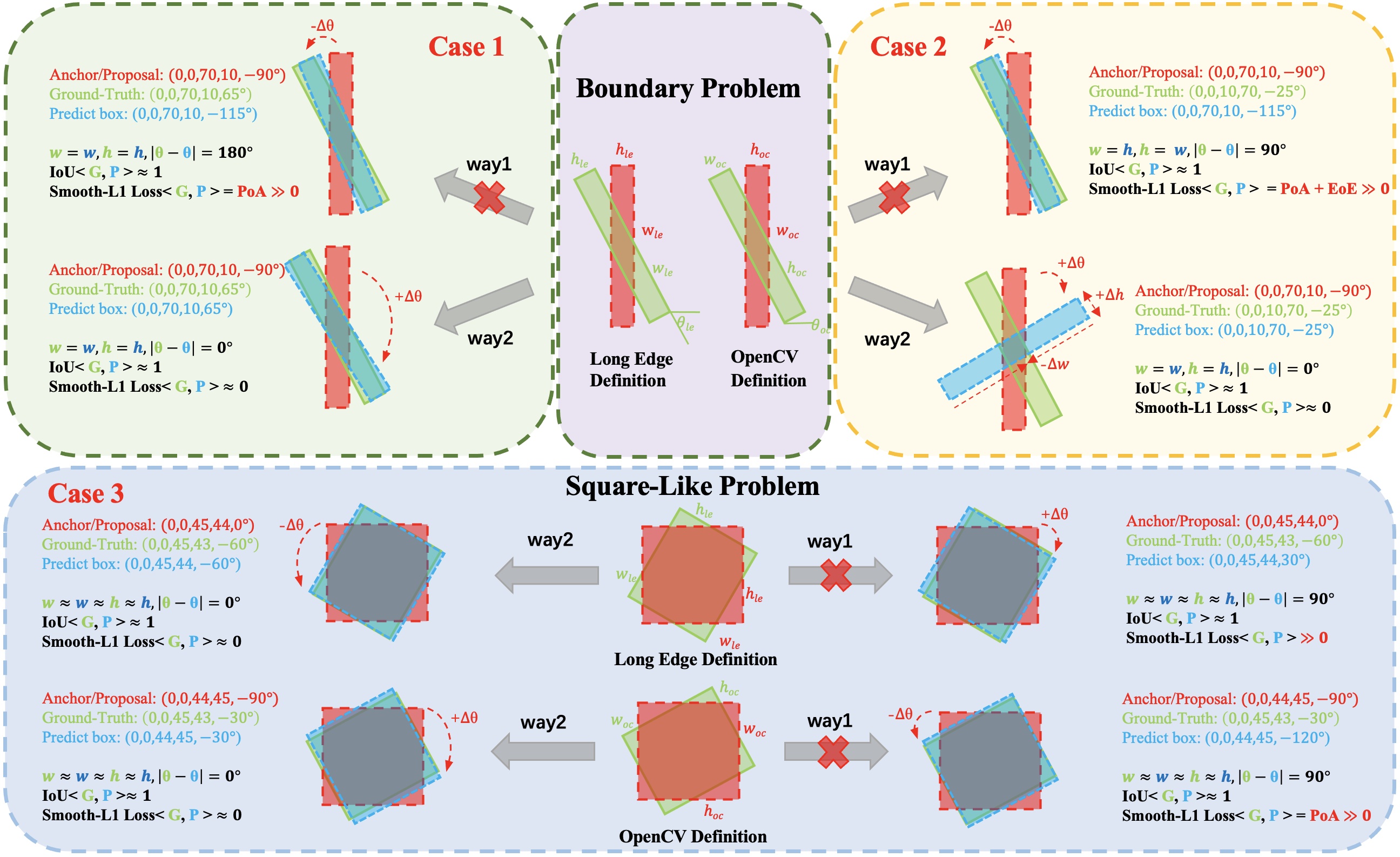}
    \centering\vspace{-7pt}
    	\caption{Boundary discontinuity under two BBox definitions (\text{top}), and illustration of the square-like problem (\text{bottom}).}
    	\label{fig:problems}
    \end{figure*}
    
    \subsubsection{Boundary Discontinuity}
    As a standing issue for regression-based rotation detectors, the boundary discontinuity ~\cite{yang2019scrdet,yang2020arbitrary} in general refers to the sharp loss increase at the boundary induced by the angle and edge parameterization. 
    
    Specifically, \textit{Case 1-2} in Fig.~\ref{fig:problems} describe the boundary discontinuity. Take \textit{Case 2} as an example, we assume that there is a red anchor/proposal $\color{black}{(0,0,70,10,-90^\circ)}$ and a green ground truth (GT) $\color{black}{(0,0,10,70,-25^\circ)}$ at the boundary position\footnote{The angle of the BBox is close to the maximum and minimum values of the angle range. For more clearly visualization, the GT has been rendered with a larger angle in Fig.~\ref{fig:problems}.}, both of which are defined in OpenCV definition $D_{oc}$. 
    The upper right corner of Fig.~\ref{fig:problems} shows two ways to regress from anchor/proposal to GT. The \text{way1} achieves the goal by only rotating anchor/proposal by an angle counterclockwise, but a very large Smooth L1 loss occurs at this time due to the periodicity of angle (PoA) and the exchangeability of edges (EoE). As discussed in CSL \cite{yang2020arbitrary}, this is because the result of the blue prediction box $\color{black}{(0,0,70,10,-115^\circ)}$ is outside the defined range. As a result, the model has to make predictions in other complex regression forms, such as rotating anchor/proposal by an large angle clockwise to the blue box while scaling $w$ and $h$ (\text{way2} in \textit{Case 2}). A similar problem (only PoA) also occurs in the long edge definition $D_{le}$, as shown in \textit{Case 1}. 
    
    When the predefined anchor/proposal and GT are not in the boundary position, \text{way1} will not produce a large loss.
    Therefore, there exists inconsistency between the boundary position and the non-boundary position regression, which makes the model very confused about in which way it should perform regression. Since non-boundary cases account for the majority, the regression results of models, especially those with weaker learning capacity, are fragile in boundary cases, as shown in the left of Fig.~\ref{fig:compare_vis}.
    
    \subsubsection{Square-Like Problem}
    In addition, there is also a square-like object detection problem in the $D_{le}$-based methods e.g.~\cite{yang2021dense}. In fact, $D_{le}$ cannot uniquely define a square BBox. For square-like objects\footnote{Many object instances are in square shape, e.g. the two categories of storage-tank (ST) and roundabout (RA) in the DOTA dataset.}, $D_{le}$-based method will encounter high IoU but high loss value similar to the boundary discontinuity, as shown by the upper part of \textit{Case 3} in Fig.~\ref{fig:problems}. In \text{way1}, the red anchor/proposal $\color{black}{(0,0,45,44,0^\circ)}$ rotates a small angle clockwise to get the blue prediction box. The IoU of green GT $\color{black}{(0,0,45,43,-60^\circ)}$ and the blue prediction box $\color{black}{(0,0,45,44,30^\circ)}$ is close to 1, but the regression loss is high due to the inconsistency of angle parameters. Therefore, the model will rotate a larger angle counterclockwise to make predictions, as described by \text{way2}. The reason for the square-like problem in $D_{le}$-based method is not the above-mentioned PoA and EoE, but the inconsistency of evaluation metric and loss. In contrast, the negative impact of EoE will be weakened when we use $D_{oc}$-based method to detect square-like objects, as shown in the comparison between \textit{Case 2} and the lower part of \textit{Case 3}. Therefore, there is no square-like problem in the $D_{oc}$-based method.
    
    Recent methods start to address these issues. SCRDet~\cite{yang2019scrdet} combines IoU and Smooth L1 loss to propose a IoU-Smooth L1 loss, which does not require the SkewIoU to be gradient backpropagable. It also solves the problem of inconsistency between loss and metric by eliminating the discontinuity of loss at the boundary. However, the gradient direction of IoU-Smooth L1 Loss is still dominated by Smooth L1 loss. RSDet~\cite{qian2021learning} devises modulated loss to smooth the loss mutation at the boundary, but it needs to calculate the loss of as many parameter combinations as possible. CSL~\cite{yang2020arbitrary} transforms angular prediction from a regression problem to a classification problem. CSL needs to carefully design their method according to the  BBox definition ($D_{le}$), and is limited by the  classification granularity with theoretical limitation for high-precision angle prediction. On the basis of CSL, DCL~\cite{yang2021dense} further solves the problem of square-like object detection introduced by $D_{le}$.

	\section{Proposed Method}\label{sec:method}

    \subsection{Gaussian Distribution Modeling}\label{sec:gdm}
    In this paper, we adopt Gaussian modeling to construct more accurate rotation regression loss.
    Most of the IoU-based loss can be considered as a distance function. Inspired by this, we propose a new regression loss based on Gaussian distribution metric. Specifically, we convert a 2-D rotated BBox $\mathcal{B}(x,y,w,h,\theta)$ into a Gaussian distribution $\mathcal{N}(\bm{\mu}_{2d},\mathbf{\Sigma}_{2d})$ (see Fig.~\ref{fig:gpd}) by the following formula:
    \begin{equation}
    \small
        \begin{aligned}
            \mathbf{\Sigma}^{1/2}_{2d}=&\mathbf{R\Lambda R}^{\top}\\
            =&
            \left(                 
              \begin{array}{cc}   
                \cos{\theta} & -\sin{\theta}\\  
                \sin{\theta} & \cos{\theta}\\  
              \end{array}
            \right)
            \left(                 
              \begin{array}{cc}   
                \frac{w}{2} & 0\\  
                0 & \frac{h}{2}\\  
              \end{array}
            \right)
            \left(                 
              \begin{array}{cc}   
                \cos{\theta} & \sin{\theta}\\  
                -\sin{\theta} & \cos{\theta}\\  
              \end{array}
            \right)\\
            =&
            \left(                 
              \begin{array}{cc}   
                \frac{w}{2}\cos^2{\theta}+\frac{h}{2}\sin^2{\theta} & \frac{w-h}{2}\cos{\theta}\sin{\theta}\\  
                \frac{w-h}{2}\cos{\theta}\sin{\theta} & \frac{w}{2}\sin^2{\theta}+\frac{h}{2}\cos^2{\theta}\\  
              \end{array}
            \right)\\
            \bm{\mu}_{2d}=&(x,y)^{\top}
        \end{aligned}
    	\label{eq:gdm_2-D}
    \end{equation}
    where $\mathbf{R}$ represents the rotation matrix, and $\mathbf{\Lambda}$ represents the diagonal matrix of eigenvalues. Gaussian representation is also adopted by \cite{ding2019object}, but it is not validated on more appropriate downstream tasks (i.e. oriented object detection) and no theoretical analysis is given.
    
    According to Eq.~\ref{eq:gdm_2-D}, we have the following properties:
    \begin{itemize}
        \item \textit{Property 1:} $\mathbf{\Sigma}^{1/2}_{2d}(w,h,\theta)=\mathbf{\Sigma}^{1/2}_{2d}(h,w,\theta-\frac{\pi}{2})$;\\
    
        \item \textit{Property 2:} $\mathbf{\Sigma}^{1/2}_{2d}(w,h,\theta)=\mathbf{\Sigma}^{1/2}_{2d}(w,h,\theta-\pi)$;\\
        
        \item \textit{Property 3:} $\mathbf{\Sigma}^{1/2}_{2d}(w,h,\theta)\approx\mathbf{\Sigma}^{1/2}_{2d}(w,h,\theta - \frac{\pi}{2})$, $w\approx h$.
    \end{itemize}
    
    
    	
    
    From the two BBox definitions recall that the conversion between two definitions is, the two sides are exchanged and the angle difference is $\frac{\pi}{2}$. Many methods are designated inherently according to the choice of definition in advance to solve some problems, such as $D_{le}$ for EoE and $D_{oc}$ for square-like problem. It is interesting to note that according to \textit{Property 1}, definition $D_{oc}$ and $D_{le}$ are equivalent based on Gaussian modeling, which makes our method free from the choice of box definitions. This does not mean that the final performance of the two definition methods will be the same. Different factors, e.g. order of edge and angle regression range, will still cause effects. But the method based on Gaussian distribution modeling does not need to bind a certain definition to solve the boundary discontinuity and square-like problem.
    
    Gaussian distribution modeling can also help resolve the boundary discontinuity and square-like problem. The prediction box and GT in \text{way1} of \textit{Case 1} in Fig.~\ref{fig:problems} satisfy the following relation: $x_{p}=x_{gt},y_{p}=y_{gt},w_{p}=h_{gt},h_{p}=w_{gt},\theta_{p}=\theta_{gt}-\frac{\pi}{2}$. According to \textit{Property 1}, the Gaussian distribution corresponding to these two boxes are the same (in the sense of same mean $\bm{\mu}$ and covariance $\mathbf{\Sigma}$), so it naturally eliminates the ambiguity in box representation. Similarly, according to \textit{Properties 2-3}, the GT and prediction box in \text{way1} of \textit{Case 1} and \textit{Case 3} in Fig.~\ref{fig:problems} are also the same or nearly the same (note the approximate equal symbol for $w \approx h$ for square-like boxes) Gaussian. 
    \rev{The Gaussian distribution has degenerated into an isotropic circle, losing the ability to predict the direction, especially the head of the object \cite{yang2022on}, but this does not prevent getting a high IoU prediction in mAP calculation for most 2-D object detection task. However, this can be a big hassle for 3-D object detection, which will be described in detail in Sec. \ref{sec:3-D_loss}.}
    
    \rev{A drawback of Gaussian model is that it cannot be directly applied to quadrilateral/polygon detection~\cite{ming2021optimization,xu2020gliding,guo2021beyond} which is an important task in the applications of aerial images and scene text. The difficulty is how to convert point set into a Gaussian. We leave it for future work \cite{hou2022grep}.}
    
    We now aim to design an easy-to-implement approximate SkewIoU loss to mitigate the inconsistency between metric and loss and achieve high-precision detection.
    \subsection{Between-distribution Metric Implementation}
    \subsubsection{Gaussian Wasserstein Distance}\label{sec:gwd}
    The Gaussian Wasserstein Distance (GWD) \cite{villani2008optimal} between two probability measures $\mathbf{X}_{p} \sim \mathcal{N}_{p}(\bm{\mu}_{p},\bm{\Sigma}_{p})$ and $\mathbf{X}_{t} \sim \mathcal{N}_{t}(\bm{\mu}_{t},\bm{\Sigma}_{t})$ can be expressed as:
    \begin{equation}
    \footnotesize
        \begin{aligned}
            \mathbf{D}_{w}(\mathcal{N}_{p}, \mathcal{N}_{t})^{2}=\lVert \bm{\mu}_{p}-\bm{\mu}_{t}\rVert_{2}^{2}+\mathbf{Tr}\left(\mathbf{\Sigma}_{p}+\mathbf{\Sigma}_{t}-2(\mathbf{\Sigma}_{p}^{1/2}\mathbf{\Sigma}_{t}\mathbf{\Sigma}_{p}^{1/2})^{1/2}\right)
        \end{aligned}
    	\label{eq:wd3}
    \end{equation}
    Note the following equation mathematically holds, which indicates GWD is symmetrical:
    \begin{equation}
        \small
        \begin{aligned}
            \mathbf{Tr}\left((\mathbf{\Sigma}_{p}^{1/2}\mathbf{\Sigma}_{t}\mathbf{\Sigma}_{p}^{1/2})^{1/2}\right)=\mathbf{Tr}\left((\mathbf{\Sigma}_{t}^{1/2}\mathbf{\Sigma}_{p}\mathbf{\Sigma}_{t}^{1/2})^{1/2}\right)
        \end{aligned}
    	\label{eq:wd4}
    \end{equation}
   For horizontal detection setting with a constant angle, we have: $\mathbf{\Sigma}_{p}\mathbf{\Sigma}_{t}=\mathbf{\Sigma}_{t}\mathbf{\Sigma}_{p}$, then Eq.~\ref{eq:wd3} becomes:
    \begin{equation}
    \small
        \begin{aligned}
            \mathbf{D}_{w}^{h}(\mathcal{N}_{p}, \mathcal{N}_{t})^{2}=&\lVert \bm{\mu}_{p}-\bm{\mu}_{t}\rVert_{2}^{2}+\lVert \mathbf{\Sigma}_{p}^{1/2}-\mathbf{\Sigma}_{t}^{1/2} \rVert_{F}^{2}\\
            =&(x_{p}-x_{t})^2+(y_{p}-y_{t})^2+\frac{(w_{p}-w_{t})^2+(h_{p}-h_{t})^2}{4}\\
            =&l_{2}\text{-norm}(\Delta x, \Delta y, \frac{\Delta w}{2}, \frac{\Delta h}{2})
        \end{aligned}
    	\label{eq:wd5}
    \end{equation}
    where $\|\cdot\|_F$ is the Frobenius norm. Note that both boxes are horizontal here, and Eq.~\ref{eq:wd5} is approximately equivalent to the $l_{2}$-norm loss (note the additional denominator of 2 for $w$ and $h$), which is consistent with the loss commonly used in horizontal detection. As Eq. \ref{eq:wd5} calculates the Euclidean distance in absolute coordinates, there is still a gap to Eq. \ref{eq:reg_loss}.
    
    \begin{figure}[!tb]
    	\centering
    	    \includegraphics[width=0.98\linewidth]{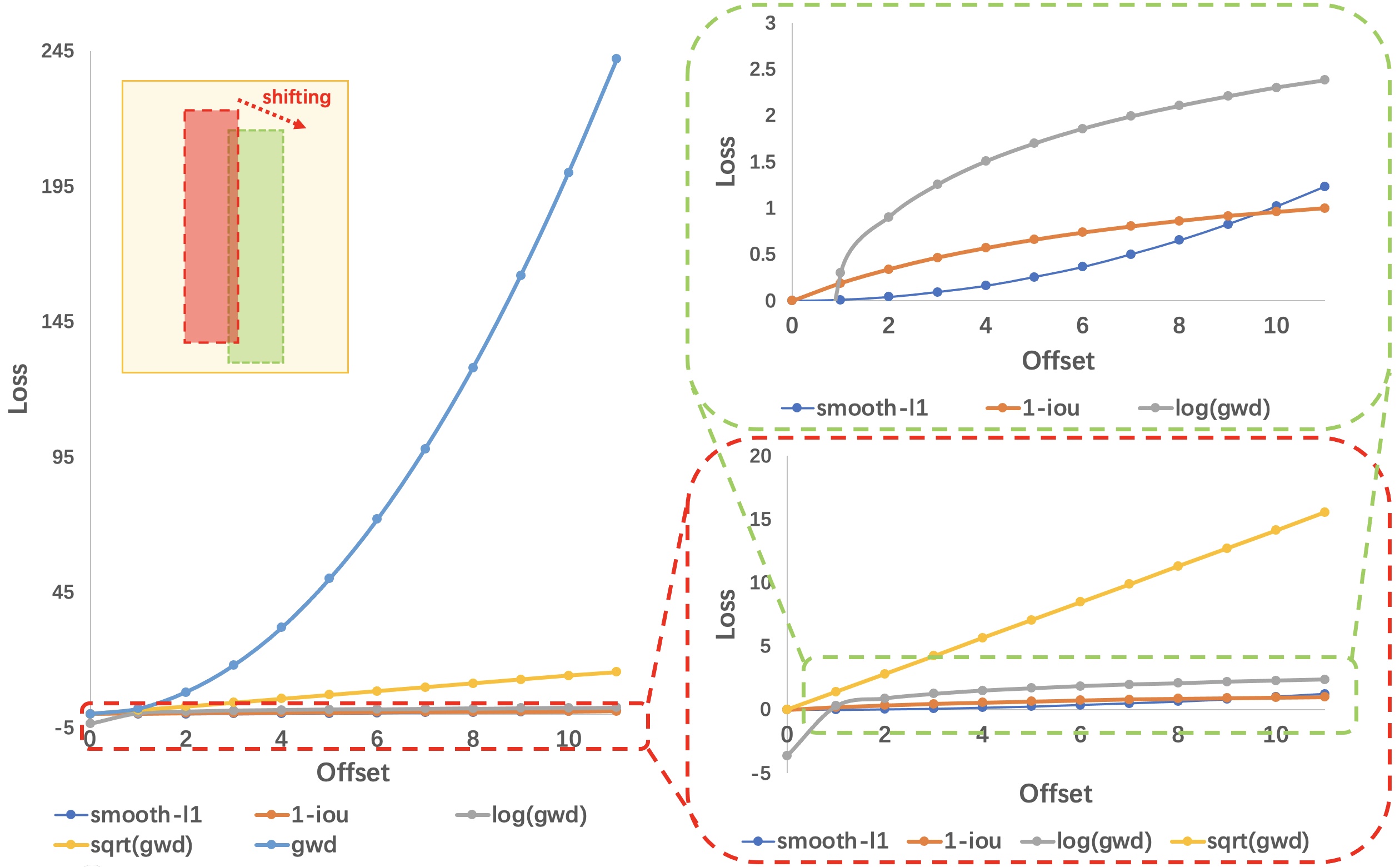}
    	\centering
    	\vspace{-8pt}
    	\caption{Illustration of the sensitivity of GWD to large errors. GWD is very sensitive to large errors and requires a suitable transformation for normalization. The $x$-axis (offset) represents the displacement in pixels}
    	\label{fig:loss_form}
    \end{figure}

    Note that GWD alone can be sensitive to large errors, as shown in the blue curve on the left of Fig. \ref{fig:loss_form}. We perform a nonlinear transformation $f$ and then convert GWD into an affinity measure $\frac{1}{\tau+f(\mathbf{D}_{w}^{2})}$ similar to IoU between two BBoxes. Then we follow the standard IoU based loss form in detection literature~\cite{rezatofighi2019generalized,zheng2020distance}, as written by:
    \begin{equation}
    \small
        \begin{aligned}
        L_{gwd} = 1-\frac{1}{\tau+f(\mathbf{D}_{w}^{2})}, \quad  \tau \geq 1
        \end{aligned}
    	\label{eq:Lgwd}
    \end{equation}
    where $f(\cdot)$ denotes a non-linear function to transform the Wasserstein distance $\mathbf{D}_{w}^{2}$ to make the loss more smooth and expressive. In this paper, we mainly use two nonlinear functions, $sqrt(\mathbf{D}_{w}^{2})$ and $\ln(\mathbf{D}_{w}^{2}+1)$. The hyperparameter $\tau$ modulates the entire loss.
    
    Fig.~\ref{fig:iou_smooth-l1-1} plots the curve under different combinations of $f(\cdot)$ and $\tau$. Compared with the Smooth L1 loss, the curve of Eq.~\ref{eq:Lgwd} is more consistent with the IoU loss curve. Also, we can find in Fig.~\ref{fig:iou_smooth-l1-3} that GWD still can measure the distance between two non-overlapping BBoxes (IoU=0), which is exactly the problem that GIoU and DIoU try to solve in horizontal detection. However, the first term of Eq. \ref{eq:wd3} is the Euclidean distance of the center point between two BBoxes. In addition to making GWD sensitive to large errors, this term also makes the regression loss lose scale invariance. Therefore, we further explore more suitable alternative metrics in subsequent sections.
    
    
    \subsubsection{Kullback-Leibler Divergence}\label{sec:kl}
    We also adopt the Kullback-Leibler divergence (KLD) \cite{kullback1951information}, which can be written by:
    \begin{equation}
    \small
        \begin{aligned}
            \mathbf{D}_{kl}(\mathcal{N}_{p}||\mathcal{N}_{t})=&\frac{1}{2}(\bm{\mu}_{p}-\bm{\mu}_{t})^{\top}\bm{\Sigma}_{t}^{-1}(\bm{\mu}_{p}-\bm{\mu}_{t})\\
            &+\frac{1}{2}\mathbf{Tr}(\bm{\Sigma}_{t}^{-1}\bm{\Sigma}_{p}) + \frac{1}{2}\ln \frac{|\bm{\Sigma}_{t}|}{|\bm{\Sigma}_{p}|}-1
        \end{aligned}
    	\label{eq:kld_pt}
    \end{equation}
    One can also adopt the opposite one: 
    \begin{equation}
    \small
        \begin{aligned}
            \mathbf{D}_{kl}&(\mathcal{N}_{t}||\mathcal{N}_{p})=\\
            &\frac{1}{2}(\bm{\mu}_{p}-\bm{\mu}_{t})^{\top}\bm{\Sigma}_{p}^{-1}(\bm{\mu}_{p}-\bm{\mu}_{t})+\frac{1}{2}\mathbf{Tr}(\bm{\Sigma}_{p}^{-1}\bm{\Sigma}_{t}) + \frac{1}{2}\ln \frac{|\bm{\Sigma}_{p}|}{|\bm{\Sigma}_{t}|}-1
        \end{aligned}
    	\label{eq:kld_tp}
    \end{equation}
    
    
    Although KLD is asymmetric, we find that the optimization principles of these two forms are similar by analyzing the gradients of various parameters and experimental results. 
    Take the relatively simple $\mathbf{D}_{kl}(\mathcal{N}_{p}||\mathcal{N}_{t})$ as an example, according to Eq. \ref{eq:gdm_2-D}, each term of Eq. \ref{eq:kld_pt} can be expressed as
    \begin{equation}
    \small
        \begin{aligned}
           \left (\bm{\mu}_{p}-\bm{\mu}_{t}\right)^{\top}\bm{\Sigma}_{t}^{-1}(\bm{\mu}_{p}-\bm{\mu}_{t})=&\frac{4\left(\Delta x\cos\theta_{t}+\Delta y\sin\theta_{t}\right)^{2}}{w_{t}^{2}}\\
            &+ \frac{4\left(\Delta y\cos\theta_{t}-\Delta x\sin\theta_{t}\right)^{2}}{h_{t}^{2}}
        \end{aligned}
    	\label{eq:kld_item1}
    \end{equation}
    \begin{equation}
    \small
            \mathbf{Tr}(\bm{\Sigma}_{t}^{-1}\bm{\Sigma}_{p})= \left(\frac{h_{p}^{2}}{w_{t}^{2}}+ \theta+\frac{w_{p}^{2}}{h_{t}^{2}}\right)\sin^{2}\Delta \theta+\left(\frac{h_{p}^{2}}{h_{t}^{2}} \theta+\frac{w_{p}^{2}}{w_{t}^{2}}\right)\cos^{2}\Delta \theta
    	\label{eq:kld_item2}
    \end{equation}
    \begin{equation}
    \small
        \begin{aligned}
            \ln \frac{|\bm{\Sigma}_{t}|}{|\bm{\Sigma}_{p}|}=\ln \frac{h_{t}^{2}}{h_{p}^{2}} + \ln \frac{w_{t}^{2}}{w_{p}^{2}}
        \end{aligned}
    	\label{eq:kld_item3}
    \end{equation}
    where $\Delta x=x_{p}-x_{t}, \Delta y=y_{p}-y_{t}, \Delta \theta=\theta_{p}-\theta_{t}$.
    
    \textit{Analysis of high-precision detection.} Without loss of generality, we set $\theta_{t}=0^{\circ}$, then
    \begin{equation}
    \small
        \begin{aligned}
           \frac{\partial \mathbf{D}_{kl}(\mu_{p})}{\partial\mu_{p}}= \left(\frac{4}{w_{t}^{2}}\Delta x, \frac{4}{h_{t}^{2}}\Delta y\right)^{\top}
        \end{aligned}
    \end{equation}
    The weights $1/w_{t}^{2}$ and $1/h_{t}^{2}$ will make the model dynamically adjust the optimization of the object position according to the scale. For example, when the object scale is small or an edge is too short, the model will pay more attention to the optimization of the offset of the corresponding direction. For this kind of object, a slight deviation on the corresponding direction will often cause a sharp drop in SkewIoU. When $\theta_{t} \neq 0^{\circ}$, the gradient of the object offset ($\Delta x$ and $\Delta y$) will be dynamically adjusted according to the $\theta_{t}$ for better optimization. In contrast, the gradient of the center point in GWD and L$_{2}$-norm are $\frac{\partial \mathbf{D}_{w}(\mu_{p})}{\partial\mu_{p}}=(2\Delta x, 2\Delta y)^{\top}$  and $\frac{\partial \mathbf{D}_{L_{2}}(\mu_{p})}{\partial\mu_{p}}=(\frac{2}{w_{a}^{2}}\Delta x, \frac{2}{h_{a}^{2}}\Delta y)^{\top}$. The former cannot adjust the dynamic gradient according to the length and width of the object. The latter is based on the length and width of the anchor ($w_{a},h_{a}$) to adjust the gradient instead of the target object ($w_{t},h_{t}$), which is almost ineffective for those detectors \cite{yang2019scrdet, yang2021r3det, ming2021dynamic, ming2021cfc, ming2021optimization, han2021align} that use horizontal anchors for rotation detection. More importantly, they are not related to the angle of the target object when $\theta_{t} \neq0^\circ$. Therefore, the detection result of the GWD-based and L$_{n}$-norm models will show a slight deviation, while the detection result of the KLD-based model is quite accurate, as shown in Fig. \ref{fig:high_precision_compare_vis}. 
    
    \begin{figure}[!tb]
    	\centering
    	\includegraphics[width=1.0\linewidth]{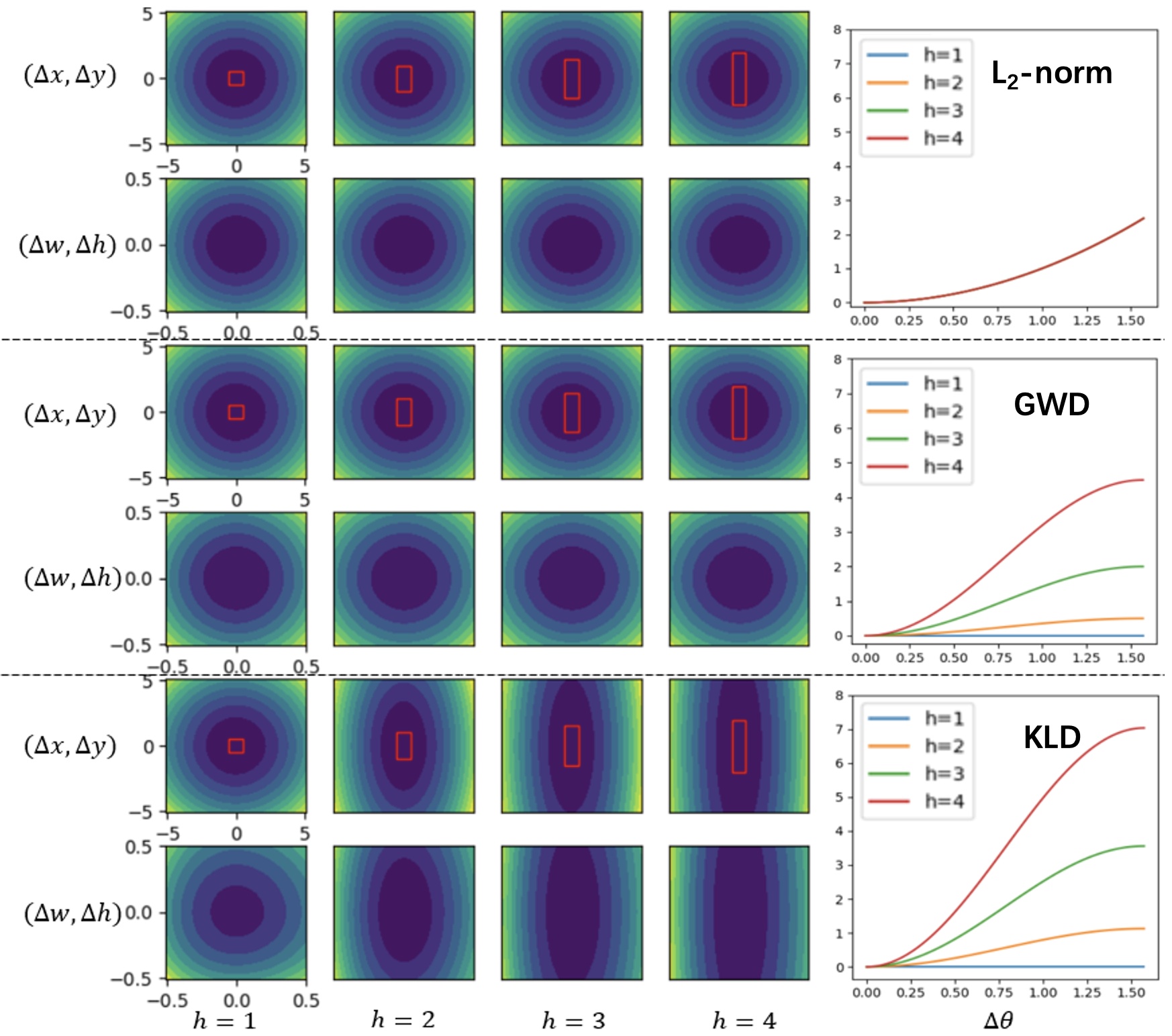}
    	\vspace{-12pt}
    	\caption{\rev{Behavior of L$_2$-norm, GWD and KLD versus parameters when the targeted height varies, as generated by a simulation test. Left: the gradient landscape of $(\Delta x,\Delta y,\Delta w,\Delta h)$; Right: gradient curve of $\Delta \theta$.}}
    	\label{fig:p_r}
    \end{figure}
    
    For $h_{p}$ (similar for $w_{p}$), we have
    \begin{equation}
    \small
           \frac{\partial \mathbf{D}_{kl}(\bm{\Sigma}_{p})}{\partial \ln{h_{p}}}=  \frac{h_{p}^{2}}{h_{t}^{2}}\cos^{2}\Delta \theta + \frac{h_{p}^{2}}{w_{t}^{2}}\sin^{2}\Delta \theta - 1 
    \end{equation}
    
    On one hand, the optimization of the $h_{p}$ and $w_{p}$ by updating their gradients is affected by the $\Delta \theta$. When $\Delta \theta=0^{\circ}$, $\frac{\partial \mathbf{D}_{kl}(\bm{\Sigma}_{p})}{\partial \ln{h_{p}}}=  \frac{h_{p}^{2}}{h_{t}^{2}} - 1, \frac{\partial \mathbf{D}_{kl}(\bm{\Sigma}_{p})}{\partial \ln{w_{p}}}=  \frac{w_{p}^{2}}{w_{t}^{2}} -1$, which means that the smaller targeted height or width leads to heavier penalty on its matching loss. This is desirable, as smaller height or width needs higher matching precision. On the other hand, the optimization of $\theta_p$ is also affected by $h_{p}$ and $w_{p}$:
    \begin{equation}
    \small
        \begin{aligned}
           \frac{\partial \mathbf{D}_{kl}(\bm{\Sigma}_{p})}{\partial\theta_{p}}= \left(\frac{h_{p}^{2}-w_{p}^{2}}{w_{t}^2}+\frac{w_{p}^{2}-h_{p}^{2}}{h_{t}^{2}}\right)\sin{2\Delta \theta}
        \end{aligned}
    \end{equation}
    When $w_{p}=w_{t}$, $h_{p}=h_{t}$, then we can get $\frac{\partial \mathbf{D}_{kl}(\bm{\Sigma}_{p})}{\partial\theta_{p}}= \left(\frac{h_{t}^{2}}{w_{t}^2}+\frac{w_{t}^{2}}{h_{t}^{2}} - 2\right)\sin{2\Delta \theta} \geq \sin{2\Delta \theta}$, the condition for the equality sign is $h_{t}=w_{t}$. This shows that the larger the aspect ratio of the object, the model will pay more attention to the optimization of the angle. This is the main reason why the KLD-based model has a huge advantage in high-precision detection indicators as a slight angle error would cause a serious accuracy drop for large aspect ratios objects. Through the above analysis, we find that when one of the parameters is optimized, the other parameters will be used as its weight to dynamically adjust the optimization rate. In other words, the optimization of parameters is no longer independent, that is, optimizing one parameter will also promote the optimization of other parameters. We believe this largely contributes to the effectiveness of KLD-based regression loss. In addition, $\mathbf{D}_{kl}(\mathcal{N}_{t}||\mathcal{N}_{p})$ has similar properties, refer to appendix in conference version for details.
    
    \rev{In general, KLD can be suited to high-precision detection especially for objects with large aspect ratio. For bounding box with larger aspect ratio, KLD gives heavier penalties to matching of shorter edge's length and the center point's position along the shorter edge's direction, as well as the matching of angle. These characteristics are desirable, as when matching bounding box with large aspect ratio, IoU can be intuitively sensitive to the shorter edge's length, the center point's position along the shorter edge's direction and the angle. Specifically, we consider a target box with $x=0$, $y=0$, $w=1$, $\theta=0$, and set $h=\{1,2,3,4\}$ to control the aspect ratio, and plot KLD versus parameter variation in Fig. \ref{fig:p_r} where L$_2$-norm and GWD are also included for comparison. When $h$ increases, KLD is more sensitive to the variation of $x$, $w$ and $\theta$, meaning it has desirable advantages for objects with large aspect ratio. Comparatively, both L$_2$-norm and GWD pay no more attention to the matching of $x$ and $w$ when $h$ increases, and L$_2$-norm is even unchanged when the difference of angle $\Delta\theta$ is fixed.}
    
    \rev{\textit{Scale invariance.} 
    Suppose there are two Gaussian distributions, denoted as $\mathbf{X}_{p} \sim \mathcal{N}_{p}(\bm{\mu}_{p},\bm{\Sigma}_{p})$ and $\mathbf{X}_{t} \sim \mathcal{N}_{t}(\bm{\mu}_{t},\bm{\Sigma}_{t})$. Then, for a full-rank matrix $\mathbf{M}$, $|\mathbf{M}|\neq 0$, we have $\mathbf{X}_{p^{'}} =\mathbf{M}\mathbf{X}_{p} \sim \mathcal{N}_{p}(\mathbf{M}\bm{\mu}_{p},\mathbf{M}\bm{\Sigma}_{p}\mathbf{M}^{\top})$, $\mathbf{X}_{t^{'}}=\mathbf{M}\mathbf{X}_{t} \sim \mathcal{N}_{t}(\mathbf{M}\bm{\mu}_{t},\mathbf{M}\bm{\Sigma}_{t}\mathbf{M}^{\top})$, denoted as $\mathcal{N}_{p^{'}}$ and $\mathcal{N}_{t^{'}}$. The Kullback-Leibler Divergence (KLD) between $\mathcal{N}_{p^{'}}$ and $\mathcal{N}_{t^{'}}$ is:
    \begin{equation}
    \small
        \begin{aligned}
            \mathbf{D}_{kl}&(\mathcal{N}_{p^{'}}||\mathcal{N}_{t^{'}})=\frac{1}{2}(\bm{\mu}_{p}-\bm{\mu}_{t})^{\top}\mathbf{M}^{\top}(\mathbf{M}^{\top})^{-1}\bm{\Sigma}_{t}^{-1}\mathbf{M}^{-1}\mathbf{M}(\bm{\mu}_{p}-\bm{\mu}_{t})\\
            &+\frac{1}{2}\mathbf{Tr}\left((\mathbf{M}^{\top})^{-1}\bm{\Sigma}_{t}^{-1}\mathbf{M}^{-1}\mathbf{M}\bm{\Sigma}_{p}\mathbf{M}^{\top}\right) +\frac{1}{2}\ln \frac{|\mathbf{M}||\bm{\Sigma}_{t}||\mathbf{M}^{\top}|}{|\mathbf{M}||\bm{\Sigma}_{p}||\mathbf{M}^{\top}|}-1 \\
            &=\mathbf{D}_{kl}(\mathcal{N}_{p}||\mathcal{N}_{t})
        \end{aligned}
    	\label{eq:proof_scale_invariance}
    \end{equation}}
    \rev{where we have $\mathbf{Tr}(\mathbf{A}\mathbf{B})=\mathbf{Tr}(\mathbf{B}\mathbf{A})$.}
    \begin{figure}[!tb]
    	\centering
    	\includegraphics[width=0.8\linewidth]{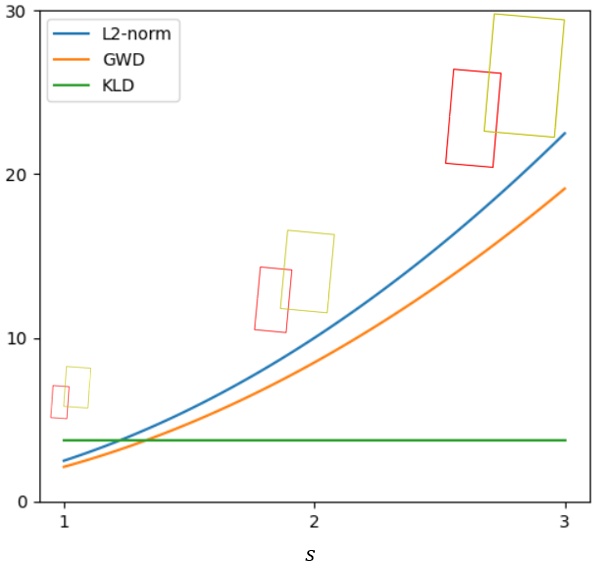}
    	\vspace{-10pt}
    	\caption{\rev{The loss of L$_2$-norm, GWD and KLD versus scaling factor, generated by a simulation test. Only the value of KLD is invariant to the scaling factor $s$. Th bounding boxes denote different scales of objects.}}
    	\label{fig:l_s}
    \end{figure}

    \rev{Therefore, KLD can achieve affine invariance. When $\mathbf{M}=k\mathbf{I}$ ($\mathbf{I}$ is identity matrix), its scale invariance holds. To visualize the scale invariance of KLD, we consider the KLD of two given boxes, and investigate the variation of KLD when the two boxes are enlarged with a scaling factor $s$.  
    As shown in Fig. \ref{fig:l_s}, the value of KLD is invariant to the scaling factor $s$. 
    Compared with this, the values of L$_2$-norm and GWD change when $s$ increases.
    }
    
    \textit{Horizontal special case.} For horizontal detection, combine Eq. \ref{eq:kld_pt} to Eq. \ref{eq:kld_item3}, we have
    \begin{equation}
    \footnotesize
        \begin{aligned}
            \mathbf{D}_{kl}^{h}(\mathcal{N}_{p}||\mathcal{N}_{t})=&\frac{1}{2}\left(\frac{w_{p}^2}{w_{t}^2}+\frac{h_{p}^{2}}{h_{t}^{2}}+\frac{4\Delta^{2}x}{w_{t}^{2}}+\frac{4\Delta^{2}y}{h_{t}^{2}}+\ln \frac{w_{t}^{2}}{w_{p}^{2}}+\ln \frac{h_{t}^{2}}{h_{p}^{2}}-2\right)\\
            =&2l_{2}\text{-norm}(\Delta t_{x}, \Delta t_{y})+l_{1}\text{-norm}(\Delta t_{w}, \Delta t_{h})\\
            &+\frac{1}{2}l_{2}\text{-norm}\left(\frac{1}{\Delta t_{w}}, \frac{1}{\Delta t_{h}}\right) -1
        \end{aligned}
    	\label{eq:kl_h}
    \end{equation}
    where the first two terms are very similar to those in Eq. \ref{eq:reg_loss}, and the divisor part of the two terms $x$ and $y$ is the main difference ($\frac{\Delta x}{w_{t}}$ vs. $\frac{\Delta x}{w_{a}}$).
    
    \textit{Variants of KLD.} 
    We introduce two variants~\cite{jeffreys1946invariant,manning1999foundations} to verify the influence of asymmetry on rotation detection:
    \begin{equation}
    \footnotesize
        \begin{aligned}
            \mathbf{D}_{js}(\mathcal{N}_{p}||\mathcal{N}_{t})=&\frac{1}{2}\left(\mathbf{D}_{kl}\left(\mathcal{N}_{t}||\frac{\mathcal{N}_{p}+\mathcal{N}_{t}}{2}\right)+\mathbf{D}_{kl}\left(\mathcal{N}_{p}||\frac{\mathcal{N}_{p}+\mathcal{N}_{t}}{2}\right)\right)\\ \mathbf{D}_{jef}(\mathcal{N}_{p}||\mathcal{N}_{t})=&\mathbf{D}_{kl}(\mathcal{N}_{t}||\mathcal{N}_{p})+\mathbf{D}_{kl}(\mathcal{N}_{p}||\mathcal{N}_{t})
        \end{aligned}
    \end{equation}
    
    \subsubsection{Bhattacharyya Distance}\label{sec:bcd}
    We adopt the Bhattacharyya Distance (BCD)~\cite{bhattacharyya1943measure}, which is specified as follows, where $\bm{\Sigma} = \frac{1}{2}(\bm{\Sigma}_{p}+\bm{\Sigma}_{t})$:
    \begin{equation}
    \small
            \mathbf{D}_{bcd}(\mathcal{N}_{p},\mathcal{N}_{t})=\frac{1}{8}(\bm{\mu}_{p}-\bm{\mu}_{t})^{\top}\bm{\Sigma}^{-1}(\bm{\mu}_{p}-\bm{\mu}_{t})
            +\frac{1}{2}\ln\frac{\det(\bm{\Sigma})}{\sqrt{\det(\bm{\Sigma}_{p}\bm{\Sigma}_{t})}}
    	\label{eq:bcd}
    \end{equation}

    Compared with the above two metrics, BCD is symmetrical and scale invariance, and has a similar parameter optimization mechanism to KLD. Experimental results show that BCD and KLD achieve the similar performance, thus we omit the analysis and verification to BCD.
    
    \rev{\subsection{Label Assignment based on Gaussian Metric}\label{sec:la}
    The label assignment strategy is a key component of object detection \cite{zhang2020bridging,ming2021dynamic}, which aims to assign targets, foreground or background, to sampled regions in an image.
    Many current methods use IoU as the basis for label assignment, e.g. the Max-IoU strategy \cite{ren2015faster,lin2017focal}. To align label assignment and regression loss, we propose a new label assignment strategy based on Gaussian metric, that is, Gaussian metric (e.g. KLD) replaces IoU as the basis for sample division.
    However, setting the threshold is a tricky problem because the Gaussian metric does not have a very intuitive physical meaning like IoU. Inspired by Adaptive Training Sample Selection (ATSS) \cite{zhang2020bridging}, we adopt a statistical approach to automatically calculate appropriate thresholds. For the $i$-th GT box ($g_{i}$), the dynamic threshold $t_{g_{i}}$ is: 
    \begin{equation}
    \small
        \begin{aligned}
        t_{g_{i}} =& m_{g_{i}} + v_{g_{i}},\quad N=kL, \quad \mathcal{G}_{ij} = \frac{1}{\tau+\mathcal{D}_{ij}}\\
        m_{g_{i}} =& \frac{1}{N}  \sum_{j=1}^{N}{\mathcal{G}_{ij}}, \quad v_{g_{i}} = \sqrt{\frac{1}{N}  \sum_{j=1}^{N}{(\mathcal{G}_{ij} 
         - m_{g_{i}})^{2}}}
        \end{aligned}
        \label{eq:atss}
    \end{equation}
    where $k$ and $L$ represent the number of proposals/anchors closest to the center point of GT box and that of feature pyramid layers in the detector neck. $\mathcal{D}_{ij}$ is the Gaussian metric between the $i$-th GT box and the $j$-th proposal/anchor. We set $\tau=2$ by default to modulate the loss.}

	\subsection{Overall Loss Function Design}\label{sec:loss}
    In this section, we take 2-D object detection as the main example. In line with~\cite{yang2020arbitrary,yang2021dense,yang2021r3det}, we use the one-stage detector RetinaNet~\cite{lin2017focal} as the baseline. Rotated rectangle is represented by five parameters ($x,y,w,h,\theta$). In our experiments we mainly follow $D_{oc}$. First of all, we need to clarify that the network has not changed the output of the original regression branch, that is, it is not directly predicting the parameters of the Gaussian distribution. The process of Gaussian distribution metric is as follows: i) predict offset ($t_{x}^{*},t_{y}^{*},t_{w}^{*},t_{h}^{*},t_{\theta}^{*}$); ii) decode prediction box; iii) convert prediction box and target ground-truth into Gaussian distribution; iv) calculate $L_{reg}$ of two Gaussian distributions. Therefore, the inference time remains unchanged. 
    
    The regression equation of ($x,y,w,h$) are already listed in Eq. \ref{eq:reg_pred} and Eq. \ref{eq:reg_target}. As for the regression equation of $\theta$, we use two forms as the baseline to be compared:
    \begin{itemize}
    	\item Direct regression, marked as \textit{Reg. ($\Delta \theta$)}. The model directly predicts the angle offset $t_{\theta}^{*}$:
    	\begin{equation}
        \begin{aligned}
        	t_{\theta}=&(\theta-\theta_{a})\cdot\pi/180, \quad t_{\theta}^{*}=&(\theta_{}^{*}-\theta_{a})\cdot\pi/180
        \end{aligned}
        \end{equation}
    	\item Indirect regression: \textit{Reg.$^*$ ($\sin{\theta}$, $\cos{\theta}$)}. The model predicts two vectors ($t_{\sin\theta}^{*}$ and $t_{\cos\theta}^{*}$) to match the two targets from the GT ($t_{\sin\theta}$ and $t_{\cos\theta}$):
    	\begin{equation}
        \begin{aligned}
            t_{\sin\theta} =& \sin{(\theta\cdot\pi/180)}, \quad t_{\cos\theta} = \cos{(\theta\cdot\pi/180)} \\ t_{\sin\theta}^{*}=&\sin{(\theta^{*}\cdot\pi/180)}, \quad t_{\cos\theta}^{*}=\cos{(\theta^{*}\cdot\pi/180)}
        \end{aligned}
        \end{equation}
    \end{itemize}
    To ensure that $t_{\sin\theta}^{*2}+t_{\cos\theta}^{*2}=1$ is satisfied, we will perform the following normalization processing:
    \begin{equation}
        \small
        \begin{aligned}
        t_{\sin\theta}^{*}=\frac{t_{\sin\theta}^{*}}{\sqrt{t_{\sin\theta}^{*2}+t_{\cos\theta}^{*2}}}, \quad
        t_{\cos\theta}^{*}=\frac{t_{\cos\theta}^{*}}{\sqrt{t_{\sin\theta}^{*2}+t_{\cos\theta}^{*2}}}
        \end{aligned}
    \end{equation}

    Indirect regression is a simpler way to avoid boundary discontinuity. The multi-task loss is defined as follows:
    \begin{equation}
    \small
    	\begin{aligned}
    	L = \frac{\lambda_{1}}{N}\sum_{n=1}^{N}obj_{n}\cdot L_{reg}(b_{n},gt_{n}) + \frac{\lambda_{2}}{N}\sum_{n=1}^{N}L_{cls}(p_{n},t_{n})
    	\label{eq:multitask_loss}
    	\end{aligned}
    \end{equation}
    where $N$ is the number of anchors, and $obj_{n}$ is a binary value ($obj_{n}=1$ for foreground and otherwise background -- no regression for background). $b_{n}$ denotes the $n$-th predicted BBox, $gt_{n}$ is the $n$-th target ground-truth. $t_{n}$ denotes the label of the $n$-th object, $p_{n}$ is the $n$-th probability distribution of various classes calculated by Sigmoid function. The hyper-parameter $\lambda_{1}$, $\lambda_{2}$ control the trade-off and are set to $\{2,1\}$ by default. The classification loss $L_{cls}$ is set as the focal loss \cite{lin2017focal} in our experiments. The regression loss $L_{reg}$ is set by Eq.~\ref{eq:Lgwd} for GWD loss (or for the BCD and KL loss). 
    
    \begin{figure}[!tb]
    	\centering
    	\subfigure[Detection comparison from the top view between existing binary-classification based methods for heading e.g. SECOND~\cite{yan2018second} and ours.]{
    		\begin{minipage}[t]{0.98\linewidth}
    			\centering
    			\includegraphics[width=0.98\linewidth]{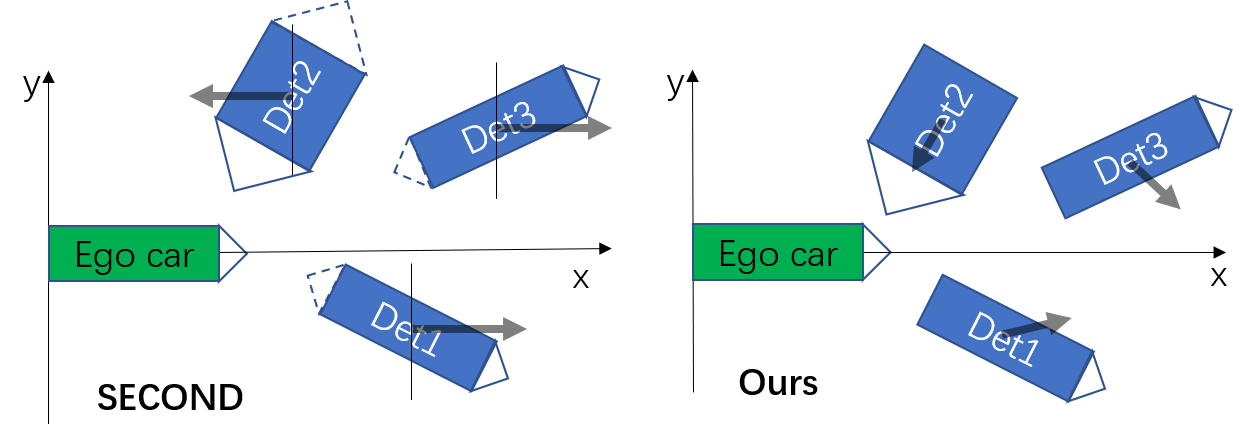}
    		\end{minipage}%
    		\label{fig:ambiguity_resolve}
    	}\\
    	\vspace{-8pt}
    	\subfigure[3-D BBox with square shape in top-view e.g.  pedestrian.]{
    		\begin{minipage}[t]{0.42\linewidth}
    			\centering
    			\includegraphics[width=0.98\linewidth,height=4.2cm]{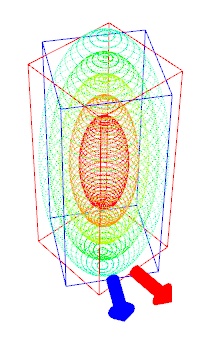}
    		\end{minipage}%
    		\label{fig:degeneration_3-D}
    	}
    	\subfigure[Top view of the 3-D BBox and the heading is arbitrary given the isotropic 2-D Gaussian.]{
    		\begin{minipage}[t]{0.45\linewidth}
    			\centering
    			\includegraphics[width=0.98\linewidth]{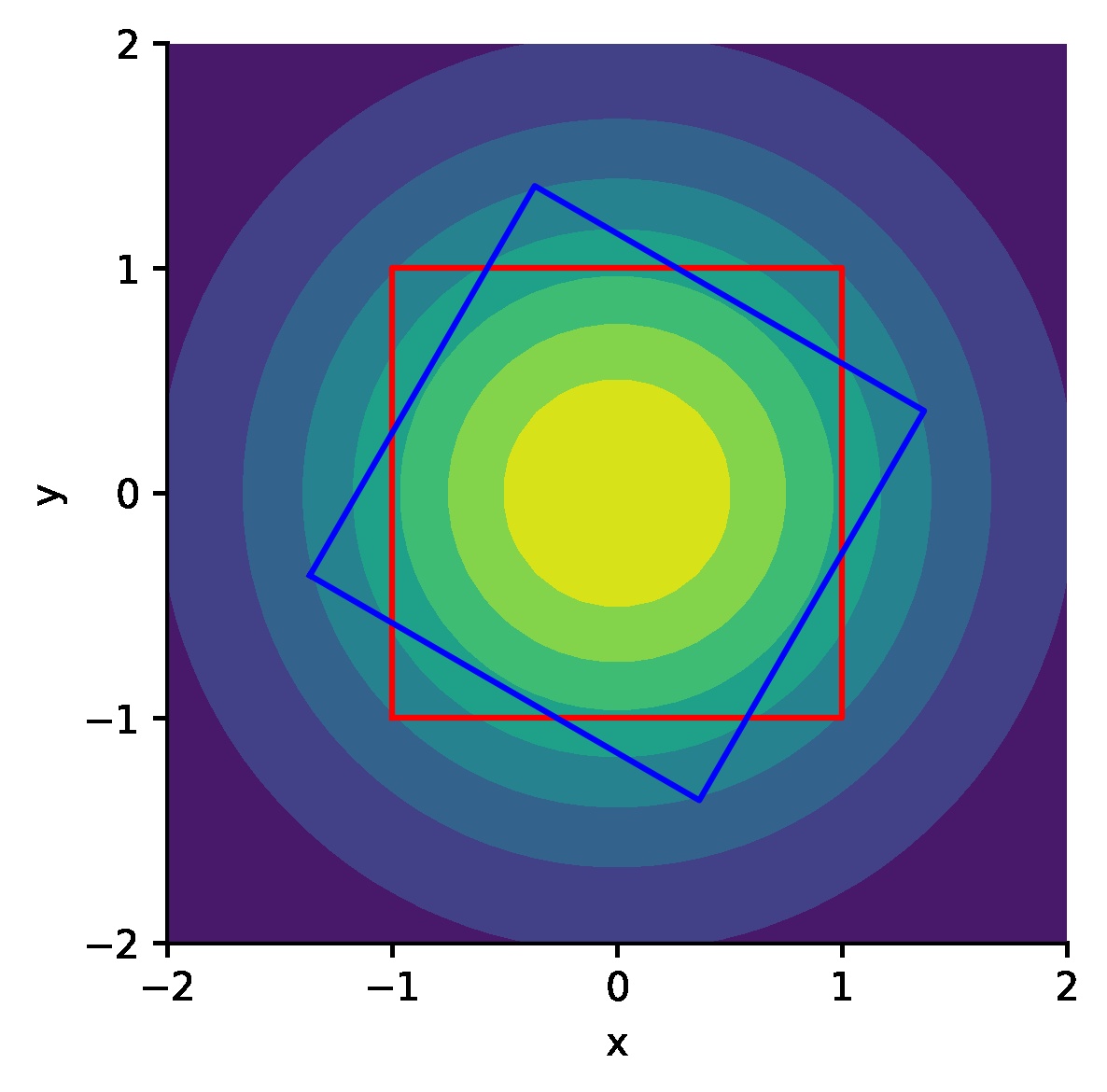}
    		\end{minipage}
    		\label{fig:degeneration_2-D}
    	}
    \centering\vspace{-12pt}
    	\caption{\rev{Orientation degeneration cases. Note that the red and blue boxes/cubes share the same Gaussian distribution representation.}}
    	\label{fig:degeneration}
    \end{figure}
    
    \begin{figure}[!tb]
    	\centering
    	\subfigure[Post-processing pipeline of our Gaussian-based loss.]{
    		\begin{minipage}[t]{0.98\linewidth}
    			\centering
    			\includegraphics[width=0.98\linewidth]{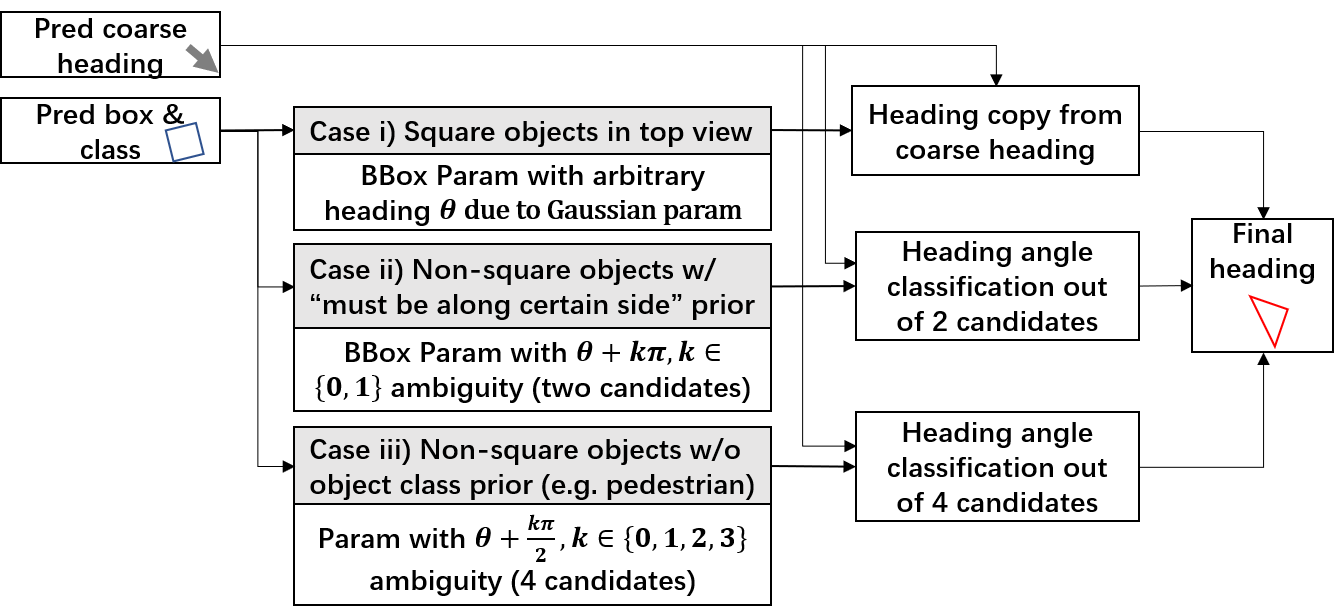}
    		\end{minipage}%
    		\label{fig:process_pipeline}
    	}
    	\subfigure[Condition rule table for our post-processing.]{
    		\begin{minipage}[t]{0.98\linewidth}
    			\centering
    			\includegraphics[width=0.9\linewidth]{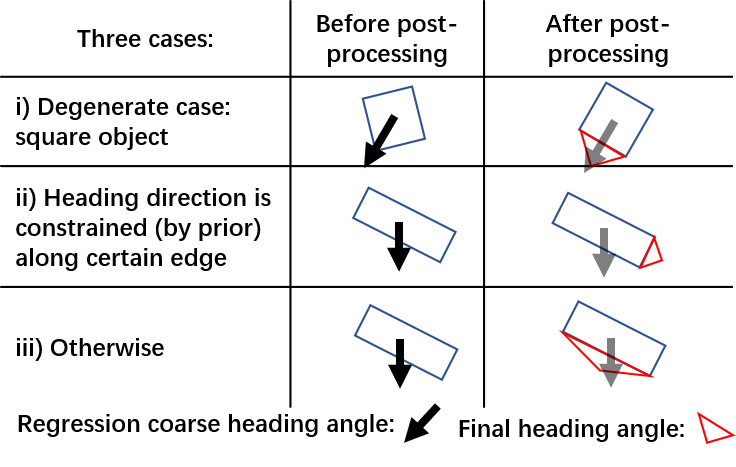}
    		\end{minipage}
    		\label{fig:condition_table}
    	}
    \centering\vspace{-12pt}
    	\caption{\rev{The proposed processing logic after the end-to-end training regression network to derive the final heading angle. The condition rule table deals with three cases according to the shape of the bounding box: i) square object; ii) the rectangle object with its heading direction confined to the angle forming an acute angle to  one of a certain bi-direction, e.g. a vehicle with its long edge as heading direction. iii) otherwise, the final direction is chosen as the one (red triangle) that forms an acute angle with the heading regression result (black arrow). }}
    	\label{fig:3-Ddetect}
    \end{figure}
    
	\subsection{\rev{Extending Gaussian Modeling to 3-D Detection}}\label{sec:3-D_loss}
	\rev{Now we show how to (non-trivially) extend the Gaussian parametric model from 2-D detection to 3-D. Rather than only using the oriented box in birds-eye-view (BEV), practical 3-D detectors are often required to predict the object's height and altitude in 3-D. More specifically, consider the common simplification in many 3-D applications and existing benchmarks assuming a constant pitch and roll angle of moving objects, we can extend Eq.~\ref{eq:gdm_2-D} to 3-D by:}
    \begin{equation}
        \begin{aligned}
            \mathbf{\Sigma}^{1/2}_{3d}=&\mathbf{R\Lambda R}^{\top},\
            \bm{\mu}_{3d}=(x,y,z)^{\top}
        \end{aligned}
    	\label{eq:gdm_3-D}
    \end{equation}
    \begin{equation}
        \begin{aligned}
        \mathbf{R}=\left(                 
                \begin{array}{ccc}   
                \cos{\theta} & -\sin{\theta} & 0\\  
                \sin{\theta} & \cos{\theta} & 0\\ 
                0 & 0 & 1
              \end{array}
            \right), \
        \mathbf{\Lambda}=\left(                 
                \begin{array}{ccc}   
                \frac{w}{2} & 0 & 0\\  
                0 & \frac{h}{2} & 0\\
                0 & 0 & \frac{l}{2}
              \end{array}
            \right)
        \end{aligned}
    \end{equation}
    \rev{where $w$, $h$, $l$ represents the width, height and length of the 3-D BBox, respectively. Heading angle is denoted by $\theta$.}
    
    \rev{In contrast to the 2-D case whereby the object bounding box IoU is the main performance interest, the object's heading information is often required in 3-D. For instance, in Waymo Open Dataset~\cite{sun2020scalability} which is a widely used benchmark for autonomous driving, the metric mAP weighted by heading accuracy (mAPH) is particularly designed.}
    
    \rev{However, the heading angle is ambiguous given only the parameters of detected BBox. In the previous OpenCV/long edge definitions, the ambiguity refers to the choice of $\theta$ and $\theta+180^\circ$ as shown in Fig.~\ref{fig:ambiguity_resolve}, which is often resolved by a binary classifier as introduced by existing 3-D detectors e.g. in SECOND \cite{yan2018second}, PointPillar \cite{lang2019pointpillars} (see the left of Fig.~\ref{fig:ambiguity_resolve}), and a similar scheme has also been devised in a recent 2-D detector~\cite{yang2022on}. While for our Gaussian-based BBox parameterization, we not only need to address the above ambiguity, but also to handle a degenerate case when the object in top view is in square form as shown in Fig.~\ref{fig:degeneration_3-D}. In this case, as shown in Fig.~\ref{fig:degeneration_2-D}, the Gaussian distribution becomes geometrically isotropic resulting in the loss no matter KLD or others is inherently agnostic to the heading of the object\footnote{\rev{Such a degeneration can also happen in 2-D for square objects. Fortunately in many 2-D benchmarks, the heading information is not of interest (mAP cannot reflect the heading accuracy) or the object itself is heading-invariant (e.g. storage-tank and roundabout in aerial images).}}.}

	\rev{To handle the above degenerating case for square-like detecting BBox in Gaussian modeling, we devise a post-processing pipeline which considers the three exclusive cases as shown in Fig.~\ref{fig:process_pipeline}. Specifically, we introduce a network layer to regress a coarse heading angle $\theta_{c}$, in addition to the BBox regression (see the left top in Fig.~\ref{fig:process_pipeline}). Then we divide the BBox into three cases for deciding the value of the final heading $\theta$: i) a square object such that the Gaussian parameterization becomes agnostic to the heading angle, and we use the coarse angle as the final heading output. ii) for rectangle object, the model can utilize an important common prior that the heading must be along a certain bi-direction e.g. the long edge direction for a vehicle and in this case, the final heading is chosen by forming an acute angle to that prior direction. iii) otherwise, the heading is taken by forming an acute angle to the coarse heading $\theta_{c}$.}         
	The post-processing logic is shown in Algorithm \ref{algorithm:degeneration_3-D}.
    
	\begin{algorithm}[t!]
	\small \small
		\caption{\rev{Post-processing for square-like degradation cases in Gaussian-based 3-D object detection.}} 
		\label{algorithm:degeneration_3-D}
		\hspace*{0.02in} {\bf Input:} 
		$(x,y,z,w,h,l,\theta)$: predicted cube parameters, $(d_{x},d_{y})$: predicted heading vector, $c$: cube's predicted class, $r$: ratio threshold, $C$: classes whose head along the long side. \\
		\hspace*{0.02in} {\bf Output:} post-processed cube parameters:
		$(x,y,z,w,h,l,\theta)$.
		\begin{algorithmic}[1]
			\State $\theta_{d} \leftarrow atan2(d_{x}, d_{y})$;
			\If{$r^{-1}<w/h<r$}
			\State $\theta \leftarrow \theta_{d}$
			\State $w,h \leftarrow max(w,h), max(w,h)$
			\EndIf
			\If{$c \in C$} // This cube's heading direction should be parallel to the long border of its BEV projected box
			    \If{$w<h$}
			        \State $\theta \leftarrow \theta + \frac{\pi}{2}$
			        \State $w,h \leftarrow h, w$ 
			    \EndIf
			    \State $n = [\frac{\theta_{d}-\theta}{\pi}]$
			    \State $\theta \leftarrow limit\_period((\theta+\pi(n \mod 2)), [-\pi,\pi))$ // We select the heading angle closer to the angle decoded from the predict heading vector.
			\Else
			    \State $n = [\frac{\theta_{d}-\theta}{\pi/2}]$
			    \State $\theta \leftarrow limit\_period((\theta+\frac{\pi}{2}(n \mod 4)), [-\pi,\pi))$ // We select the heading angle closer to the angle decoded from the predict heading vector.
			    \If{$n$ is odd} // If we rotate the box by odd multiples of $pi/2$, we have to swap the cubes w and h to guarantee the box's shape does not change. 
			        \State $w,h \leftarrow h.w$
			    \EndIf
			\EndIf
		\end{algorithmic}
	\end{algorithm}

    
    

	
	\section{Experiments}\label{sec:experiments}
	For 2-D rotated object detection, we use Tensorflow \cite{abadi2016tensorflow} for implementation under our previously released rotation detection framework~\cite{yang2021alpharotate} by default unless otherwise specified. We have made our proposed detectors open sourced\footnote{\url{https://github.com/yangxue0827/RotationDetection}}. For 3-D object detection, we use third-party tools, MMDetection3D \cite{chen2019mmdetection} and the source code is also publicly available\footnote{\url{https://github.com/zhanggefan/mmdet3d-gaussian}}. \rev{All the 2-D experiments are performed with GeForce RTX 2080 Ti and 12G memory, while the 3-D experiments are performed with GeForce RTX 3090 Ti and 24G memory.}

    \subsection{2-D Datasets and Implementation Details}
    \textit{DOTA} \cite{xia2018dota} is one of the largest datasets for oriented object detection in aerial images with three versions: DOTA-v1.0, DOTA-v1.5 and DOTA-v2.0. 
    DOTA-v1.0 contains 15 common categories, 2,806 images and 188,282 instances. The proportions of the training set, validation set, and testing set in DOTA-v1.0 are 1/2, 1/6, and 1/3, respectively. 
    In contrast, DOTA-v1.5 uses the same images as DOTA-v1.0, but extremely small instances (less than 10 pixels) are also annotated. Moreover, a new category, 
    containing 402,089 instances in total is added in this version. While DOTA-v2.0 contains 18 common categories (two new categories),
    11,268 images and 1,793,658 instances. Compared to DOTA-v1.5, it further includes the new categories. The 11,268 images in DOTA-v2.0 are split into training, validation, test-dev, and test-challenge sets. We divide the images into 600 $\times$ 600 subimages with an overlap of 150 pixels and scale it to 800 $\times$ 800, in line with the cropping protocol in literature.
    
    \textit{DIOR-R} \cite{cheng2022anchor} is an aerial image dataset annotated by rotated BBoxes. There are 23,463 images and 190,288 instances, covering 20 object classes. DIOR-R has a high variation of object size, both in spatial resolutions, and in the aspect of inter‐class and intra‐class size variability across objects. Different imaging conditions, weathers, seasons, image quality are the major challenges of DIOR-R. Besides, it has high inter‐class similarity and intra‐class diversity.
    
    \textit{UCAS-AOD} \cite{zhu2015orientation} contains 1,510 aerial images of about 659 $\times$ 1,280 pixels, with 2 categories of 14,596  instances. In line with~\cite{azimi2018towards,xia2018dota}, we sample 1,110 images for training and 400 for testing. 
    
    \textit{HRSC2016} \cite{liu2017high} contains images from two scenarios including ships on sea and ships close inshore. 
    All images are collected from six famous harbors. 
    The training, validation and test set include 436,181 and 444 images, respectively.
    
    \textit{ICDAR2015} \cite{karatzas2015icdar} is commonly used for oriented scene text detection and spotting. This dataset includes 1,000 training images and 500 testing images. 
    
    \textit{ICDAR2017 MLT} \cite{nayef2017icdar2017} is a multi-lingual text dataset, which includes 7,200 training images, 1,800 validation images and 9,000 testing images. The dataset is composed of complete scene images in 9 languages, and text regions in this dataset can be in arbitrary orientations, being more diverse and challenging.
    
    \textit{MSRA-TD500} \cite{yao2012detecting} is proposed for detecting long and oriented texts. It contains 300 training and 200 test images annotated in terms of text lines.
    
    \textit{FDDB} \cite{jain2010fddb} is a dataset designed for unconstrained face detection, in which faces have a wide variability of face scales, poses, and appearance. This dataset contains annotations for 5,171 faces in a set of 2,845 images taken from the faces in the Wild dataset~\cite{berg2005s}. In our paper, we manually use 70\% as the training set and the rest as the validation set.
    
    \begin{table}[tb!]
        \caption{Ablation test of GWD-based regression loss form and hyperparameter on DOTA. The based detector is RetinaNet.}
        \vspace{-7pt}
        \label{tab:func_tau}
        \centering
        \resizebox{0.48\textwidth}{!}{
            \begin{tabular}{c|cccc|c|c}
            \toprule
            $1-\frac{1}{\tau+f(\mathbf{D}_{w}^{2})}$ & $\tau=1$ & $\tau=2$ & $\tau=3$ & $\tau=5$ & $\mathbf{D}_{w}^{2}$ & Baseline \\
            \hline
            $f(\mathbf{D}_{w}^{2})=sqrt(\mathbf{D}_{w}^{2})$ & 68.56 & \textbf{68.93} & 68.37 & 67.77 & \multirow{2}{*}{49.11} & \multirow{2}{*}{65.73} \\
            $f(\mathbf{D}_{w}^{2})=\ln(\mathbf{D}_{w}^{2}+1)$ & 67.87 & 68.09 & 67.48 & 66.49 & \\
            \bottomrule
            \end{tabular}}
    \end{table}
    
    \begin{table}[tb!]
        \caption{Ablation of KLD regression losses using RetinaNet as based detector.}\vspace{-7pt}
        \label{tab:func_form}
        \centering
        \resizebox{0.48\textwidth}{!}{
            \begin{tabular}{c|c|c|c|c}
            \toprule
            Dataset & $\mathbf{D}_{kl}(\mathcal{N}_{p}||\mathcal{N}_{t})$ & $\mathbf{D}_{kl}(\mathcal{N}_{t}||\mathcal{N}_{p})$ & $\mathbf{D}_{js}(\mathcal{N}_{p}||\mathcal{N}_{t})$ & $\mathbf{D}_{jef}(\mathcal{N}_{p}||\mathcal{N}_{t})$ \\
            \hline
            DOTA-v1.0 & 70.17 & 70.55 & 69.67 & \textbf{70.56}\\
            HRSC2016 & 82.83 & 83.82 & \textbf{84.06} & 83.66\\
            \bottomrule
        \end{tabular}}
    \end{table}
    
    \begin{table}[tb!]
        \caption{Ablation study of normalization. The based detector is RetinaNet.}
        \vspace{-7pt}
        \label{tab:norm}
        \centering
        \resizebox{0.48\textwidth}{!}{
            \begin{tabular}{c|c|ccc|c}
            \toprule
            \multirow{2}{*}{Loss} & \multirow{2}{*}{Norm by Eq. \ref{eq:Lgwd}} & \multicolumn{3}{c|}{HRSC2016} & DOTA-v1.0\\
            \cline{3-6}
            & & Hmean$_{50}$ & Hmean$_{75}$ & Hmean$_{50:95}$ & AP$_{50}$ \\
            \hline
            \multirow{2}{*}{Smooth L1} & w/ & 78.99 & 43.12 & 43.47 & 64.95\\
            & w/o & \textbf{84.80} & \textbf{48.42} & \textbf{47.76} & \textbf{65.73}\\
            \bottomrule
            \end{tabular}}
    \end{table}
    
    \begin{table}[tb!]
        \caption{Ablation study under different BBox definitions.}
        \vspace{-7pt}
        \label{tab:ablation_rs}
        \centering
        \resizebox{0.48\textwidth}{!}{
            \begin{tabular}{c|c|c|c|c}
            \toprule
            Base Detector & Box Def. & Reg. Loss & Dataset & mAP$_{50}$ \\
            \hline
            \multirow{8}{*}{RetinaNet \cite{lin2017focal}} & \multirow{4}{*}{$D_{le}$} & Smooth L1 & \multirow{8}{*}{DOTA-v1.0} & 64.17 \\
            & & GWD & & 66.31 \textbf{\textcolor[rgb]{0,0.6,0}{\small (+2.14)}}\\
            & & BCD & & 68.56 \textbf{\textcolor[rgb]{0,0.6,0}{\small (+4.39)}}\\
            & & KLD & & \textbf{68.88 \textcolor[rgb]{0,0.6,0}{\small (+4.71)}}\\
            \cline{2-3} \cline{5-5}
            & \multirow{4}{*}{$D_{oc}$} & Smooth L1 & & 65.73 \\
            & & GWD & & 68.93 \textbf{\textcolor[rgb]{0,0.6,0}{\small (+3.20)}}\\
            & & BCD & & 71.23 \textbf{\textcolor[rgb]{0,0.6,0}{\small (+5.50)}}\\
            & & KLD & & \textbf{71.28 \textcolor[rgb]{0,0.6,0}{\small (+5.55)}}\\
            \bottomrule
            \end{tabular}}
    \end{table}
    
    \begin{table*}[tb!]
        \caption{High-precision detection experiment under different regression loss. `R', `F' and `G' indicate random rotation, flipping, and graying, respectively. The resolution of HRSC2016, MSRA-TD500, ICDAR2015 and FDDB are $500\times500$, $800\times1,000$, $800\times1,000$ and $800\times800$, respectively.}
        \label{tab:ablation_high_precision}
      \vspace{-7pt}
        \centering
        \resizebox{0.95\textwidth}{!}{
            \begin{tabular}{c|c|c|c|cccc|c}
            \toprule
            Base Detector & Dataset & Data Aug. & Reg. Loss & Hmean$_{50}$/AP$_{50}$ & Hmean$_{60}$/AP$_{60}$ & Hmean$_{75}$/AP$_{75}$ & Hmean$_{85}$/AP$_{85}$ & Hmean$_{50:95}$/AP$_{50:95}$ \\
            \hline
            \multirow{4}{*}{RetinaNet \cite{lin2017focal}} & \multirow{8}{*}{HRSC2016} & \multirow{8}{*}{R+F+G} & Smooth L1 & 84.28 & 74.74 & 48.42 & 12.56 & 47.76 \\
            & & &  GWD & 85.56 \textbf{\textcolor[rgb]{0,0.6,0}{\small (+1.28)}} & 84.04 \textbf{\textcolor[rgb]{0,0.6,0}{\small (+9.30)}} & 60.31 \textbf{\textcolor[rgb]{0,0.6,0}{\small (+11.89)}} & 17.14 \textbf{\textcolor[rgb]{0,0.6,0}{\small (+4.58)}} & 52.89 \textbf{\textcolor[rgb]{0,0.6,0}{\small (+5.13)}} \\
            & & & BCD & 86.38 \textbf{\textcolor[rgb]{0,0.6,0}{\small (+2.10)}} & 85.32 \textbf{\textcolor[rgb]{0,0.6,0}{\small (+10.58)}} & 68.50 \textbf{\textcolor[rgb]{0,0.6,0}{\small (+20.08)}} & 15.67 \textbf{\textcolor[rgb]{0,0.6,0}{\small (+3.11)}} & 55.09 \textbf{\textcolor[rgb]{0,0.6,0}{\small (+7.33)}} \\
            & & & KLD &  \textbf{87.45 \textcolor[rgb]{0,0.6,0}{\small (+3.17)}} & \textbf{86.72 \textcolor[rgb]{0,0.6,0}{\small (+11.98)}} & \textbf{72.39 \textcolor[rgb]{0,0.6,0}{\small (+23.97)}} & \textbf{27.68 \textcolor[rgb]{0,0.6,0}{\small (+15.12)}} & \textbf{57.80 \textcolor[rgb]{0,0.6,0}{\small (+10.04)}} \\
            \cline{1-1} \cline{4-9}
            \multirow{4}{*}{R$^3$Det \cite{yang2021r3det}} & &
            & Smooth L1 & 88.52 & 79.01 & 43.42 & 4.58 & 46.18  \\
            & & &  GWD & 89.43 \textbf{\textcolor[rgb]{0,0.6,0}{\small (+0.91)}} & 88.89 \textbf{\textcolor[rgb]{0,0.6,0}{\small (+9.88)}} & 65.88 \textbf{\textcolor[rgb]{0,0.6,0}{\small (+22.46)}} & 15.02 \textbf{\textcolor[rgb]{0,0.6,0}{\small (+10.44)}} & 56.07 \textbf{\textcolor[rgb]{0,0.6,0}{\small (+9.89)}} \\
            & & &  BCD &  \textbf{90.06 \textcolor[rgb]{0,0.6,0}{\small (+1.54)}} & \textbf{89.75 \textcolor[rgb]{0,0.6,0}{\small (+10.74)}} & \textbf{76.24 \textcolor[rgb]{0,0.6,0}{\small (+32.82)}} & \textbf{23.42 \textcolor[rgb]{0,0.6,0}{\small (+18.84)}} & \textbf{60.26 \textcolor[rgb]{0,0.6,0}{\small (+14.08)}} \\
            & & & KLD &  \textbf{89.97 \textcolor[rgb]{0,0.6,0}{\small (+1.45)}} & \textbf{89.73 \textcolor[rgb]{0,0.6,0}{\small (+10.72)}} & \textbf{77.38 \textcolor[rgb]{0,0.6,0}{\small (+33.96)}} & \textbf{25.12 \textcolor[rgb]{0,0.6,0}{\small (+20.54)}} & \textbf{61.40 \textcolor[rgb]{0,0.6,0}{\small (+15.22)}} \\
            \cline{1-9}
            \multirow{12}{*}{RetinaNet \cite{lin2017focal}} & \multirow{4}{*}{MSRA-TD500} &
            \multirow{4}{*}{R+F} & Smooth L1 & 70.98 & 62.42 & 36.73 & 12.56 & 37.89\\
            & & &  GWD & 76.76 \textbf{\textcolor[rgb]{0,0.6,0}{\small (+5.78)}}  & 68.58 \textbf{\textcolor[rgb]{0,0.6,0}{\small (+6.16)}}  & 44.21 \textbf{\textcolor[rgb]{0,0.6,0}{\small (+7.48)}}  & 17.75 \textbf{\textcolor[rgb]{0,0.6,0}{\small (+5.19)}}  & 43.62 \textbf{\textcolor[rgb]{0,0.6,0}{\small (+5.73)}} \\
            & & &  BCD &  75.24 \textbf{\textcolor[rgb]{0,0.6,0}{\small (+4.26)}} & 69.50 \textbf{\textcolor[rgb]{0,0.6,0}{\small (+7.08)}} & \textbf{48.13 \textcolor[rgb]{0,0.6,0}{\small (+11.40)}} & \textbf{20.33 \textcolor[rgb]{0,0.6,0}{\small (+7.77)}} & \textbf{45.26 \textbf{\textcolor[rgb]{0,0.6,0}{\small (+7.37)}} }\\
            & & & KLD & \textbf{76.96 \textcolor[rgb]{0,0.6,0}{\small (+5.98)}} & \textbf{70.08 \textcolor[rgb]{0,0.6,0}{\small (+7.66)}} & 46.95 \textbf{\textcolor[rgb]{0,0.6,0}{\small (+10.22)}} & 19.59 \textbf{\textcolor[rgb]{0,0.6,0}{\small (+7.03)}} & 45.24 \textbf{\textbf{\textcolor[rgb]{0,0.6,0}{\small (+7.35)}} }\\
            \cline{2-9}
            & \multirow{16}{*}{ICDAR2015} & \multirow{4}{*}{F} & Smooth L1 & 69.78 & 64.15 & 36.97 & 8.71 & 37.73\\
            & & &  GWD & 74.29 \textbf{\textcolor[rgb]{0,0.6,0}{\small (+4.51)}} & 68.34 \textbf{\textcolor[rgb]{0,0.6,0}{\small (+4.19)}} & 43.39 \textbf{\textcolor[rgb]{0,0.6,0}{\small (+6.42)}} & 10.50 \textbf{\textcolor[rgb]{0,0.6,0}{\small (+1.79)}} & 41.68 \textbf{\textcolor[rgb]{0,0.6,0}{\small (+3.95)}}\\
            & & &  BCD &  \textbf{76.63 \textcolor[rgb]{0,0.6,0}{\small (+6.85)}} & \textbf{71.07 \textcolor[rgb]{0,0.6,0}{\small (+6.92)}} & 43.10 \textbf{\textcolor[rgb]{0,0.6,0}{\small (+6.13)}} & 10.24 \textbf{\textcolor[rgb]{0,0.6,0}{\small (+1.53)}} & \textbf{42.78 \textcolor[rgb]{0,0.6,0}{\small (+5.05)}}\\
            & & & KLD & 75.32 \textbf{\textcolor[rgb]{0,0.6,0}{\small (+5.54)}} & 69.94 \textbf{\textcolor[rgb]{0,0.6,0}{\small (+5.79)}} & \textbf{44.46 \textcolor[rgb]{0,0.6,0}{\small (+7.49)}} & \textbf{10.70 \textcolor[rgb]{0,0.6,0}{\small (+1.99)}} & 42.68 \textbf{\textcolor[rgb]{0,0.6,0}{\small (+4.95)}}\\
            \cline{3-9}
            & & \multirow{4}{*}{R+F} & Smooth L1 & 74.83 & 69.46 & 42.02 & 11.59 & 41.98\\
            & & & GWD & 76.15 \textbf{\textcolor[rgb]{0,0.6,0}{\small (+1.32)}} & 71.26 \textbf{\textcolor[rgb]{0,0.6,0}{\small (+1.80)}} & \textbf{45.59 \textcolor[rgb]{0,0.6,0}{\small (+3.57)}} & \textbf{11.65 \textcolor[rgb]{0,0.6,0}{\small (+0.06)}} & 43.58 \textbf{\textcolor[rgb]{0,0.6,0}{\small (+1.60)}} \\
            & & &  BCD &  \textbf{78.03 \textcolor[rgb]{0,0.6,0}{\small (+3.20)}} & 72.50 \textbf{\textcolor[rgb]{0,0.6,0}{\small (+3.04)}} & 45.44 \textbf{\textcolor[rgb]{0,0.6,0}{\small (+3.42)}} & 10.53 \textcolor[rgb]{0.6,0.6,0.6}{\small (-1.06)} & 43.58 \textbf{\textcolor[rgb]{0,0.6,0}{\small (+1.60)}} \\
            & & & KLD &  77.92 \textbf{\textcolor[rgb]{0,0.6,0}{\small (+3.09)}} & \textbf{72.77 \textcolor[rgb]{0,0.6,0}{\small (+3.31)}} & 43.27 \textbf{\textcolor[rgb]{0,0.6,0}{\small (+1.25)}} & 11.09 \textcolor[rgb]{0.6,0.6,0.6}{\small (-0.50)} & \textbf{43.65 \textcolor[rgb]{0,0.6,0}{\small (+1.67)}} \\
            \cline{1-1} \cline{3-9}
            \multirow{8}{*}{R$^3$Det \cite{yang2021r3det}} &  &
            \multirow{4}{*}{F}  & Smooth L1 & 74.28 & 68.12 & 35.73 & 8.01 & 39.10\\
            & & &  GWD & 75.59 \textbf{\textcolor[rgb]{0,0.6,0}{\small (+1.31)}} & 68.36 \textbf{\textcolor[rgb]{0,0.6,0}{\small (+0.24)}} & 40.24 \textbf{\textcolor[rgb]{0,0.6,0}{\small (+4.51)}} & 9.15 \textbf{\textcolor[rgb]{0,0.6,0}{\small (+1.14)}} & 40.80 \textbf{\textcolor[rgb]{0,0.6,0}{\small (+1.70)}} \\
            & & &  BCD &  \textbf{79.02 \textcolor[rgb]{0,0.6,0}{\small (+4.74)}} & \textbf{72.82 \textcolor[rgb]{0,0.6,0}{\small (+4.70)}} & \textbf{45.68 \textcolor[rgb]{0,0.6,0}{\small (+9.95)}} & \textbf{10.42 \textcolor[rgb]{0,0.6,0}{\small (+2.41)}} & \textbf{44.22 \textcolor[rgb]{0,0.6,0}{\small (+5.12)}} \\
            & & & KLD &  \textbf{77.72 \textcolor[rgb]{0,0.6,0}{\small (+2.43)}} & \textbf{71.99 \textcolor[rgb]{0,0.6,0}{\small (+3.87)}} & \textbf{43.95 \textcolor[rgb]{0,0.6,0}{\small (+8.22)}} & \textbf{10.43 \textcolor[rgb]{0,0.6,0}{\small (+2.42)}} & \textbf{43.29 \textcolor[rgb]{0,0.6,0}{\small (+4.19)}} \\
            \cline{3-9}
            &  & \multirow{4}{*}{R+F}  & Smooth L1 & 75.53 & 69.69 & 37.69 & 9.03 & 40.56\\
            & & &  GWD & 77.09 \textbf{\textcolor[rgb]{0,0.6,0}{\small (+1.56)}} & 71.52 \textbf{\textcolor[rgb]{0,0.6,0}{\small (+1.83)}} & 41.08 \textbf{\textcolor[rgb]{0,0.6,0}{\small (+3.39)}} & 10.10 \textbf{\textcolor[rgb]{0,0.6,0}{\small (+1.07)}} & 42.17 \textbf{\textcolor[rgb]{0,0.6,0}{\small (+1.61)}} \\
            & & &  BCD &  \textbf{80.49 \textcolor[rgb]{0,0.6,0}{\small (+4.96)}} & \textbf{74.73 \textcolor[rgb]{0,0.6,0}{\small (+5.04)}} & \textbf{45.42 \textcolor[rgb]{0,0.6,0}{\small (+7.73)}} & \textbf{10.89 \textcolor[rgb]{0,0.6,0}{\small (+1.86)}} & \textbf{44.55 \textcolor[rgb]{0,0.6,0}{\small (+3.99)}} \\
            & & & KLD & 79.63 \textbf{\textcolor[rgb]{0,0.6,0}{\small (+4.63)}} & 73.30 \textbf{\textcolor[rgb]{0,0.6,0}{\small (+3.61)}} & 43.51 \textbf{\textcolor[rgb]{0,0.6,0}{\small (+5.82)}} & 10.61 \textbf{\textcolor[rgb]{0,0.6,0}{\small (+1.58)}} & 43.61 \textbf{\textcolor[rgb]{0,0.6,0}{\small (+3.05)}} \\
            \hline
            \multirow{4}{*}{RetinaNet \cite{lin2017focal}} & \multirow{4}{*}{FDDB} & \multirow{4}{*}{F} & Smooth L1 & 95.92 & 87.50 & 55.81 & 12.67 & 52.77 \\
            & & & GWD & 97.44 \textbf{\textcolor[rgb]{0,0.6,0}{\small (+1.52)}} & 94.68 \textbf{\textcolor[rgb]{0,0.6,0}{\small (+7.18)}} & 80.84 \textbf{\textcolor[rgb]{0,0.6,0}{\small (+25.03)}} & 36.38 \textbf{\textcolor[rgb]{0,0.6,0}{\small (+23.71)}} & 65.77 \textbf{\textcolor[rgb]{0,0.6,0}{\small (+13.00)}} \\
            & & & BCD & 96.67 \textbf{\textcolor[rgb]{0,0.6,0}{\small (+0.75)}} & 94.60 \textbf{\textcolor[rgb]{0,0.6,0}{\small (+7.10)}} & 83.09 \textbf{\textcolor[rgb]{0,0.6,0}{\small (+27.28)}} & 40.72 \textbf{\textcolor[rgb]{0,0.6,0}{\small (+28.05)}} & 67.03 \textbf{\textcolor[rgb]{0,0.6,0}{\small (+14.26)}} \\
            & & & KLD & \textbf{97.51 \textcolor[rgb]{0,0.6,0}{\small (+1.59)}} & \textbf{95.40 \textcolor[rgb]{0,0.6,0}{\small (+7.90)}} & \textbf{85.33 \textcolor[rgb]{0,0.6,0}{\small (+29.52)}} & \textbf{42.20 \textcolor[rgb]{0,0.6,0}{\small (+29.53)}} & \textbf{68.01 \textcolor[rgb]{0,0.6,0}{\small (+15.24)}} \\
            \bottomrule
        \end{tabular}}
    \end{table*}

    The model uses ResNet50 \cite{he2016deep} as the default backbone unless otherwise specified, and is initialized with ImageNet \cite{deng2009imagenet} pretrained weights.
    We perform experiments on six aerial benchmarks, three scene text benchmarks and one face benchmark to verify the generality of our techniques. Weight decay and momentum are set 0.0001 and 0.9, respectively. We use MomentumOptimizer over 4 GPUs with a total of 4 images per mini-batch (1 image per GPU).
    All the models are trained by 20 epochs in total, and learning rate is reduced tenfold at 12 epochs and 16 epochs, respectively. The initial learning rate for RetinaNet is 1e-3. The number of iterations per epoch for DOTA-v1.0, DOTA-v1.5, DOTA-v2.0, DIOR-R, UCAS-AOD, HRSC2016, ICDAR2015, MLT, MSRA-TD500 and FDDB are 54k, 64k, 80k, 17k, 5k, 10k, 10k, 10k, 5k and 4k respectively, and doubled if data augmentation (random rotation, flipping, and graying) and multi-scale training are used. 
    
    \subsection{3-D Datasets and Implementation Details}\label{sec:3-D_dataset}
    \textit{KITTI} \cite{geiger2012we} contains 7,481 training and 7,518 testing samples for 3-D object detection benchmark. The training samples are generally divided into the train split (3,712 samples) and the val split (3,769 samples). The evaluation is classified into Easy, Moderate or Hard according to the object size, occlusion and truncation. All results are evaluated by the mean average precision with a 3-D SkewIoU threshold of 0.7 for cars and 0.5 for pedestrian and cyclists. 
    
    \rev{\textit{Waymo Open Dataset} \cite{sun2020scalability} (WOD) is a dataset for autonomous driving. There are totally 1,150 sequences, including 798 sequences in training set with 158,801 LiDAR frames, 202 sequences in validation set with 39,987 LiDAR frames, and 150 sequences in test set with 29,647 LiDAR frames. The official 3-D detection evaluation metrics include the 3-D bounding box mean average precision (mAP) and mAP weighted by heading accuracy (mAPH). The mAP and mAPH are based on an 3-D SkewIoU threshold of 0.7 for vehicles and 0.5 for pedestrians and cyclists.}
    
    \rev{For experiments on KITTI and WOD, we use PointPillars \cite{lang2019pointpillars} as the baseline by plugging our new loss. 
    Experiments are all conducted with a single model for 3-class joint detection: vehicle, cyclists, and pedestrian.
    As a common protocol in 3-D detection, all the detectors are trained from scratch. The training schedule of PointPillars on KITTI follows that of MMDetection3D \cite{chen2019mmdetection}: AdamW optimizer \cite{loshchilov2018decoupled} with 48 samples per mini-batch (12 samples per GPU), a cosine-shaped one-cycle learning rate scheduler that spans 160 epochs. The learning rate starts from 1e-4 and reaches its peak value 1e-3 at the 60 epochs, and then goes down gradually to 1e-7 in the end. We train PointPillars on WOD with FP16 mixed-precision enabled and with similar schedule used by MMDetection3D: AdamW optimizer with 32 samples per mini-batch (8 samples per GPU), and with a linear learning rate scheduler that spans 24 epochs. The learning rate starts from 1e-3 and decays to 1e-4 and 1e-5 at the beginning of the $21^{st}$ and $24^{th}$ epochs respectively. 
    }
    
    \begin{table*}[tb!]
        \caption{Performance comparison on the KITTI \textit{val} split. Numbers are AP scores with 40 recall positions i.e. mAP quoted from~\cite{chen2019mmdetection}. `Mod.' denotes the moderate level of detection as defind by the dataset and the column `mAP mod.' refers to the overall mAP for the moderate level of objects.}
        \vspace{-7pt}
        \label{tab:kitti_val}
        \centering
        \resizebox{1.0\textwidth}{!}{
			\begin{tabular}{c||c|ccc|ccc|ccc|c|ccc|ccc|ccc}
				\toprule
				\multirow{2}{*}{Method} & \multicolumn{1}{c|}{mAP} & \multicolumn{3}{c|}{Car - 3-D Detection} & \multicolumn{3}{c|}{Ped. - 3-D Detection} & \multicolumn{3}{c|}{Cyc. - 3-D Detection} & \multicolumn{1}{c|}{mAP} & \multicolumn{3}{c|}{Car - BEV Detection} & \multicolumn{3}{c|}{Ped. - BEV Detection} & \multicolumn{3}{c}{Cyc. - BEV Detection}\\
				& Mod. & Easy & Mod. & Hard & Easy & Mod. & Hard & Easy & Mod. & Hard & Mod. & Easy & Mod. & Hard & Easy & Mod. & Hard & Easy & Mod. & Hard\\
				\hline
				PointPillars \cite{lang2019pointpillars} & 64.28 & 88.26 & 78.90 & \textbf{76.06} & 57.10 & 50.96 & 46.38 & 83.77 & 62.99 & 59.65 & 70.10 & \textbf{93.81} & 88.08 & \textbf{86.80} & 61.49 & 55.51 & 51.13 & 87.20 & 66.69 & 63.02 \\
				+ GWD & 65.50 & 87.38 & 78.57 & 75.87 & \textbf{61.69} & \textbf{55.19} & \textbf{50.04} & 81.61 & 62.74 & 59.18 & 71.48 & 92.02 & 88.30 & 85.72 & \textbf{64.67} & \textbf{58.49} & \textbf{53.45} & 86.92 & 67.66 & 63.37 \\
				+ BCD & 66.07 & 87.56 & 78.69 & 75.86 & 58.44 & 52.91 & 48.12 & \textbf{87.08} & \textbf{66.62} & \textbf{62.68} & \textbf{72.02} & 92.07 & \textbf{88.38} & 85.68 & 63.24 & 57.75 & 53.23 & \textbf{90.14} & \textbf{69.95} & \textbf{65.78} \\
				\ +KLD & \textbf{66.19} & \textbf{89.55} & \textbf{80.36} & 76.02 & 59.95 & 52.94 & 48.22 & 85.61 & 65.27 & 61.45 & 71.18 & 93.33 & 88.11 & 85.44 & 64.46 & 57.26 & 52.53 & 87.40 & 68.19 & 64.47 \\ 
				\bottomrule
		\end{tabular}}
		\vspace{-5pt}
    \end{table*}
    
    \begin{table}[tb!]
    \caption{\rev{Performance comparison \zhang{on Waymo Open Dataset \textit{val} set} with 202 sequences for 3-D object detection.}}
    \vspace{-7pt}
    \label{tab:waymo_val}
    \centering
    \resizebox{0.49\textwidth}{!}{
		\begin{tabular}{c|c|cccc|cccc}
			\toprule
			\multirow{2}{*}{Method} & \multirow{2}{*}{Difficulty} & \multicolumn{4}{c|}{mAP} & \multicolumn{4}{c}{mAPH} \\
			\cline{3-10}
			& & Veh. & Ped. & Cyc. & Overall & Veh. & Ped. & Cyc. & Overall\\
			\hline
			PointPillars & \multirow{4}{*}{Level 1} & 71.07 & 72.89 & 63.26 & 69.08 & 70.52 & 57.95 & 61.05 & 63.17 \\
			+GWD & & 72.89 & 72.86 & 62.75 & 69.50 & 72.35 & 59.23 & 60.74 & 64.11 \\
			+BCD & & 72.86 & \textbf{73.50} & \textbf{63.33} & \textbf{69.90} & 72.34 & \textbf{59.74} & \textbf{61.30} & \textbf{64.46} \\
			+KLD & & \textbf{72.93} & 71.87 & 62.83 & 69.21 & \textbf{72.41} & 58.07 & 60.83 & 63.77\\
			\hline
			PointPillars & \multirow{4}{*}{Level 2} & 62.84 & 64.82 & 60.89 & 62.85 & 62.34 & 51.31 & 58.76 & 57.47 \\
			+GWD & & 64.60 & 64.71 & 60.43 & 63.25 & 64.11 & 52.41 & 58.49 & 58.34\\
			+BCD & & 64.58 & \textbf{65.47} & \textbf{60.95} & \textbf{63.66} & 64.10 & \textbf{53.01} & \textbf{59.00} & \textbf{58.70} \\
			+KLD & & \textbf{64.62} & 63.73 & 60.48 & 62.94 & \textbf{64.14} & 51.32 & 58.56 & 58.01 \\
			\bottomrule
		\end{tabular}}
    \end{table}
    
    \begin{table}[tb!]
        \caption{\zhang{Ablation of heading post-processing on Waymo Open Dataset \textit{val} set.}}
        \vspace{-7pt}
        \label{tab:post_processing}
        \centering
        \resizebox{0.49\textwidth}{!}{
			\begin{tabular}{c|c|c|cccc|cccc}
				\toprule
				\multirow{2}{*}{Method} & \multirow{2}{*}{Post-Proc.} & \multirow{2}{*}{Difficulty} & \multicolumn{4}{c|}{mAP} & \multicolumn{4}{c}{mAPH} \\
				\cline{4-11}
				& & & Veh. & Ped. & Cyc. & Overall & Veh. & Ped. & Cyc. & Overall\\
				\hline
				\multirow{4}{*}{\shortstack{PointPillars\\+GWD}} & w/o & \multirow{2}{*}{Level 1} & \textbf{73.54} & \textbf{74.09} & \textbf{64.24} & \textbf{70.62} & \textbf{72.99} & 55.15 & \textbf{61.77} & 63.30 \\
				& w/ & & 72.89 & 72.86 & 62.75 & 69.50 & 72.35 & \textbf{59.23} & 60.74 & \textbf{64.11} \\
				\cline{2-11}
				& w/o & \multirow{2}{*}{Level 2} & \textbf{64.94} & \textbf{65.76} & \textbf{61.88} & \textbf{64.19} & \textbf{64.44} & 48.76 & \textbf{59.49} & 57.57 \\
				& w/ & & 64.60 & 64.71 & 60.43 & 63.25 & 64.11 & \textbf{52.41} & 58.49 & \textbf{58.34}\\
				\bottomrule
		\end{tabular}}
    \end{table}

    \subsection{Ablation Study and Further Comparison}\label{sec:ablation}

    \textit{Ablation study of GWD-based regression loss form and hyperparameter:} 
    Tab. \ref{tab:func_tau} compares two forms of GWD-based loss. The performance of directly using GWD ($\mathbf{D}_{w}^{2}$) as the regression loss is extremely poor: \text{49.11\%}, due to its rapid growth trend, as shown in the blue curve on the left of Fig. \ref{fig:loss_form}. In other words, the regression loss $\mathbf{D}_{w}^{2}$ is too sensitive to large errors. In contrast, Eq. \ref{eq:Lgwd} achieves a significant improvement by fitting IoU loss. Eq. \ref{eq:Lgwd} introduces two new hyperparameters, the non-linear function $f(\cdot)$ to transform the Wasserstein distance, and the constant $\tau$ to modulate the entire loss. From Tab. \ref{tab:func_tau}, the overall performance of using $sqrt$ outperforms that using $\ln$, about \text{0.98$\pm$0.3\%} higher. 
    For $f(\cdot)=sqrt$ with $\tau=2$, the model achieves the best performance: \text{68.93\%}. The results are consistent for BCD and KLD. All the subsequent experiments follow this setting for hyperparameters unless otherwise specified.
    
    \begin{table}[tb!]
        \centering
        \caption{More ablation experiments on more datasets (MLT and UCAS-AOD).}
        \label{tab:ablation_more_dataset}\vspace{-7pt}
        \resizebox{0.48\textwidth}{!}{
            \begin{tabular}{c|c|c|ccc}
            \toprule
            \multirow{2}{*}{Base Detector} & \multirow{2}{*}{Reg. Loss} & \multicolumn{1}{c|}{MLT} & \multicolumn{3}{c}{UCAS-AOD} \\
            \cline{3-6}
            & & Hmean$_{50}$ & car & plane & mAP$_{50}$\\
            \hline
            \multirow{4}{*}{RetinaNet} & Smooth L1 & 48.42 & 92.62 & 96.50 & 94.56  \\
            & GWD & 54.58 \textbf{\textcolor[rgb]{0,0.6,0}{\small(+6.16)}} & 94.03 \textbf{\textcolor[rgb]{0,0.6,0}{\small(+1.41)}} & 96.86 \textbf{\textcolor[rgb]{0,0.6,0}{\small(+0.36)}} & 95.44 \textbf{\textcolor[rgb]{0,0.6,0}{\small(+0.88)}}  \\
            &  BCD & 56.79
            \textbf{\textcolor[rgb]{0,0.6,0}{\small(+8.37)}} & \textbf{94.99
            \textcolor[rgb]{0,0.6,0}{\small(+2.37)}} & \textbf{98.10 \textcolor[rgb]{0,0.6,0}{\small(+1.60)}} & \textbf{96.54 \textcolor[rgb]{0,0.6,0}{\small(+1.98)}}  \\
            & KLD & \textbf{57.59 \textcolor[rgb]{0,0.6,0}{\small(+9.17)}} & 94.34
            \textbf{\textcolor[rgb]{0,0.6,0}{\small(+1.72)}} & 97.94
            \textbf{\textcolor[rgb]{0,0.6,0}{\small(+1.44)}} & 96.14
            \textbf{\textcolor[rgb]{0,0.6,0}{\small(+1.58)}}\\
            \bottomrule
        \end{tabular}}
        
    \end{table}
    
    \begin{table*}[tb!]
        \caption{Comparison between different solutions for inconsistency between metric and loss (IML), boundary discontinuity (BD) and square-like problem (SLP) on DOTA dataset. The $\checkmark$ indicates that the method has corresponding problem. $^\dagger$ and $^\ddagger$ represent the large aspect ratio object and the square-like object, respectively. The bold \textbf{\color{red}{red}} and \textbf{\color{blue}{blue}} fonts indicate the top two performances respectively.}
        \label{tab:ablation_study}
        \vspace{-7pt}
        \centering
        \resizebox{1.0\textwidth}{!}{
            \begin{tabular}{c|c|c|c|cc|c|ccccc|cc|cc|ccc|c|c}
            \toprule
            \multirow{2}{*}{Base Detector} & \multirow{2}{*}{Solution} & \multirow{2}{*}{Box Def.} & \multirow{2}{*}{IML} & \multicolumn{2}{c|}{BD} & \multirow{2}{*}{SLP} & \multicolumn{9}{c|}{v1.0 tranval/test} & \multicolumn{3}{c|}{v1.0 train/val} & \multicolumn{1}{c|}{v1.5} & \multicolumn{1}{c}{v2.0}\\
            \cline{5-6} \cline{8-21}
            & & & & EoE & PoA & & BR$^\dagger$ & SV$^\dagger$ & LV$^\dagger$ & SH$^\dagger$ & HA$^\dagger$ & ST$^\ddagger$ & RA$^\ddagger$ & 7-mAP$_{50}$ & mAP$_{50}$ & mAP$_{50}$ & mAP$_{75}$ & mAP$_{50:95}$ & mAP$_{50}$ & mAP$_{50}$\\
            \hline
            \multirow{12}{*}{RetinaNet \cite{lin2017focal}} & Reg. ($\Delta \theta$) & $D_{oc}$ & $\checkmark$ & $\checkmark$ & $\checkmark$ & $\times$ & 42.17 & 65.93 & 51.11 & 72.61 & 53.24 & 78.38 & 62.00 & 60.78 & 65.73 & 64.70 & 32.31 & 34.50 & 58.87 & 44.16\\
            & Reg. ($\Delta \theta$) & $D_{le}$ & $\checkmark$ & $\times$ & $\checkmark$ & $\checkmark$ & 38.31 & 60.48 & 49.77 & 68.29 & 51.28 & 78.60 & 60.02 & 58.11 & 64.17 & 62.21 & 26.06 & 31.49 & 56.10 & 43.06\\
            & Reg.$^*$ ($\sin{\theta}$, $\cos{\theta}$) & $D_{le}$ & $\checkmark$ & $\times$ & $\times$ & $\checkmark$ & 41.52 & 63.94 & 44.95 & 71.18 & 53.22 & 78.11 & 60.54 & 59.07 & 65.78 & 63.22 & 30.63 & 33.19 & 57.17 & 43.92\\
            & IoU-Smooth L1 \cite{yang2019scrdet} & $D_{oc}$ & $\checkmark$ & $\times$ & $\times$ & $\times$ & \textbf{\color{blue}{44.32}} & 63.03 & 51.25 & 72.78 & 56.21 & 77.98 & 63.22 & 61.26 & 66.99 & 64.61 & 34.17 & 36.23 & 59.17 & 46.31\\
            & Modulated \cite{qian2021learning} & $D_{oc}$ & $\checkmark$ & $\times$ & $\times$ & $\times$ & 42.92 & 67.92 & 52.91 & 72.67 & 53.64 & \textbf{\color{blue}{80.22}} & 58.21 & 61.21 & 66.05 & 63.50 & 33.32 & 34.61 & 57.75 & 45.17\\
            & Modulated \cite{qian2021learning} & Quad. & $\checkmark$ & $\times$ & $\times$ & $\times$ & 43.21 & 70.78 & 54.70 & 72.68 & 60.99 & 79.72 & 62.08 & 63.45 & 67.20 & 65.15 & 40.59 & 39.12 & \textbf{\color{blue}{61.42}} & 46.71\\
            & RIL \cite{ming2021optimization} & Quad. & $\checkmark$ & $\times$ & $\times$ & $\times$ & 40.81 & 67.63 & 55.45 & 72.42 & 55.49 & 78.09 & 64.75 & 62.09 & 66.06 & 64.07 & 40.98 & 39.05 & 58.91 & 45.35\\
            & CSL \cite{yang2020arbitrary} & $D_{le}$ & $\checkmark$ & $\times$ & $\times$ & $\checkmark$ & 42.25 & 68.28 & 54.51 & 72.85 & 53.10 & 75.59 & 58.99 & 60.80 & 67.38 & 64.40 & 32.58 & 35.04 & 58.55 & 43.34\\
            & DCL (BCL) \cite{yang2021dense} & $D_{le}$ & $\checkmark$ & $\times$ & $\times$ & $\times$ & 41.40 & 65.82 & 56.27 & 73.80 & 54.30 & 79.02 & 60.25 & 61.55 & 67.39 & 65.93 & 35.66 & 36.71 & 59.38 & 45.46\\
            & GWD (ours) & $D_{oc}$ & $\checkmark$ & $\times$ & $\times$ & $\times$ & 44.07 & 71.92 & 62.56 & 77.94 & 60.25 & 79.64 & 63.52 & 65.70 & 68.93 & 65.44 & 38.68 & 38.71 & 60.03 & 46.65\\
            & BCD (ours) & $D_{oc}$ & $\times$ & $\times$ & $\times$ & $\times$ & \textbf{\color{red}{45.16}} & \textbf{\color{blue}{74.04}} & \textbf{\color{blue}{72.19}} & \textbf{\color{blue}{84.07}} & \textbf{\color{blue}{65.07}} & \textbf{\color{red}{80.23}} & \textbf{\color{blue}{64.52}} & \textbf{\color{blue}{69.33}} & \textbf{\color{blue}{71.23}} & \textbf{\color{blue}{67.83}} & \textbf{\color{blue}{42.40}} & \textbf{\color{blue}{41.24}} & 60.78 & \textbf{\color{blue}{47.48}}\\
            & KLD (ours) & $D_{oc}$ & $\times$ & $\times$ & $\times$ & $\times$ & 44.00 & \textbf{\color{red}{74.45}} & \textbf{\color{red}{72.48}} & \textbf{\color{red}{84.30}} & \textbf{\color{red}{65.54}} & 80.03 & \textbf{\color{red}{65.05}} & \textbf{\color{red}{69.41}} & \textbf{\color{red}{71.28}} & \textbf{\color{red}{68.14}} & \textbf{\color{red}{44.48}} & \textbf{\color{red}{42.15}} & \textbf{\color{red}{62.50}} & \textbf{\color{red}{47.69}}\\ 
            \cline{1-21}
            \multirow{5}{*}{R$^3$Det \cite{yang2021r3det}} & Reg. ($\Delta \theta$) & $D_{oc}$ & $\checkmark$ & $\checkmark$ & $\checkmark$ & $\times$ & 44.15 & 75.09 & 72.88 & 86.04 & 56.49 & 82.53 & 61.01 & 68.31 & 70.66 & 67.18 & 38.41 & 38.46 & 62.91 & 48.43\\
            & DCL (BCL) \cite{yang2021dense} & $D_{le}$ & $\checkmark$ & $\times$ & $\times$ & $\times$ & 46.84 & 74.87 & 74.96 & 85.70 & 57.72 & \textbf{\color{red}{84.06}} & \textbf{\color{red}{63.77}} & 69.70 & 71.21 & 67.45 & 35.44 & 37.54 & 61.98 & 48.71\\
            & GWD (ours) & $D_{oc}$ & $\checkmark$ & $\times$ & $\times$ & $\times$ & 46.73 & \textbf{\color{blue}{75.84}} & 78.00 & \textbf{\color{blue}{86.71}} & 62.69 & \textbf{\color{blue}{83.09}} & 61.12 & 70.60 & 71.56 & \textbf{\color{red}{69.28}} & 43.35 & 41.56 & 63.22 & 49.25\\
            & BCD (ours) & $D_{oc}$ & $\times$ & $\times$ & $\times$ & $\times$ & \textbf{\color{blue}{47.80}} & \textbf{\color{red}{76.23}} & \textbf{\color{red}{79.20}} & \textbf{\color{red}{86.90}} & \textbf{\color{blue}{65.24}} & 83.07 & 61.20 & \textbf{\color{red}{71.38}} & \textbf{\color{red}{72.22}} & \textbf{\color{blue}{69.23}} & \textbf{\color{blue}{43.96}} & \textbf{\color{blue}{41.91}} & \textbf{\color{blue}{63.53}} & \textbf{\color{blue}{49.71}} \\
            & KLD (ours) & $D_{oc}$ & $\times$ & $\times$ & $\times$ & $\times$ & \textbf{\color{red}{48.34}} & 75.09 & \textbf{\color{blue}{78.88}} & 86.52 & \textbf{\color{red}{65.48}} & 82.08 & \textbf{\color{blue}{61.51}} & \textbf{\color{blue}{71.13}} & \textbf{\color{blue}{71.73}} & 68.87 & \textbf{\color{red}{44.48}} & \textbf{\color{red}{42.11}} & \textbf{\color{red}{65.18}} & \textbf{\color{red}{50.90}} \\
            \bottomrule
            \end{tabular}}
    \end{table*}
    
    \begin{table*}
    	\centering
    	\caption{Accuracy (\%) on DIOR-R. The short names c1-c20 for categories in our experiment are defined as: Airplane, Airport, Baseball field, Basketball court, Bridge, Chimney, Dam, Expressway service area, Expressway toll station, Golf field, Ground track field, Harbor, Overpass, Ship, Stadium, Storage tank, Tennis court, Train station, Vehicle, and Wind mill. $^\ddagger$ indicates that data augmentation and multi-scale training and testing are used.}
    	\vspace{-7pt}
    	\resizebox{1.0\textwidth}{!}{
    		\begin{tabular}{c|c|c|c|c|c|c|c|c|c|c|c|c|c|c|c|c|c|c|c|c|c|c}
    		\toprule
    		Base Detector & Solution & c1 &  c2 &  c3 &  c4 &  c5 &  c6 &  c7 &  c8 &  c9 &  c10 &  c11 &  c12 &  c13 &  c14 &  c15 & c16 & c17 & c18 & c19 & c20 &  mAP$_{50}$\\
    		\hline
    		\multirow{11}{*}{RetinaNet \cite{lin2017focal}} & \textit{Reg.} ($\Delta \theta$) & 57.10 & 28.85 & 66.69 & 80.50 & 21.00 & 72.38 & 17.72 & 59.90 & 50.56 & 73.05 & 73.18 & 16.10 & 37.44 & 54.20 & 63.64 & 43.57 & 79.56 & 35.40 & 25.77 & 51.59 & 50.41 \\
    		& \textit{Reg.$^*$} ($\sin{\theta}$, $\cos{\theta}$) & 57.67 & 28.48 & 68.65 & 80.52 & 19.29 & 72.51 & 22.25 & 64.16 & 52.45 & 73.58 & 76.26 & 18.59 & 39.82 & 54.21 & 58.00 & 43.50 & 79.74 & 40.49 & 25.73 & 53.24 & 51.46\\
    		& IoU-Smooth L1 \cite{yang2019scrdet} & 57.17 & 24.86 & 67.62 & 80.51 & 21.36 & 72.60 & 22.48 & 60.15 & 51.15 & 73.01 & 74.26 & 19.80 & 37.02 & 54.58 & 63.09 & 43.55 & 79.85 & 37.61 & 25.96 & 52.69 & 50.97\\
    		& RIL \cite{ming2021optimization} & 50.02 & 34.91 & 66.83 & 80.44 & 22.73 & 72.19 & 25.06 & 61.39 & 48.83 & 68.69 & 73.26 & 27.71 & 41.19 & 56.50 & 63.09 & 42.30 & 75.40 & 33.91 & 24.50 & 51.47 & 51.02\\
    		& CSL \cite{yang2020arbitrary} & 57.25 & 26.35 & 67.60 & 80.17 & 19.10 & 72.55 & 20.54 & 65.25 & 53.10 & 72.65 & 73.61 & 25.18 & 37.13 & 56.43 & 63.56 & 41.21 & 78.62 & 40.82 & 26.34 & 52.47 & 51.50 \\
    		& DCL \cite{yang2021dense} & 57.21 & 21.51 & 67.35 & 80.68 & 18.84 & 72.43 & 20.24 & 64.33 & 51.48 & 72.96 & 72.18 & 24.13 & 37.21 & 55.93 & 61.94 & 47.50 & 80.21 & 39.82 & 25.53 & 51.47 & 51.15\\
    		& Modulated \cite{qian2021learning} & 56.70 & 34.22 & 68.11 & 82.44 & 25.03 & 72.60 & 26.91 & 70.77 & 53.45 & 75.75 & 74.55 & 29.78 & 45.79 & 61.59 & 62.76 & 42.85 & 78.57 & 38.16 & 26.72 & 52.87 & 53.98 \\
    		& GWD (ours) & 59.22 & 20.55 & 69.43 & 80.85 & 16.99 & 72.58 & 20.18 & 64.03 & 53.18 & 72.24 & 76.33 & 17.10 & 35.27 & 58.34 & 68.11 & 44.15 & 80.81 & 37.90 & 26.13 & 54.23 & 51.38 \\
    		& BCD (ours) & 59.75 & 30.48 & 69.24 & 81.08 & 24.93 & 72.38 & 23.85 & 67.87 & 53.59 & 75.03 &  73.18 & 31.42 & 45.26 & 65.49 & 64.64 & 49.17 & 81.10 & 35.40 & 29.01 & 54.37 & 54.36 \\
    		& KLD (ours) & 59.12 & 33.23 & 68.91 & 81.25 & 27.82 & 75.45 & 26.76 & 73.15 & 53.16 & 76.56 & 77.46 & 33.29 & 47.05 & 66.03 & 65.56 & 43.83 & 81.12 & 37.41 & 28.52 & 54.43 & 55.50 \\
    		& KLD$^\ddagger$ (ours) & 78.45 & \textbf{48.09} & 76.53 & 89.80 & 34.38 & \textbf{77.37} & 33.13 & 84.64 & 66.56 & 76.03 & 82.66 & 40.69 & 52.61 & 79.61 & 74.68 & 61.37 & 88.18 & 47.90 & 38.40 & 63.56 & 64.73 \\
    		\hline
    		\multirow{3}{*}{FPN \cite{lin2017feature}} & \textit{Reg.} ($\Delta \theta$) & 62.36 & 30.31 & 71.13 & 80.62 & 29.65 & 72.26 & 22.91 & 72.50 & 65.66 & 73.53 & 76.55 & 26.19 & 45.62 & 78.89 & 68.83 & 71.18 & 80.95 & 38.58 & 47.04 & 63.50 & 58.91 \\
    		& KLD (ours) & 62.57 & 30.10 & 70.84 & 81.04 & 33.39 & 72.48 & 22.89 & 73.40 & 67.01 & 76.33 & 75.59 & 36.57 & 50.19 & 80.55 & 63.09 & 70.71 & 81.07 & 49.86 & 48.53 & 64.29 & 60.52 \\
    		& KLD$^\ddagger$ (ours) & \textbf{87.77} & 42.82 & \textbf{80.69} & \textbf{90.01} & \textbf{44.89} & 72.70 & \textbf{34.59} & \textbf{85.88} & \textbf{76.20} & \textbf{77.41} & \textbf{84.18} & \textbf{45.46} & \textbf{56.29} & \textbf{88.85} & \textbf{78.64} & \textbf{79.81} & \textbf{89.33} & \textbf{57.26} & \textbf{56.02} & \textbf{72.19} & \textbf{70.05} \\
    		\bottomrule
    	\end{tabular}}	
    	\label{table:dior_r}
    \end{table*}
    
    \begin{table}[tb!]
        \centering
        \caption{\rev{Performance comparison by mAP$_{50}$ of using different approximate SkewIoU losses on DOTA-v1.0 dataset. Base model is RetinaNet.}}
        \vspace{-7pt}
        \label{tab:iou_related}
        \resizebox{0.48\textwidth}{!}{
            \begin{tabular}{c|ccc|c}
            \toprule
            Reg. Loss & Implement & Consistency & Scale Invariance & mAP$_{50}$ \\
            \hline
            Smooth L1 \cite{ren2015faster} & easy & $\times$ & $\times$ & 64.55 \\
            PIoU \cite{chen2020piou} & medium & $\checkmark$ & $\checkmark$ & 65.85 \\
            plain SkewIoU \cite{zhou2019iou} & hard & $\checkmark$ & $\checkmark$ & 68.27 \\
            \hline
            GWD (ours) & easy & $\times$ & $\times$ & 67.05 \\
            KLD (ours) & easy & $\checkmark$ & $\checkmark$ & \textbf{69.94} \\
            \bottomrule
        \end{tabular}}
        
    \end{table}
    
        
    
    \begin{table}[tb!]
        \centering
        \caption{\rev{Evaluation by mAP of the combination of different strategies for label assignments and regression losses. }}
        \label{tab:label_assign}\vspace{-7pt}
        \resizebox{0.48\textwidth}{!}{
            \begin{tabular}{c|cc|c}
            \toprule
            Base Detector & Label Assignment & Regression Loss & AP$_{50}$\\
            \hline
            \multirow{5}{*}{RetinaNet-$D_{le}$ \cite{lin2017focal}} & Max-IoU & Smooth L1 & 68.56 \\
            & ATSS-IoU & Smooth L1 & 70.63 \textbf{\textcolor[rgb]{0,0.6,0}{\small(+2.07)}} \\
            & Max-IoU & KLD & 70.35 \textbf{\textcolor[rgb]{0,0.6,0}{\small(+1.79)}} \\
            & ATSS-IoU & KLD & 71.13 \textbf{\textcolor[rgb]{0,0.6,0}{\small(+2.57)}} \\
            & ATSS-KLD & KLD & \textbf{72.13 \textcolor[rgb]{0,0.6,0}{\small(+3.57)}} \\

            \bottomrule
        \end{tabular}}
        
    \end{table}
    
    \begin{table}[tb!]
      \caption{Performance evaluation of our KLD loss on horizontal detection.}
      \label{tab:kld_coco}\vspace{-7pt}
      \centering
      \resizebox{0.48\textwidth}{!}{
      \begin{tabular}{c|c|c|c|c|c|c|c}
            \toprule
            Detector & Regression Loss & AP & AP$_{50}$ & AP$_{75}$ & AP$_{s}$ & AP$_{m}$ & AP$_{l}$ \\
            \hline
            \multirow{3}{*}{RetinaNet \cite{lin2017focal}} & Smooth L1 & 37.2 & 56.6 & 39.7 & 21.4 & 41.1 & 48.0\\
            & GIoU & 37.4 & \textbf{56.7} & 39.7 & 22.2 & 41.7 & 48.1 \\
            & KLD & \textbf{38.0} & 56.4 & \textbf{40.6} & \textbf{23.3} & \textbf{43.2} & \textbf{49.3} \\
            \hline
            \multirow{3}{*}{Faster RCNN \cite{ren2015faster}} & Smooth L1 & 37.9 & \textbf{58.8} & 41.0 & 22.4 & 41.4 & 49.1 \\
            & GIoU & \textbf{38.3} & 58.7 & 41.5 & 22.5 & 41.7 & \textbf{49.7} \\
            & KLD & 38.2 & 58.7 & \textbf{41.7} & \textbf{22.6} & \textbf{41.8} & 49.3\\
            \hline
            \multirow{2}{*}{FCOS \cite{tian2019fcos}} & IoU & 36.6 & 56.0 & 38.8 & 21.0 & 40.6 & 47.0\\
            & KLD & \textbf{36.8} & \textbf{56.3} & \textbf{39.1} & \textbf{21.7} & \textbf{40.8} & \textbf{47.5} \\
            \bottomrule
        \end{tabular}}
    \end{table}
    
    \textit{Ablation study of KLD variants:} Keeping the same loss pattern, we compare four KLD-based distance functions in Tab. \ref{tab:func_form}, and conclude that the asymmetry of KLD does not have much impact on performance. In subsequent experiments, we use $\mathbf{D}_{kl}(\mathcal{N}_{p}||\mathcal{N}_{t})$ as the basic setting.
    
    \textit{Ablation study of normalization:} Note the extra normalization in Eq. \ref{eq:Lgwd} questions if the GWD/BCD/KLD actually contributes or simply produces noise in the results. Hence, we also perform a normalization operation on the Smooth L1 loss to eliminate the interference caused by normalization. Tab. \ref{tab:norm} shows a significant performance drop after normalization. These results show that the effectiveness of GWD/BCD/KLD does not come from Eq. \ref{eq:Lgwd}.
    
    \textit{Ablation under different rotating box definitions:} Tab. \ref{tab:ablation_rs} studies RetinaNet under different regression losses on DOTA-v1.0, and both rotating box definitions: $D_{le}$ and $D_{oc}$ are tested. For the Smooth L1 loss, the accuracy of $D_{le}$-based method is \text{1.56\%} lower than the $D_{le}$-based ones, at \text{64.17\%} and \text{65.73\%}, respectively. GWD/BCD/KLD-based methods obtain an increase by \text{2.14\%/4.39\%/4.71\%} and \text{3.20\%/5.50\%/5.55\%} under the above two definitions. Although Gaussian distribution modeling makes our method free from the choice of box definitions, it does not mean that the final performance of the two definition methods will be the same, as shown in Tab. \ref{tab:ablation_rs} (GWD: \text{66.31\%} vs. \text{68.93\%}; BCD: \text{68.56\%} vs. \text{71.23\%}; KLD: \text{68.88\%} vs. \text{71.28\%}). Different factors, such as order of edges and angle regression range, will still cause differences in model learning, but the methods based on Gaussian distribution need not to bind a certain definition. Therefore, $D_{oc}$ is used in all the subsequent experiments, unless otherwise specified.
    
    \textit{High-precision detection:} 
    Tab. \ref{tab:ablation_high_precision} shows results by two detectors on three datasets where the IoU for AP is at least 50\%. For HRSC2016 with a large number of ship of high aspect ratios, GWD achieves \text{11.89\%} improvement over Smooth L1 by AP$_{75}$, BCD and KLD even get \text{20.08\%} and \text{23.97\%} gain. Even with a stronger R$^3$Det detector, GWD/BCD/KLD still obtains improvement by \text{22.46\%}/\text{32.82\%}/\text{33.96\%} by AP$_{75}$, and \text{9.89\%}/\text{14.08\%}/\text{15.22\%} by AP$_{50:95}$. Similar results are obtained on MASR-TD500, ICDAR2015, FDDB that BCD/KLD output higher quality BBoxes than GWD and Smooth L1.
    
    \begin{table*}[tb!]
		\caption{\rev{Average precision (AP) of different objects on DOTA-v1.0. Here R-101 denotes ResNet-101 (likewise for R-50, R-152), and RX-101, and H-104 represent ResNeXt101~\cite{xie2017aggregated} and Hourglass-104~\cite{newell2016stacked}, respectively. MS indicates that multi-scale training/testing is used. $^\dagger$ means that the label assignment strategy is ATSS-KLD. The bold \textbf{\color{red}{red}} and \textbf{\color{blue}{blue}} fonts indicate the top two performances respectively.}}
		\vspace{-7pt}
        \label{tab:DOTA}
        \centering
		\resizebox{1.0\textwidth}{!}{
			\begin{tabular}{l|l|c|c|ccccccccccccccc|c}
				\toprule
				& Method & Backbone & MS &  PL &  BD &  BR &  GTF &  SV &  LV &  SH &  TC &  BC &  ST &  SBF &  RA &  HA &  SP &  HC &  mAP$_{50}$\\
				\hline
	            \multirow{10}{*}{\rotatebox{90}{\shortstack{Two-stage}}}
				&ICN \cite{azimi2018towards} & R-101 & $\checkmark$ & 81.40 & 74.30 & 47.70 & 70.30 & 64.90 & 67.80 & 70.00 & 90.80 & 79.10 & 78.20 & 53.60 & 62.90 & 67.00 & 64.20 & 50.20 & 68.20 \\
				&RoI-Trans. \cite{ding2018learning} & R-101 & $\checkmark$ & 88.64 & 78.52 & 43.44 & 75.92 & 68.81 & 73.68 & 83.59 & 90.74 & 77.27 & 81.46 & 58.39 & 53.54 & 62.83 & 58.93 & 47.67 & 69.56 \\
				&SCRDet \cite{yang2019scrdet} & R-101 & $\checkmark$ & 89.98 & 80.65 & 52.09 & 68.36 & 68.36 & 60.32 & 72.41 & 90.85 & 87.94 & 86.86 & 65.02 & 66.68 & 66.25 & 68.24 & 65.21 & 72.61\\
				&Gliding Vertex \cite{xu2020gliding} & R-101 & & 89.64 & 85.00 & 52.26 & 77.34 & 73.01 & 73.14 & 86.82 & 90.74 & 79.02 & 86.81 & 59.55 & \textbf{\color{red}{70.91}} & 72.94 & 70.86 & 57.32 & 75.02 \\
				&Mask OBB \cite{wang2019mask} & RX-101 & $\checkmark$ & 89.56 & \textbf{\color{red}{85.95}} & 54.21 & 72.90 & 76.52 & 74.16 & 85.63 & 89.85 & 83.81 & 86.48 & 54.89 & \textbf{\color{blue}{69.64}} & 73.94 & 69.06 & 63.32 & 75.33 \\
				&FPN-CSL \cite{yang2020arbitrary} & R-152 & $\checkmark$ & \textbf{\color{blue}{90.25}} & \textbf{\color{blue}{85.53}} & 54.64 & 75.31 & 70.44 & 73.51 & 77.62 & 90.84 & 86.15 & 86.69 & 69.60 & 68.04 & 73.83 & 71.10 & 68.93 & 76.17\\
				&RSDet-II \cite{qian2021learning} & R-152 & $\checkmark$ & 89.93& 84.45 & 53.77 & 74.35 & 71.52 & 78.31 & 78.12 & \textbf{\color{red}{91.14}} & 87.35 & 86.93 & 65.64 & 65.17 & 75.35 & \textbf{\color{red}{79.74}} & 63.31 & 76.34 \\
				& SCRDet++ \cite{yang2022scrdet++} & R-101 & $\checkmark$ & 90.05 & 84.39 & 55.44 & 73.99 & 77.54 & 71.11 & 86.05 & 90.67 & 87.32 & \textbf{\color{blue}{87.08}} & \textbf{\color{blue}{69.62}} & 68.90 & 73.74 & 71.29 & 65.08 & 76.81\\
				& ReDet \cite{han2021redet} & ReR-50 & $\checkmark$ & 88.81 & 82.48 & \textbf{\color{red}{60.83}} & \textbf{\color{blue}{80.82}} & \textbf{\color{blue}{78.34}} & \textbf{\color{red}{86.06}} & \textbf{\color{red}{88.31}} & 90.87 & \textbf{\color{red}{88.77}} & 87.03 & 68.65 & 66.90 & \textbf{\color{red}{79.26}} & \textbf{\color{blue}{79.71}} & \textbf{\color{blue}{74.67}} & \textbf{\color{blue}{80.10}} \\
				\cline{2-20}
				& FPN-BCD (ours) & R-152 & $\checkmark$ & \textbf{\color{red}{90.33}} & 85.43 & \textbf{\color{blue}{59.33}} & \textbf{\color{red}{82.11}} & \textbf{\color{red}{79.35}} & \textbf{\color{blue}{83.02}} & \textbf{\color{blue}{87.31}} & \textbf{\color{blue}{90.88}} & \textbf{\color{blue}{88.04}} & \textbf{\color{red}{87.18}} & \textbf{\color{red}{75.30}} & 66.73 & \textbf{\color{blue}{76.72}} & 74.77 & \textbf{\color{red}{75.61}} & \textbf{\color{red}{80.14}} \\
				\hline
				\hline
				\multirow{14}{*}{\rotatebox{90}{\shortstack{Single-stage}}} 
				&PIoU \cite{chen2020piou} & DLA-34 \cite{yu2018deep} & & 80.90 & 69.70 & 24.10 & 60.20 & 38.30 & 64.40 & 64.80 & \textbf{\color{red}{90.90}} & 77.20 & 70.40 & 46.50 & 37.10 & 57.10 & 61.90 & 64.00 & 60.50 \\
				&O$^2$-DNet \cite{wei2020oriented} & H-104 & $\checkmark$ & 89.31 & 82.14 & 47.33 & 61.21 & 71.32 & 74.03 & 78.62 & 90.76 & 82.23 & 81.36 & 60.93 & 60.17 & 58.21 & 66.98 & 61.03 & 71.04 \\
				& DAL \cite{ming2021dynamic} & R-101 & $\checkmark$ & 88.61 & 79.69 & 46.27 & 70.37 & 65.89 & 76.10 & 78.53 & 90.84 & 79.98 & 78.41 & 58.71 & 62.02 & 69.23 & 71.32 & 60.65 & 71.78\\
				&BBAVectors \cite{yi2021oriented} & R-101 & $\checkmark$ & 88.35 & 79.96 & 50.69 & 62.18 & 78.43 & 78.98 & \textbf{\color{blue}{87.94}} & 90.85 & 83.58 & 84.35 & 54.13 & 60.24 & 65.22 & 64.28 & 55.70 & 72.32 \\
				&DRN \cite{pan2020dynamic} & H-104 & $\checkmark$ & \textbf{\color{red}{89.71}} & 82.34 & 47.22 & 64.10 & 76.22 & 74.43 & 85.84 & 90.57 & 86.18 & 84.89 & 57.65 & 61.93 & 69.30 & 69.63 & 58.48 & 73.23 \\
				& PolarDet \cite{zhao2021polardet} & R-101 & $\checkmark$ & \textbf{\color{blue}{89.65}} & \textbf{\color{red}{87.07}} & 48.14 & 70.97 & \textbf{\color{blue}{78.53}} & \textbf{\color{blue}{80.34}} & 87.45 & 90.76 & 85.63 & 86.87 & 61.64 & \textbf{\color{blue}{70.32}} & 71.92 & 73.09 & 67.15 & 76.64 \\
				& DCL \cite{yang2021dense} & R-152 & $\checkmark$ & 89.10 & 84.13 & 50.15 & 73.57 & 71.48 & 58.13 & 78.00 & \textbf{\color{blue}{90.89}} & 86.64 & 86.78 & 67.97 & 67.25 & 65.63 & 74.06 & 67.05 & 74.06 \\
				& RDD \cite{zhong2020single} & R-101 & $\checkmark$ & 89.15 & 83.92 & 52.51 & 73.06 & 77.81 & 79.00 & 87.08 & 90.62 & \textbf{\color{blue}{86.72}} & \textbf{\color{blue}{87.15}} & 63.96 & 70.29 & \textbf{\color{blue}{76.98}} & 75.79 & 72.15 & 77.75 \\
				\cline{2-20}
				& GWD (ours) & R-152 & $\checkmark$ &  89.06 & 84.32 & \textbf{\color{blue}{55.33}} & 77.53 & 76.95 & 70.28 & 83.95 & 89.75 & 84.51 & 86.06 & \textbf{\color{red}{73.47}} & 67.77 & 72.60 & 75.76 & \textbf{\color{blue}{74.17}} & 77.43 \\
				\cline{2-20}
				& \multirow{2}{*}{BCD (ours)} &  \multirow{2}{*}{R-152} & & 88.80 & 84.41 & 53.73 & 70.26 & 77.85 & 76.31 & 85.18 & 90.83 & 85.91 & 85.61 & 64.77 & 64.15 & 76.60 & \textbf{\color{blue}{77.19}} & 71.27 & 76.86 \\
				& & & $\checkmark$ & 89.35 & \textbf{\color{blue}{86.35}} & 54.79 & \textbf{\color{blue}{80.98}} & 77.79 & 74.83 & 84.19 & 90.79 & 86.26 & 85.74 & \textbf{\color{blue}{72.02}} & 69.19 & \textbf{\color{red}{77.20}} & 76.39 & 71.91 & \textbf{\color{blue}{78.52}} \\
				\cline{2-20}
				& \multirow{2}{*}{KLD (ours)} & R-50 & & 88.91 & 83.71 & 50.10 & 68.75 & 78.20 & 76.05 & 84.58 & 89.41 & 86.15 & 85.28 & 63.15 & 60.90 & 75.06 & 71.51 & 67.45 & 75.28 \\
				& & R-50 & $\checkmark$ & 88.91  & 85.23 & 53.64  & \textbf{\color{red}{81.23}}  & 78.20  & 76.99  & 84.58  & 89.50  & \textbf{\color{red}{86.84}}  & 86.38  & 71.69  & 68.06  & 75.95  & 72.23  & \textbf{\color{red}{75.42}}  & 78.32 \\
				\cline{2-20}
				& \rev{KLD$^\dagger$ (ours)} & R-50 & $\checkmark$ & 89.00 & 83.70 & \textbf{\color{red}{56.03}} & 78.12 & \textbf{\color{red}{80.77}} & \textbf{\color{red}{85.35}} & \textbf{\color{red}{88.22}} & \textbf{\color{red}{90.90}} & 84.51 & \textbf{\color{red}{87.23}} & 66.71 & \textbf{\color{red}{73.44}} & 76.06 & \textbf{\color{red}{81.94}} & 70.91 & \textbf{\color{red}{79.53}}\\
				\hline
				\hline
				\multirow{14}{*}{\rotatebox{90}{\shortstack{Refine-stage}}} 
				& CFC-Net \cite{ming2021cfc} & R-101 & $\checkmark$ & 89.08 & 80.41 & 52.41 & 70.02 & 76.28 & 78.11 & 87.21 & \textbf{\color{red}{90.89}} & 84.47 & 85.64 & 60.51 & 61.52 & 67.82 & 68.02 & 50.09 & 73.50\\
				& R$^3$Det \cite{yang2021r3det} & R-152 & $\checkmark$ & 89.80 & 83.77 & 48.11 & 66.77 & 78.76 &
				83.27 & 87.84 & 90.82 & 85.38 & 85.51 & 65.67 & 62.68 & 67.53 & 78.56 & 72.62 & 76.47\\
				& DAL \cite{ming2021dynamic} & R-50 & $\checkmark$ & 89.69 & 83.11 & 55.03 & 71.00 & 78.30 & 81.90 & 88.46 & \textbf{\color{red}{90.89}} & 84.97 & \textbf{\color{blue}{87.46}} & 64.41 & 65.65 & 76.86 & 72.09 & 64.35 & 76.95 \\
				& DCL \cite{yang2021dense} & R-152 & $\checkmark$ & 89.26 & 83.60 & 53.54 & 72.76 & 79.04 & 82.56 & 87.31 & 90.67 & 86.59 & 86.98 & 67.49 & 66.88 & 73.29 & 70.56 & 69.99 & 77.37 \\
				& RIDet \cite{ming2021optimization} & R-50 & $\checkmark$ & 89.31 & 80.77 & 54.07  & 76.38   & \textbf{\color{blue}{79.81}}  & 81.99 & \textbf{\color{blue}{89.13}}  & 90.72  & 83.58   & 87.22  & 64.42  & 67.56  & 78.08  & 79.17  & 62.07  & 77.62  \\ 
				& S$^2$A-Net \cite{han2021align} & R-101 & $\checkmark$ & 89.28 & 84.11 & 56.95 & 79.21 & \textbf{\color{red}{80.18}} & 82.93 & \textbf{\color{red}{89.21}} & 90.86 & 84.66 & \textbf{\color{red}{87.61}} & 71.66 & 68.23 & \textbf{\color{blue}{78.58}} & 78.20 & 65.55 & 79.15\\
				\cline{2-20}
				& \multirow{2}{*}{R$^3$Det-GWD (ours)} & R-50 & $\checkmark$ & 88.89 & 83.58 & 55.54 & 80.46 & 76.86 & 83.07 &	86.85 &	89.09 &	83.09 &	86.17 &	71.38 & 64.93 & 76.21 & 73.23 & 64.39 & 77.58 \\
				& & R-152 & $\checkmark$ & 
				89.28 & 83.70 & \textbf{\color{red}{59.26}} & 79.85 & 76.42 & 83.87 & 86.53 & 89.06 & 85.53 & 86.50 & \textbf{\color{blue}{73.04}} & 67.56 & 76.92 & 77.09 & 71.58 & 79.08 \\
				\cline{2-20}
				& \multirow{3}{*}{R$^3$Det-BCD (ours)} &  R-50 & & 89.53 & 85.02 & 54.32 & 73.08 & 79.11 & 84.83 & 88.05 & 90.82 & 85.92 & 85.75 & 66.80 & 63.96 & 77.11 & 73.82 & 66.31 & 77.63 \\
				& &  R-50 & $\checkmark$ & 89.77 & \textbf{\color{blue}{86.11}} & 56.48 & \textbf{\color{red}{81.94}} & 79.37 & \textbf{\color{red}{85.03}} & 88.15 & 90.83 & \textbf{\color{blue}{86.77}} & 87.04 & 71.87 & 66.67 & 78.08 & 74.31 & 68.80 & 79.41 \\
				& &  R-152 & $\checkmark$ & 89.82 & \textbf{\color{red}{86.62}} & 57.36 & 79.68 & 79.73 & \textbf{\color{blue}{84.86}} & 87.92 & \textbf{\color{blue}{90.88}} & 86.74 & 86.92 & \textbf{\color{red}{73.44}} & 69.95 & \textbf{\color{red}{78.83}} & 75.01 & 76.11 & \textbf{\color{blue}{80.26}} \\
				\cline{2-20}
				& \multirow{3}{*}{R$^3$Det-KLD (ours)} & R-50 & & 88.90 & 84.17 & 55.80 & 69.35 & 78.72 & 84.08 & 87.00 & 89.75 & 84.32 & 85.73 & 64.74 & 61.80 & 76.62 & 78.49 & 70.89 & 77.36 \\
				& & R-50 & $\checkmark$ & \textbf{\color{blue}{89.90}} & 84.91 & \textbf{\color{blue}{59.21}} & 78.74 & 78.82 & 83.95 & 87.41 & 89.89 & 86.63 & 86.69 & 70.47 & \textbf{\color{blue}{70.87}} & 76.96 & \textbf{\color{blue}{79.40}} & \textbf{\color{blue}{78.62}} & 80.17 \\
				& & R-152 & $\checkmark$ & \textbf{\color{red}{89.92}} & 85.13 & 59.19 & \textbf{\color{blue}{81.33}} & 78.82 & 84.38 & 87.50 & 89.80 & \textbf{\color{red}{87.33}} & 87.00 & 72.57 & \textbf{\color{red}{71.35}} & 77.12 & \textbf{\color{red}{79.34}} & \textbf{\color{red}{78.68}} & \textbf{\color{red}{80.63}} \\
				\bottomrule
		\end{tabular}}
    \end{table*}
    
    \zhang{\textit{Ablation study on 3-D detection tasks:}
    We extended the framework based on Gaussian distribution modeling from 2-D to 3-D object detection. Tab. \ref{tab:kitti_val} shows the performance comparison in 3-D detection and BEV detection on KITTI val split, and significant performance improvements are also achieved. On the moderate level of 3-D detection, GWD, BCD and KLD improve the PointPillars by \text{1.22\%}, \text{1.79\%} and \text{1.91\%}. On the moderate level of BEV detection, BCD and GWD achieve gains of \text{1.38\%}, \text{1.92\%} and \text{1.08\%}. Tab. \ref{tab:waymo_val} shows similar trend on Waymo Open Dataset.}
    
    \zhang{\textit{Ablation study of heading post-processing:}
    We also conducted experiments to validate the effectiveness of the proposed heading post-processing technique. We trained PointPillars-GWD with heading direction binary classifier used in the original work as a reference. As shown in Tab. \ref{tab:post_processing}, our post-processing method applied to PointPillars improves the mAPH metric for the square-like category (pedestrian) by \text{4.08\%/3.65\%} (from a 2.8\%/2.55\% decrease to a 1.28\%/1.10\% increase) in terms of Level 1/2, but it also brings a slight performance degradation to mAP due to the need to learn new parameters (heading vectors), which are meaningless for mAP. Therefore, our proposed post-processing method can be useful to the case when the heading information is needed e.g. in autonomous driving, especially for square-like categories.}
     
    \textit{Ablation study on more datasets:}
    To make the results more credible, we continue to verify on the other two 2-D datasets, as shown in Tab. \ref{tab:ablation_more_dataset}. The improvements of GWD/BCD/KLD on MLT and UCAS-AOD are still considerable, with an increase of \text{6.16\%}/\text{8.37\%}/\text{9.17\%} and \text{0.88\%}/\text{1.98\%}/\text{1.58\%} respectively. 
    
    \textit{Comparison between different solutions:}
    For the three issues of inconsistency between metric and loss, boundary discontinuity and square-like problem, Tab. \ref{tab:ablation_study} compares the five different solutions, including IoU-Smooth L1 Loss \cite{yang2019scrdet}, Modulated loss \cite{qian2021learning}, RIL \cite{ming2021optimization}, CSL \cite{yang2020arbitrary}, DCL \cite{yang2021dense} on DOTA dataset. For fairness, these methods are all implemented on the same baseline, and are trained and tested under the same environment and hyperparameters. In particular, we detail the accuracy of the seven categories, including large aspect ratio (e.g. BR, SV, LV, SH, HA) and square-like object (e.g. ST, RD), which contain many corner cases in the dataset. These categories are assumed can better reflect the real-world challenges and advantages of our method. Many methods that solve the boundary discontinuity have achieved significant improvements in the large aspect ratio object category, and the methods that take into account the square-like problem perform well in the square-like object. 
    
    We compare the two baselines, \textit{Reg.} ($\Delta \theta$)  vs. \textit{Reg.$^*$} ($\sin{\theta}$, $\cos{\theta}$), provided in Sec. \ref{sec:loss}. Under the same BBox definition $D_{le}$, the indirect regression method has obvious performance advantages, especially for large aspect ratio objects, due to its immunity to boundary discontinuity.
    
    There is rarely a unified method to solve all problems, and methods are proposed for part of problems e.g. IoU-Smooth L1 Loss. However, the gradient direction of IoU-Smooth L1 Loss is still dominated by Smooth L1 loss, so the metric and loss cannot be regarded as truly consistent. In contrast, due to the three nice properties of Gaussian distribution modeling, it need not to make additional judgments and can elegantly solve all the problems. Without bells and whistles, the combination of RetinaNet and BCD/KLD directly surpasses R$^3$Det (\text{71.23\%}/\text{71.28\%} vs. \text{70.66\%} in AP$_{50}$ and \text{69.33\%}/\text{69.41\%} vs. \text{68.31\%} in 7-AP$_{50}$). Even combined with R$^3$Det, BCD/KLD can still further improve performance of the large aspect ratio object (\text{3.07\%}/\text{2.82\%} in 7-AP$_{50}$) and high-precision detection (\text{5.55\%}/\text{6.07\%} in AP$_{75}$ and \text{3.45\%}/\text{3.65\%} in AP$_{50:95}$). BCD-based and KLD-based methods show the best performance in almost all indicators. Fig.~\ref{fig:compare_vis} and Fig.~\ref{fig:high_precision_compare_vis} visualize the comparison between Smooth L1 loss-based and GWD-based detector. Similar conclusions can still be drawn on the more challenging datasets (DOTA-v1.5, DOTA-v2.0, and DIOR-R in Tab. \ref{table:dior_r}), with more tiny objects (less than 10 pixels). 
    
    \rev{\textit{Comparing different approximate SkewIoU losses:} Tab. \ref{tab:iou_related} compares the proposed techniques with the plain SkewIoU Loss \cite{zhou2019iou} and the recently proposed approximate SkewIoU loss (PIoU) \cite{chen2020piou} on DOTA-v1.0 under the same experimental conditions. Since the official PIoU and plain SkewIoU are implemented based on PyTorch \cite{paszke2017automatic}, we also implemented the PyTorch version\footnote{\url{https://github.com/open-mmlab/mmrotate}} \cite{zhou2022mmrotate} of KLD and GWD. Clearly, KLD outperforms all other losses, at \text{69.94\%}. 
    }
    
    \rev{\textit{Comparing combined strategies for different label assignments and regression losses:}
    Not limited to regression loss, the proposed KLD can also be used as an effective division basis for label assignment. Tab. \ref{tab:label_assign} shows that when KLD is applied to both label assignment and regression loss, the performance rises to 72.13\% with a gain of \text{3.57\%} compared with the baseline. Since KLD does not have a clear physical meaning like IoU, dynamic threshold calculation by ATSS \cite{zhang2020bridging} is required. Compared to the combination of ATSS-IoU label assignment and KLD loss, using ATSS-KLD can obtain further improvement from 71.13\% to 72.13\%.
    }
    
    \textit{Horizontal detection verification:}
    KLD can be degenerated into the common regression loss in horizontal detection (see Eq.~\ref{eq:kl_h}). Tab. \ref{tab:kld_coco} compares the regression loss Smooth L1 and IoU/GIoU for horizontal detection with our KLD loss on MS COCO \cite{lin2014microsoft}. KLD still performs competitively on the Faster RCNN, RetinaNet and FCOS, and even has an improvement of \text{0.6\%} on RetinaNet. The GT for rotation detection is the minimum circumscribed rectangle, which means that GT can well reflect the true scale and direction information. The ``horizontal special case'' described in this paper also meets the above requirements, and the horizontal circumscribed rectangle is equal to the minimum circumscribed rectangle at this time. Although the GT of MS COCO is a horizontal box, it is not the minimum circumscribed rectangle. It means that it loses direction and accurate scale information of the object. For example, for a baseball bat placed obliquely in the image, the height and width of its horizontal circumscribed rectangle do not represent the exact height and width of the object. This causes that when KLD is applied to MS COCO, the optimization mechanism of KLD that dynamically adjusts the angle gradient according to the aspect ratio is meaningless, which in our analysis may affect the final performance. Hence, we believe this is a defect in the dataset annotation itself. In fact, it is inappropriate to use MS COCO to discuss $\theta=0^\circ$, because this dataset discards $\theta$. In addition, $\theta=0^\circ$ describes the instances in the horizontal position, but not mean all instances of the dataset are in a horizontal position. This paper uses MS COCO to discuss the ``horizontal special case'' to show that even if the dataset has certain labeling defects, our KLD can still be robust.
    
    
    \subsection{Overall Comparison}
  Evaluation is performed on DOTA-v1.0 and HRSC2016.
    
    \textit{Results on DOTA-v1.0:} Tab. \ref{tab:DOTA} compares with state-of-the-art methods, which in fact use different image resolutions, network structures, training strategies and tricks, which make the comparison less direct. For overall performance, BCD-based and KLD-based methods have achieved the best mAP among the two-stage, single-stage and refine-stage methods, in \text{80.14\%}, \text{79.53\%} and \text{80.63\%} respectively.
    
    \begin{table}[tb!]
		\caption{Performance on HRSC2016 via different backbones. For mAP, the numbers in bracket `07' or `12' means following the 2007 or 2012 evaluation metric. `-' means the raw papers did not provide the results.}
        \label{tab:HRSC2016}\vspace{-7pt}
        \centering
		\resizebox{0.42\textwidth}{!}{
			\begin{tabular}{lccc}
				\toprule
				Method & Backbone & mAP$_{50}$ (07) & mAP$_{50}$ (12)\\
				
				\midrule
				RoI-Trans. \cite{ding2018learning} & R-101 & 86.20 & -- \\
				RSDet \cite{qian2021learning} & R-50 & 86.50 & --\\
				DRN \cite{pan2020dynamic} & H-104 & -- & 92.70 \\
				SBD \cite{liu2019omnidirectional} & R-50 & -- & 93.70 \\
				Gliding Vertex \cite{xu2020gliding} & R-101 & 88.20 & -- \\
				OPLD  \cite{song2020learning} & R-101 & 88.44 & --\\
				BBAVectors \cite{yi2021oriented} & R-101 & 88.60 & -- \\
				S$^2$A-Net \cite{han2021align} & R-101 & \textbf{\color{red}{90.17}} & 95.01 \\
				R$^3$Det \cite{yang2021r3det} & R-101 & 89.26 & 96.01 \\
                R$^3$Det-DCL \cite{yang2021dense} & R-101 & 89.46 & 96.41\\
                FPN-CSL \cite{yang2020arbitrary} & R-101 & 89.62 & 96.10 \\
                DAL \cite{ming2021dynamic} & R-101 & 89.77 & -- \\
                \hline
                R$^3$Det-GWD (ours) & R-101 & 89.85 & 97.37\\
                R$^3$Det-BCD (ours) & R-101 & \textbf{\color{blue}{90.07}} & \textbf{\color{blue}{97.42}}\\
                R$^3$Det-KLD (ours) & R-101 & 89.87 & \textbf{\color{red}{97.62}}\\
				\bottomrule
		\end{tabular}}
    \end{table}
    
    \textit{Results on HRSC2016:}
    It contains large aspect ratio ship instances with arbitrary orientation, posing a challenge to detection accuracy. Tab. \ref{tab:HRSC2016} shows that R$^3$Det variants: -GWD, -BCD and -KLD achieve competitive performances: \text{89.85\%}/\text{90.07\%}/\text{89.87\%} and \text{97.37\%}/\text{97.42\%}/\text{97.62\%} by the 2007/2012 evaluation metrics, respectively.
    
	
	\section{Conclusion}\label{sec:conclusion}
	\rev{We have shown how to model the rotated objects as Gaussian distributions and present a novel regression loss and an efficient label assignment strategy, to model the deviation between rotated BBoxes for object detection. 
	
	We extend Gaussian modeling to 3-D detection by devising a heading regression and post-processing method that takes advantage of the Gaussian loss while preserving the supervision to object’s heading angle in degeneration cases during training. Experimental results on benchmarks (2-D/3-D, aerial/text/face images) show its effectiveness.} 
	

	
	%

	\section*{Acknowledgment}
    This work was partly supported by National Key Research and Development Program of China (2020AAA0107600), Shanghai Municipal Science and Technology Major Project (2021SHZDZX0102), and National Science of Foundation China (U20B2068, 61972250, 72061127003). 
	
	\ifCLASSOPTIONcaptionsoff
	\newpage
	\fi
	
	\bibliographystyle{IEEEtran}
	\bibliography{IEEEegbib}

	\begin{IEEEbiography}[{\includegraphics[width=1in,height=1.25in,clip,keepaspectratio]{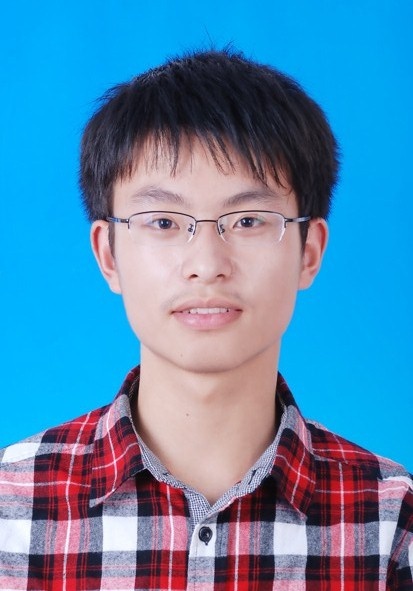}}]{Xue Yang} is a Ph.D. candidate in Computer Science, Shanghai Jiao Tong University, Shanghai, China. He received the B. E. in Automation, Central South University, Hunan, China, in 2016. He also received the M. S. from Chinese Academy of Sciences University, Beijing, China, in 2019. His research interest is computer vision. He published first-authored papers in TPAMI, IJCV, CVPR, ECCV, ICCV, ICML, NeurIPS, AAAI and ACM MM. He is also the leading contributor to the MMRotate and AlphaRotate open-source projects for oriented object detection, and with 5000+ stars in Github.


    \end{IEEEbiography}
    
    \begin{IEEEbiography}[{\includegraphics[width=1in,height=1.25in,clip,keepaspectratio]{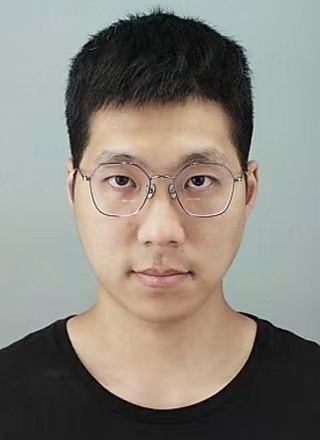}}]{Gefan Zhang} received his B.E. degree in automotive engineering at Tongji University in 2016. In 2018 he became a part-time student in pursuit of M.S. degree at the Department of Computer Science and Engineering, Shanghai Jiao Tong University. His research interest is 3-D object detection, especially lidar point cloud object detection. He is currently also working as a lidar perception engineer at COWA Robot and with focus on 3-D detection for autonomous driving.
    \end{IEEEbiography}
    
    \begin{IEEEbiography}[{\includegraphics[width=1in,height=1.25in,clip,keepaspectratio]{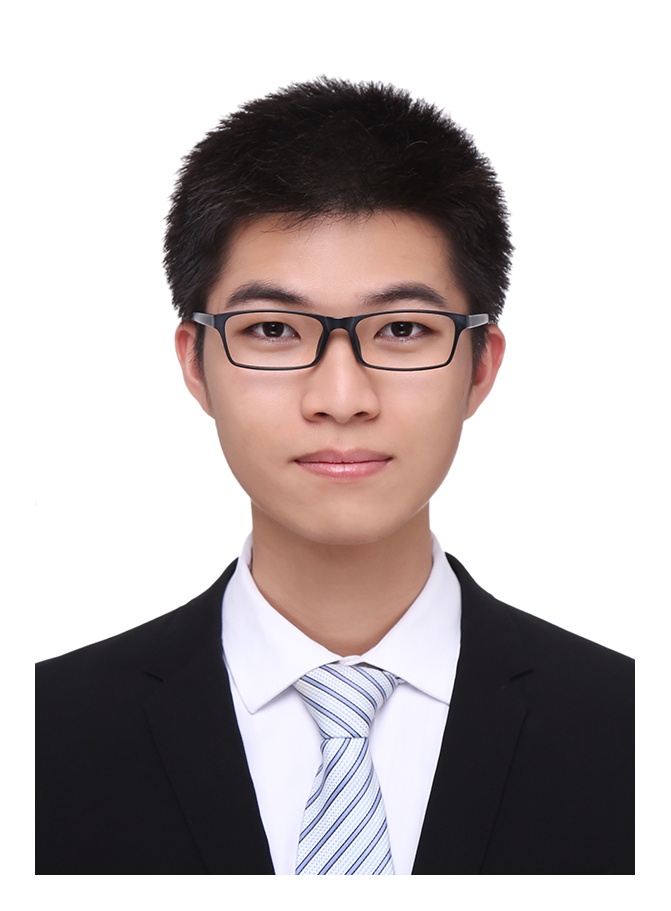}}]{Xiaojiang Yang} received the B.E. degree in Physics from Nankai University in 2018 (with minor in Math). He is currently working toward the Ph.D. degree in the Department of Computer Science and Engineering, Shanghai Jiao Tong University, Shanghai, China. His research interests include machine learning and vision, especially for generative models and representation learning. He has first-authored papers in top venues including ICLR and IEEE TNNLS.
    \end{IEEEbiography}
    
    \begin{IEEEbiography}[{\includegraphics[width=1in,height=1.25in,clip,keepaspectratio]{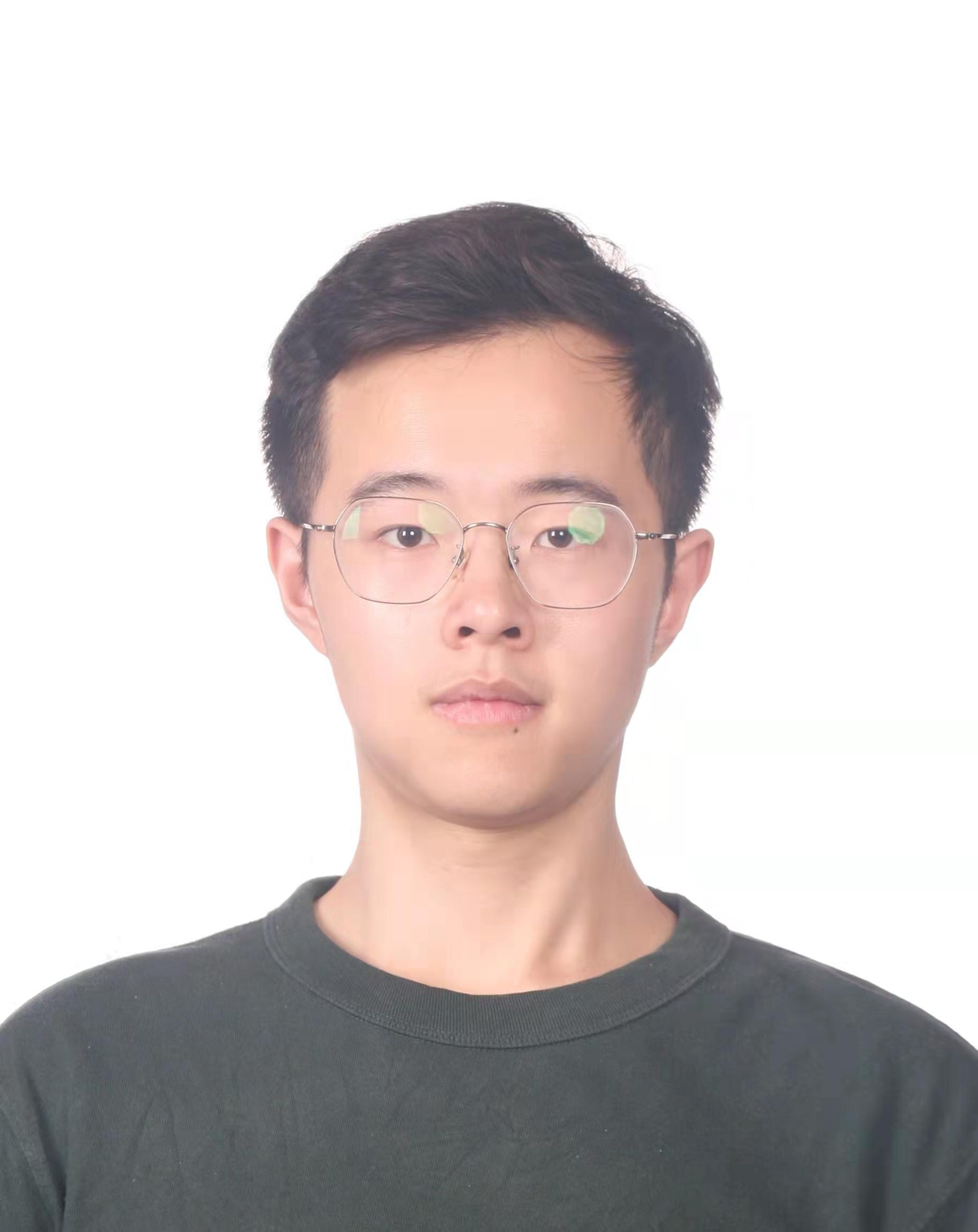}}]{Yue Zhou} received the B.S. degree in Electronic and Information Engineering from Beijing University of Posts and Telecommunications, Beijing, China, in 2017. He is currently pursuing the Ph.D. degree in Electrical  Engineering, Shanghai Jiao Tong University, Shanghai, China. His research interest is computer vision, especially for object detection. He has published papers in CVPR and the leading contributor to the MMRotate open-source project for rotation detection.
    \end{IEEEbiography}
    
    \begin{IEEEbiography}[{\includegraphics[width=1in,height=1.25in,clip,keepaspectratio]{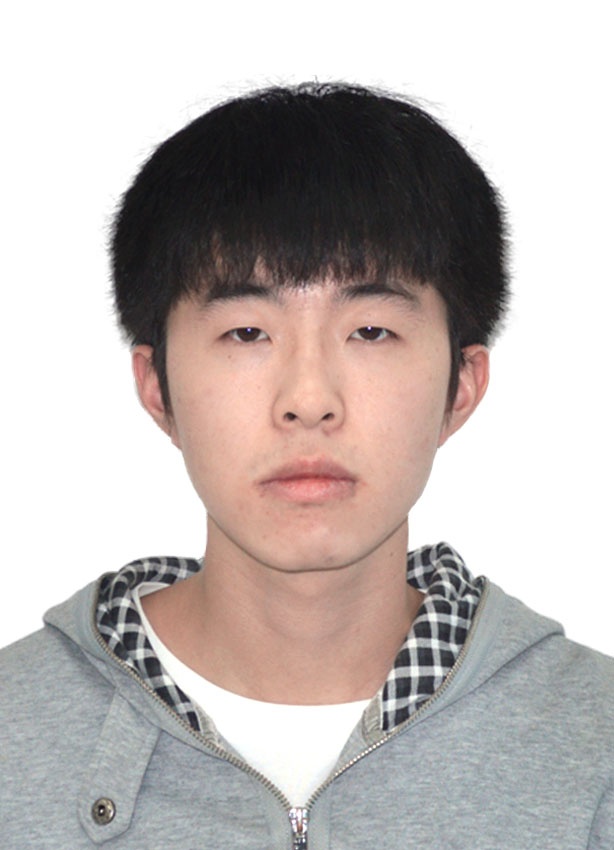}}]{Wentao Wang} received the B.E. degree from the Beijing University of Chemistry Technology, Beijing, China, in 2016 and received the M.S. degree from Institute of Electrics, Chinese Academy of Sciences, Beijing, China, in 2019. He is pursuing the Ph.D. degree in Computer Science, Shanghai Jiao Tong University, Shanghai, China. His research interests include deep learning and computer vision. He has published first-authored papers in CVPR, ICCV, ACM MM.
    \end{IEEEbiography}
    \begin{IEEEbiography}[{\includegraphics[width=1in,height=1.25in,clip,keepaspectratio]{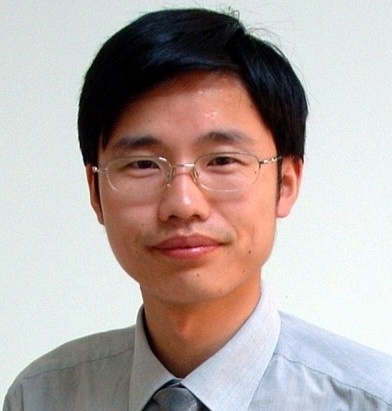}}]{Jin Tang} received the B.Eng. degree in automation and the Ph.D. degree in computer science from Anhui University, Hefei, China, in 1999 and 2007, respectively. He is currently a Professor with the School of Computer Science and Technology, Anhui University, Hefei, China. His current research interests include computer vision, pattern recognition, and deep learning, especially for remote sensing and object detection.
    \end{IEEEbiography}
    \begin{IEEEbiography}[{\includegraphics[width=1in,height=1.25in,clip,keepaspectratio]{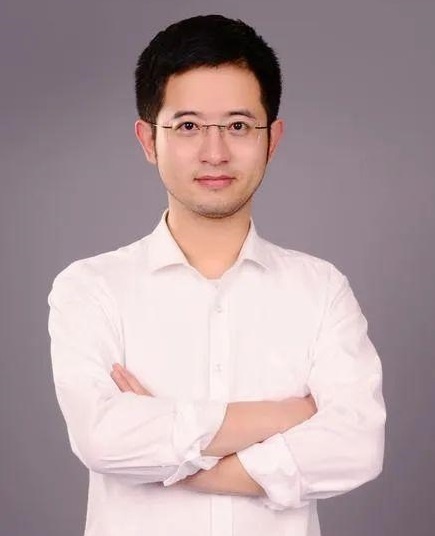}}]{Tao He} received the B. E. and M. E. degrees from Shanghai Jiao Tong University in Electrical Engineering in 2005 and 2008, respectively. He received his PhD in Mechanical and Aerospace Engineering, from Tokyo Institute of Technology, Tokyo, Japan in 2012. He was also once a post-doc with CMU. He has been working on autonomous driving for over one decade and currently he is the founder and the Chief Executive Officer (CEO) of COWAROBOT Co., Ltd. He is the winner of Forbes 40 under 40 China.
    \end{IEEEbiography}
	\begin{IEEEbiography}[{\includegraphics[width=1in,height=1.25in,clip,keepaspectratio]{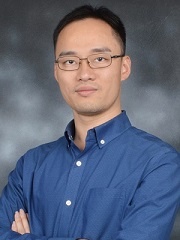}}]{Junchi Yan}(S'10-M'11-SM'21) is  an Associate Professor with Department of Computer Science and Engineering, Shanghai Jiao Tong University, Shanghai, China. He is also affiliated with Shanghai AI Laboratory, Shanghai, China. Before that, he was a Senior Research Staff Member with IBM Research where he started his career since April 2011, and obtained his PhD in Electrical Engineering, Shanghai Jiao Tong University in 2015. His research interest is machine learning. He served Area Chair for CVPR/AAAI/ICML/NeurIPS, Associate Editor for Pattern Recognition.
    \end{IEEEbiography}
	
\end{document}